\documentclass[10pt]{article} % For LaTeX2e
\usepackage[preprint]{tmlr}

\usepackage{bm}
\usepackage{amsmath, amsthm, amssymb,amsfonts,latexsym,mathtools}
\usepackage{graphicx}

\usepackage{hyperref}
\usepackage{url}
\newcommand{\Part}[3]{ \frac{ \partial^{#3} #1 }{ \partial #2^{#3} } }%partial derivative
\newcommand{\V}[1]{\bm{#1} } %vector command
\newcommand{\Tr}[1]{\mathrm{Tr}_{#1} }
\newcommand{\Ave}[1]{\left\langle {#1} \right\rangle} %thermal average 
\newcommand{\sgn}[1]{{\rm sgn}\left({#1} \right)}

\newcommand{\Extr}[1]{ \mathop{\rm Extr}_{ #1 } }

\newcommand{\mR}{\mathbb{R}}
\newcommand{\mN}{\mathbb{N}}
\newcommand{\lb}{\left(}
\newcommand{\rb}{\right)}
\newcommand{\lbb}{\left\{}
\newcommand{\rbb}{\right\}}
\newcommand{\lsb}{ \left[ }
\newcommand{\rsb}{ \right] }

\newcommand{\T}[1]{\tilde{#1}}
\newcommand{\Req}[1]{Eq.\ (\ref{eq:#1})} %{eq.\ (\ref{eq:#1})}

\newcommand{\NReq}[1]{(\ref{eq:#1})}
\newcommand{\BReqs}[2]{Eqs.\ (\ref{eq:#1}) and (\ref{eq:#2})}   %{Eqs.\ (\ref{eq:#1},\ref{eq:#2})}

\newcommand{\BReqss}[2]{Eqs.\ (\ref{eq:#1})--(\ref{eq:#2})}

\newcommand{\Rfig}[1]{Fig.\ \ref{fig:#1}}
\newcommand{\Rfigs}[2]{Figs.\ \ref{fig:#1} and \ref{fig:#2}}
\newcommand{\Rfigss}[2]{Figs.\ \ref{fig:#1}--\ref{fig:#2}}
\newcommand{\NRfig}[1]{\ref{fig:#1}}
\newcommand{\Lfig}[1]{\label{fig:#1}}
\newcommand{\Leq}[1]{\label{eq:#1}}
\newcommand{\Rsec}[1]{sec.\ \ref{sec:#1}}
\newcommand{\Rsecss}[2]{secs.\ \ref{sec:#1}--\ref{sec:#2}}
\newcommand{\Lsec}[1]{\label{sec:#1}}
\newcommand{\be}{\begin{eqnarray}}
\newcommand{\ee}{\end{eqnarray}}
\newcommand{\ba}{\begin{array}}
\newcommand{\ea}{\end{array}}
\newcommand{\no}{\nonumber}

\newcommand{\subbe}{\begin{subequations}}
\newcommand{\subee}{\end{subequations}}

\newcommand{\mc}[1]{\mathcal{#1}}
\newcommand{\argmax}{\mathop{\rm arg\,max}\limits}
\newcommand{\argmin}{\mathop{\rm arg\,min}\limits}

\newcommand{\E}{\mathbb{E}}

\newcommand{\out}{\mathcal{M}}
\newcommand{\loss}{\mathcal{L}}

\newcommand{\lossS}{\mathcal{\ell}}
\newcommand{\dloss}{g}
\newcommand{\Dz}{\int Dz\,}
\newcommand{\fe}{\phi}
\newcommand{\remid}[1]{\mathrel{}\middle#1\mathrel{}}

\title{When resampling/reweighting improves feature learning in imbalanced classification? A toy-model study}
\author{
      \name Tomoyuki Obuchi \email obuchi@i.kyoto-u.ac.jp \\
      \addr Department of Systems Science\\
      Kyoto University
      \AND
      \name Toshiyuki Tanaka \email tt@i.kyoto-u.ac.jp \\
      \addr Department of Systems Science\\
      Kyoto University
      }

\def\month{4}  % Insert correct month for camera-ready version
\def\year{2025} % Insert correct year for camera-ready version
\def\openreview{\url{https://openreview.net/forum?id=spqbyeGyLR}} % Insert correct link to OpenReview for camera-ready version

\begin{document} 
\maketitle

%%%%%%%%%%%%%%%%%%%%%%%%%%%%%%%%%%%%%%%%%%%%%%%%%%%%%
%%%%%%%%%%%%%%%%%%%%%%%%%%%%%%%%%%%%%%%%%%%%%%%%%%%%%
%%%%%%%%%%%%%%%%%%%%%%%%%%%%%%%%%%%%%%%%%%%%%%%%%%%%%
\begin{abstract}
A toy model of binary classification is studied with the aim of clarifying the class-wise resampling/reweighting effect on the feature learning performance under the presence of class imbalance. In the analysis, a high-dimensional limit of the input space is taken while keeping the ratio of the dataset size against the input dimension finite and the non-rigorous replica method from statistical mechanics is employed. The result shows that there exists a case in which the no resampling/reweighting situation gives the best feature learning performance irrespectively of the choice of losses or classifiers, supporting recent findings in~\citet{kang2019decoupling,cao2019learning}. It is also revealed that the key of the result is the symmetry of the loss and the problem setting. Inspired by this, we propose a further simplified model exhibiting the same property in the multiclass setting. These clarify when the class-wise resampling/reweighting becomes effective in imbalanced classification. 
\end{abstract}
%\keywords{Sensing, Image processing, Super-resolution imaging, Black hole, sparse modelling}

%%%%%%%%%%%%%%%%%%%%%%%%%%%%%%%%%%%%%%%%%%%%%%%%%%%%%
%%%%%%%%%%%%%%%%%%%%%%%%%%%%%%%%%%%%%%%%%%%%%%%%%%%%%
%%%%%%%%%%%%%%%%%%%%%%%%%%%%%%%%%%%%%%%%%%%%%%%%%%%%%
\section{Introduction}\Lsec{Introduction}
Real-world datasets for classification occasionally exhibit strong class imbalance with a long-tailed class distribution~\citep{van2018inaturalist,iNaturalist2018,liu2019large}.
Classifiers %based on deep neural networks (DNN)
applied to such datasets tend to perform poorly for minority classes, which poses a major challenge in areas such as visual recognition.
Although several methods to mitigate class imbalance have been proposed so far~\citep{chawla2002smote,he2009learning,6137280},
recent advances of deep learning have shed new light on this issue,
resulting in numerous studies from the perspective of applying those approaches to classifiers based on deep neural networks (DNNs)~\citep{liu2019large,huang2016learning,wang2017learning,cui2018large,8953234,cui2019class,cao2019learning,kang2019decoupling,jamal2020rethinking,tan2020equalization,menon2020long,kini2021label}. %Many key concepts in these methods can be applied not only to DNNs but also to a wider range of classification models~\citep{chawla2002smote,he2009learning,6137280}.

Among those approaches proposed so far, we focus on two simple strategies, reweighting and resampling, which are commonly employed to mitigate class imbalance. The resampling strategy tries to balance the samples in the dataset by oversampling the minority classes and/or undersampling the majority classes, while the reweighting strategy puts an additional weight to each term of the loss %according to some criterion: a simple way focused here is based on the class size.
in order to counterweight the class imbalance. 
The effectiveness of these strategies has been empirically verified in a wide range of studies~\citep{cui2019class,cao2019learning,kang2019decoupling,jamal2020rethinking,chawla2002smote,he2009learning}.
In spite of these pieces of work, transparent description or understanding about when they are useful or not would still be incomplete.
In particular, how class imbalance may affect
the quality of feature learning would be an important problem
in the context of representation learning in DNNs, 
but a thorough understanding of this issue is still missing.
%although there is a nice analysis on the decision boundary location~\citep{6137280}. 

Recently, \citet{kang2019decoupling} reported an interesting observation that feature learning becomes better if no resampling is applied. More specifically, on the basis of their extensive experiment on visual recognition tasks using DNNs, they reported that the best classification performance was achieved when the whole network was first trained without any resampling and then only the last output layer (final classifier) was retrained with class-balanced resampling. This observation can be interpreted as follows: one can learn the best feature representation in the initial training phase if one does not use resampling at all, and the good classification performance achieved by the retrained network is ascribed to exploitation of the good feature representation acquired in the initial training. A similar behavior was also reported in~\citet{cao2019learning}. One can therefore expect that Kang et al.'s observation would provide a useful generic insight into efficiency of resampling with regard to feature learning. 

In this paper, we provide a theoretical analysis on a toy model to examine the effect of resampling and reweighting, especially aiming to clarify under what conditions the observation by Kang et al.\ holds. In our toy model, we treat a binary classification problem, in which the sample-generation process is assumed to be stochastic. More specifically, inputs are independent and identically-distributed (i.i.d.) from a probability distribution on $\mathbb{R}^N$, with parameters controlling the class imbalance and the variances. The two class centers are assumed to be located at $\pm\bm{w}_0/\sqrt{N}\in\mathbb{R}^N$.
These constitute a standard setting for theoretical treatments of binary classification~\citep{barkai1994statistical}.

In the above setting, the performance of feature learning can be quantified as the accuracy of estimating $\V{w}_0$, since it represents the most discriminative direction of the two classes
under isotropic class-conditional input distributions. Our analysis, which considers the asymptotic $N\to\infty$, reveals that the accuracy of estimating $\V{w}_0$ is maximized when one does not employ resampling/reweighting at all, irrespectively of the degree of the class imbalance, under the conditions that the variances of samples of the two classes are equal, that the decision boundary is located equidistantly from the two class centers, and that a specific ansatz of the analytical framework we use, the so-called replica symmetric (RS) ansatz, is correct.  This finding provides an analytical support for Kang et al.'s observation.  More interestingly, this finding remains valid with a rather wide range of classifiers and losses, which can be shown on the basis of the symmetry in formulae derived in our analysis. Although the equal-variance condition for sample distributions might seem somewhat artificial, it may be achieved in the feature representation at the last hidden layer of a classification DNN as a result of DNN training, and some recent studies partially support this~\citep{doi:10.1073/pnas.2015509117,doi:10.1073/pnas.2103091118}.

Our results are derived via statistical-mechanical techniques~\citep{barkai1994statistical,barkai1993scaling,biehl1993statistical,watkin1994optimal,lootens1995analysing,tanaka2013statistical}, which are applicable in the limit $N \to \infty$. Especially, the replica method~\citep{nishimori2001statistical,dotsenko2005introduction,mezard2009information} plays a key role in computing the quantities of interest. Although the replica method is mathematically non-rigorous, the results derived in this paper are conjectured to be correct, which is supported by an excellent agreement with results of numerical experiments, as well as by an accumulation of many model analyses over many decades in which the replica method is eventually shown to give the exact results~\citep{montanari2006analysis,talagrand2003spin,talagrand2011mean1,talagrand2011mean2}. 

The remainder of this paper is organized as follows. In the next section, the problem setup and the formulation are explained. In \Rsec{Theoretical}, the analysis details using the replica method are explained. The derived formulae using the replica method are utilized to systematically examine behaviors of the quantities of interest. The result shows that there exists a case in which the absence of resampling/reweighing gives the best feature learning performance irrespectively of the choice of losses or classifiers, yielding a theoretical support for Kang et al.'s observation. On the basis of this theoretical result, we also provide a further simplified model for multiclass classification, for which the same consequence about the resampling/reweighting holds. In \Rsec{Numerical}, numerical experiments are conducted to verify the replica results. The last section concludes the paper.

%%%%%%%%%%%%%%%%%%%%%%%%%%%%%%%%%%%%%%%%%%%%%%%%%%%%%
%%%%%%%%%%%%%%%%%%%%%%%%%%%%%%%%%%%%%%%%%%%%%%%%%%%%%
%%%%%%%%%%%%%%%%%%%%%%%%%%%%%%%%%%%%%%%%%%%%%%%%%%%%%
\section{Problem setup, formulation, and related work}\Lsec{Problem}
%Here we explain our problem setup and formulation.

%%%%%%%%%%%%%%%%%%%%%%%%%%%%%%%%%%%%%%%%%%%%%%%%%%%%%
%%%%%%%%%%%%%%%%%%%%%%%%%%%%%%%%%%%%%%%%%%%%%%%%%%%%%
\subsection{Data-generation process}\Lsec{Data generation}
Let us consider a classification problem with two classes labeled by $y=\pm 1$ whose distribution is
\be
P_Y(y)=\sum_{y'=\pm1}r_{y'}\delta_{y,y'},
\ee
where $\delta_{a,b}$ denotes the Kronecker delta,
which equals 1 when $a=b$ and 0 otherwise, 
and where $r_{\pm1}\in \lsb 0,1 \rsb$ with $r_{+1}+r_{-1}=1$ control the degree of class imbalance: $r_{+1}=r_{-1}=1/2$ corresponds to the balanced case. The input space is assumed to be $\mR^{N}$ and the input vector $\V{x}\in\mR^N$ is assumed to be generated from the following model given a label $y\in\{-1,1\}$:
\be
\V{x}=y\frac{\V{w}_0}{\sqrt{N}}+\V{\xi}(y),
\Leq{model}
\ee
where $\pm \V{w}_{0}/\sqrt{N}\in \mR^{N}$ represent the class centers corresponding to the two classes $y=\pm1$. We impose the normalization condition $\|\V{w}_0\|^2/N=1$. In \Req{model}, $\V{\xi}(y)$ is assumed to be a random i.i.d.\ vector obeying the zero-mean Gaussian distribution 
\be
P_{\V{\Xi}\mid Y}\lb  \V{\xi} \remid{|}  \sigma^2_{y}\rb
=
\lb 2\pi \sigma_{y}^2 \rb^{-N/2}e^{-\frac{1}{2\sigma^2_{y}}\left\|  \V{\xi} \right\|^2}, 
\Leq{Gaussian}
\ee
where the label-$y$ dependence appears only through the variance $\sigma^2_{y}$ which is assumed finite for both $y=\pm1$. The variance $\sigma^2_y$ expresses the cluster size of the class $y$ in the input space. We call this model the Gaussian class-conditional model (GCCM) (see, e.g.,~\cite{ChatterjiLong2021})\footnote{The same model is sometimes referred to by other names, such as the Gaussian mixture model.}. Let $D^M=\{(\V{x}_{\mu},y_{\mu})\}_{\mu=1}^{M}$ be a dataset of $M$ i.i.d.\ datapoints following the above data-generation process: 
\be
\Leq{likelihood}
P(D^M \mid \V{w}_0)
=
\prod_{\mu=1}^{M}r_{y_\mu}%P_{Y}(y_{\mu})
P_{\V{\Xi}\mid Y}\lb   \V{x}_{\mu}-y_{\mu}\frac{\V{w}_0}{\sqrt{N}} \remid{|}  \sigma^2_{y_{\mu}}\rb.
\ee
Although we derive our results on the basis of the Gaussianity assumption \NReq{Gaussian}, we expect that the same results hold for a much wider class of distributions with $\sigma_y^2<\infty$ thanks to the universality appearing through the central limit theorem in the limit $N\to\infty$. Also, it is possible that qualitatively similar results may hold for a broader class of distributions beyond the Gaussian universality, such as those studied in~\citet{adomaityte2024classification}.

%In \Req{model}, $\V{\xi}(y)$ is a random i.i.d.\ vector following the distribution $P_{\V{\Xi}\mid Y}(\V{\xi}\mid y)=\prod_{i=1}^{N}P_{\Xi\mid Y}(\xi_i\mid y)$, where the components of $\V{\xi}$ are also assumed to be i.i.d.: we further assume that $P_{\Xi\mid Y}(\xi\mid y)$ is zero-mean and that the label-$y$ dependence appears only through the variance $\sigma^2_{y}$ which is finite for both $y=\pm1$; to make this explicit, we will also write $P_{\V{\Xi}\mid Y}(\V{\xi}\mid y)=P_{\V{\Xi}\mid Y}(\V{\xi}\mid \sigma^2_y)$. The variance $\sigma^2_y$ in a sense expresses the cluster size of the class $y$ in the input space. 

%%%%%%%%%%%%%%%%%%%%%%%%%%%%%%%%%%%%%%%%%%%%%%%%%%%%%
%%%%%%%%%%%%%%%%%%%%%%%%%%%%%%%%%%%%%%%%%%%%%%%%%%%%%
\subsection{Classifier and loss}\Lsec{Classifier and}
A generic classifier can be formulated as first mapping the input $\bm{x}$ onto a one-dimensional feature $z=f(\bm{x})\in\mathbb{R}$ via a feature mapping $f$, and then, on the basis of the feature $z=f(\bm{x})$,
producing a soft decision $P(y\mid \bm{x})$ which is an estimate of the conditional distribution of the class label $y$ for the input $\bm{x}$. In our formulation, we consider the simplest case where a linear feature mapping $f(\bm{x})=\bm{w}^\top\bm{x}/\sqrt{N}$ is to be used, with $\bm{w}\in\mathbb{R}^N$ its parameter. As the soft-decision classifier given the one-dimensional feature $z=f(\bm{x})$, which we call the feature-based classifier, we assume $P(y\mid \bm{x})=\mathcal{M}(y(f(\bm{x})+b))$ with a function $\mathcal{M}:\mathbb{R}\to[0,1]$. The bias term $b$ is the parameter of the feature-based classifier. We may also write $P(y\mid \V{x};\V{w},b)=\out(y(\V{w}^{\top}\V{x}/\sqrt{N}+b))$ in order to make explicit the dependence of the classifier on the parameters $\bm{w}$ and $b$.  By an appropriate choice of the function $\mathcal{M}$, this formulation covers several standard classifier models,
such as a perceptron and a logistic function:
%we can treat some classifiers belonging to generalized linear models in a unified manner: the probability of $y$ given a feature point $\V{x}$ is denoted as $P(y\mid \V{x};\V{w},b)=\out(\V{w}^{\top}\V{x}/\sqrt{N}+b,y)$ where the model parameters $\V{w}$ and $b$ are termed weights and bias. Some standard models are perceptron and logit functions:
\be
&&
\out_{\rm pe}(h)\coloneqq %\delta_{\sgn{h},1},
\frac{1+\sgn{h}}{2}=\begin{cases}
  1,&h>0,\\
  1/2,&h=0,\\
  0,&h<0,
\end{cases}
%\\ &&
%\out_{\rm pr}(h,y)\coloneqq \frac{1+y}{2}\Phi(h)+\frac{1-y}{2}(1-\Phi(h)),
\\ &&
\out_{\rm lo}(h)\coloneqq \frac{e^{h}}{2\cosh(h)}.
\ee 
%where $\Phi(h)$ denotes the cumulative distribution function of the standard normal distribution. 
%A well-known perceptron can also be derived from a limit of these models: $\out_{\rm perceptron}(h,y)=\lim_{\beta \to \infty} \out_{\rm probit}(\beta h,y)=\lim_{\beta \to \infty} \out_{\rm logit}(\beta h,y)$. 
%These models have a property $\out(-h,-y)=\out(h,y)$ reflecting the symmetry between the two classes. In the following, we use $\out$ to denote general model having the above parameterization and symmetry.
We say that a feature-based classifier using $\mathcal{M}(h)$ is symmetric
if the function $\mathcal{M}$ satisfies
$\mathcal{M}(h)+\mathcal{M}(-h)=1$ for all $h\in\mathbb{R}$.
$\out_{\rm pe}$ and $\out_{\rm lo}$ shown above are examples
defining symmetric feature-based classifiers. 

The weights $\V{w}$ of the feature mapping $f$ are usually determined by minimizing an empirical loss, which is a (sometimes weighted) sum of loss values over all the datapoints. We let $\loss(\V{w};\V{x},y,\out)$ denote the loss of the weights $\V{w}$ given the datapoint $(\V{x},y)$ and the model $\out$. According to the above assumptions of the linear feature mapping and those on the model, generic loss function can be written as a function $\lossS(h,y)$ of two arguments $h=\V{w}^{\top}\V{x}/\sqrt{N}+b$ and $y$, in such a way that the relation $\loss(\V{w};\V{x},y,\out)=\lossS(\V{w}^{\top}\V{x}/\sqrt{N}+b,y)$ holds. A common loss function in the recent practice is cross entropy (CE) and the corresponding loss function takes the form
\be
\lossS_{\rm CE}(\V{w}^{\top}\V{x}/\sqrt{N}+b,y)= -\log \out(y(\V{w}^{\top}\V{x}/\sqrt{N}+b)).
\ee
As $\lossS_{\mathrm{CE}}(h,y)$ depends on $h$ and $y$ only through their
product $yh$, it has the symmetry $\lossS_{\rm CE}(-h,-y)=\lossS_{\rm CE}(h,y)$;
it is also the case with many other standard losses which are represented as functions of $yh$, such as zero-one, exponential, smoothed or non-smoothed hinge losses. Hereafter the symbol $\lossS$ is used to denote a generic loss having this symmetry. To express a specific model and loss, we use an appropriate subscript: for example, if we use the logistic function and the CE, the resultant loss will be denoted as $\lossS_{\rm CElo}$. Furthermore, if $\lossS$ is used with a single argument, we assume that it expresses the one for the positive label: $\lossS(h)=\lossS(h,+1)$. This is a convenient shorthand notation when we work with the above symmetry. 

The empirical loss considered in this paper has class-wise reweighing factors $s_y\in[0,1]$, which are assumed to satisfy the condition $s_{+1}+s_{-1}=1$. %, where $s_+(s_-)$ corresponds to $s_y$ with $y=+1(-1)$.
Given a loss $\lossS$ and a dataset $D^M=\{(\V{x}_{\mu},y_{\mu})\}_{\mu=1}^{M}$, the empirical reweighted loss is thus written by
\be
\mc{H}(\V{w}\mid D^M;b,s)=\sum_{\mu=1}^{M}s_{y_{\mu}}\lossS(\V{w}^{\top}\V{x}_{\mu}/\sqrt{N}+b,y_{\mu}).
\Leq{Hamiltonian}
\ee
This class-wise reweighting of the loss is intended to mitigate possible undesirable effects of the class imbalance arising when $r_{\pm1}\not=1/2$. It can be considered as a special case of the sample-wise weighting which has been investigated, e.g., in~\citet{he2009learning}. We would also like to mention that resampling offers yet another alternative for the purpose of mitigation of class imbalance. Resampling, however, amounts to determining the sample-wise weights in the empirical loss by the respective sample counts, %resampling offers an alternative means, but also only changes the weight of each data point in the total loss,
and hence the average effect of the resampling can be incorporated into the class-wise reweighting factors. We analyze properties of the above loss and its minimizer $\hat{\V{w}}=\argmin_{\V{w}:\|\V{w}\|^2=N}\mc{H}(\V{w}\mid D^M;b,s)$ under the constraint $\|\V{w}\|^2=N$, especially focusing on the overlap between $\hat{\V{w}}$ and $\V{w}_0$: $\V{w}_0$ defines the most discriminative direction of the two classes, and under the fixed-norm constraint $\|\hat{\V{w}}\|^2=N$ how well the estimate $\hat{\V{w}}$ aligns with the direction of $\V{w}_0$ is fully captured by the overlap $m=\hat{\V{w}}^{\top}\V{w}_0/N$, implying that the quality of feature learning in our setting is solely characterized by $m$.
Although estimating the bias $b$ may also be performed via minimization of $\mc{H}$ with respect to $b$, the analytical framework explained in the next section allows us to do it in a more flexible manner and we leave $b$ as a tunable parameter. 
 
Before proceeding, we have a noteworthy remark about the Bayesian inference. Suppose that we know the data-generation process but do not know the specific values of either $\V{w}_0$ or $r_{\pm1}$. We thus introduce $\V{w}$ as a random variable to estimate $\V{w}_0$ and $r_{\pm1}$ as hyperparameters playing the role of class-wise reweighting factors. With an appropriate prior $P(\V{w})$, the posterior distribution of $\V{w}$ given the dataset $D^M$ becomes 
\be
&&
P(\V{w}\mid D^M)
=
\frac{
P(\V{w})\prod_{\mu=1}^{M}
r_{y_{\mu}}%(r\delta_{y_{\mu},1}+(1-r)\delta_{y_{\mu},-1})
P_{\V{\Xi}\mid Y}\lb   \V{x}_{\mu}-y_{\mu}\frac{\V{w}}{\sqrt{N}}  \remid{|} \sigma_{y_{\mu}}^2 \rb 
}{
\int d\V{w}
P(\V{w})\prod_{\mu=1}^{M}
r_{y_\mu}%(r\delta_{y_{\mu},1}+(1-r)\delta_{y_{\mu},-1})
P_{\V{\Xi}\mid Y}\lb   \V{x}_{\mu}-y_{\mu}\frac{\V{w}}{\sqrt{N}}  \remid{|} \sigma_{y_{\mu}}^2 \rb 
}
\no 
\\ && 
=
\frac{
P(\V{w})
\prod_{\mu=1}^{M}
P_{\V{\Xi}\mid Y}\lb   \V{x}_{\mu}-y_{\mu}\frac{\V{w}}{\sqrt{N}}  \remid{|} \sigma_{y_{\mu}}^2 \rb 
}{
\int d\V{w}
P(\V{w})
\prod_{\mu=1}^{M}
P_{\V{\Xi}\mid Y}\lb   \V{x}_{\mu}-y_{\mu}\frac{\V{w}}{\sqrt{N}} \remid{|} \sigma_{y_{\mu}}^2 \rb 
}.
\ee
Hence the posterior distribution does not depend on $r_{\pm1}$, meaning that the reweighting via adjusting $r_{\pm1}$ has no effect on estimation of $\V{w}_0$. If we treat $r_{\pm1}$ as random variables with a prior $P(\{r_{\pm1}\})$, the result is seemingly different but the marginal posterior $P(\V{w}\mid D^M)=\int dr~P(\V{w},r\mid D^M)$ is still independent of the choice of $P(\{r_{\pm1}\})$ as long as the priors of $r_{\pm1}$ and $\V{w}$ are independent. Hence, to study the effect of reweighting/resampling on the feature learning, the Bayesian inference framework based on the true data-generation process is inappropriate. 

%%%%%%%%%%%%%%%%%%%%%%%%%%%%%%%%%%%%%%%%%%%%%%%%%%%%%
%%%%%%%%%%%%%%%%%%%%%%%%%%%%%%%%%%%%%%%%%%%%%%%%%%%%%
\subsection{Statistical mechanical formulation}\Lsec{Statistical mechanical}
To investigate the estimator $\hat{\V{w}}=\argmin_{\V{w}:\|\V{w}\|^2=N}\mc{H}(\V{w}\mid D^M;b,s)$, it is convenient to introduce the following distribution:
\be
&&
P_{\beta}(\V{w}\mid D^M;b,s)\coloneqq \frac{1}{Z}\delta(N-\|\V{w}\|^2)e^{-\beta \mc{H}(\V{w}\mid D^M;b,s)},
\Leq{Boltzmann}
\ee
where $\delta(\cdot)$ denotes the Dirac measure, where $\beta\ge0$ is the inverse temperature parameter, and where
\be
Z=Z(D^M;b,s)\coloneqq\int d\V{w}~\delta(N-\|\V{w}\|^2)e^{-\beta \mc{H}(\V{w}\mid D^M;b,s)},
\Leq{Partition}
\ee
is the normalization coefficient. In the limit $\beta\to \infty$, the distribution $P_\beta$ concentrates on the set of minimizers of $\mc{H}(\V{w}\mid D^M;b,s)$, and hence any properties of the estimator $\hat{\V{w}}$ can be computed from the average over the distribution in the limit. Further, $\fe= -(\beta N)^{-1}\log Z$ plays the role of the cumulant generating function of $\V{w}$ and converges to the per-variable average loss in the limit $\beta \to \infty$, that is, $\lim_{\beta \to \infty}\fe=u\coloneqq \min_{\V{w}:\|\V{w}\|^2=N}\mc{H}(\V{w}\mid D^M;b,s)/N$. This means that $\fe$ contains all the necessary information for our purpose and hereafter we concentrate on computing it. According to the physics terminology, in the following we call $P_{\beta}$ the Boltzmann distribution, $\beta^{-1}$ the temperature, $Z$ the partition function, $\fe$ the free energy, and $u$ the energy. The average over the Boltzmann distribution is denoted by the angular brackets as
\be
\Ave{(\cdots)}
=\int d\V{w}\,P_{\beta}(\V{w}\mid D^M;b,s)(\cdots)
=\Tr{\V{w}}\frac{e^{-\beta \mc{H}(\V{w}\mid D^M;b,s)}}{Z}(\cdots),
\Leq{BAve}
\ee
where the symbol $\Tr{\V{w}}=\int d\V{w}\,\delta(N-\|\V{w}\|^2)$ is introduced for notational simplicity of our development later. 

A problem arises in the computation of the free energy $\fe$: it depends on the random variable $D^M$ and hence its direct evaluation is difficult. However, $\fe$ is expected to exhibit what is called the self-averaging property, implying that it converges to its expectation value over $D^M$ in the limit $N\to \infty$. Hence we may instead compute $\lsb \fe \rsb_{D^M}$, where the square brackets express the average over the data-generation process:
\be
&&
\lsb (\cdots) \rsb_{D^M}
=\lb \prod_{\mu=1}^{M}\sum_{y_{\mu}=\pm 1}\int d\V{x}_{\mu}\rb
P(D^M \mid \V{w}_0)
(\cdots),
%\\ &&
%P(D^M \mid \V{w}_0)
%=\prod_{\mu=1}^{M}
%r_{y_\mu}%(r_+\delta_{y_{\mu},1}+(1-r_+)\delta_{y_{\mu},-1})
%P_{\V{\Xi}\mid Y}\lb   \V{x}_{\mu}-y_{\mu}\frac{\V{w}_0}{\sqrt{N}}  \remid{|} \sigma_{y_{\mu}}^2 \rb.
\ee
where $P(D^M \mid \V{w}_0)$ is given in~\Req{likelihood}. 
Unfortunately, the evaluation of $\lsb \fe \rsb_{D^M}=-(\beta N)^{-1}\lsb \log Z \rsb_{D^M}$ is still difficult. The replica method is a great aid in such a situation, via making use of the following identity:   
\be
\lsb \log Z \rsb_{D^M}
=
\lim_{n\to 0}\frac{1}{n}\log \lsb Z^n \rsb_{D^M}.
\Leq{replica trick}
\ee
In addition to this identity, we assume that $n$ is a positive integer. This assumption enables us to explicitly compute the average $\lsb (\cdots) \rsb_{D^M}$ on the right-hand side of \Req{replica trick}. After computing this average, we take the limit $n\to 0$ by relying on an expression of the average that is analytically continuable from $\mathbb{N}$ to $\mathbb{R}$, under what is called the RS ansatz. The details are in sec.~\ref{sec:Theoretical}.

%%%%%%%%%%%%%%%%%%%%%%%%%%%%%%%%%%%%%%%%%%%%%%%%%%%%%
%%%%%%%%%%%%%%%%%%%%%%%%%%%%%%%%%%%%%%%%%%%%%%%%%%%%%
\subsection{Related work}
The statistical mechanical approach to neural networks and machine learning problems has a long history: some pioneering pieces of work are for associative memory~\citep{doi:10.1073/pnas.79.8.2554}, generative model~\citep{hinton_BM}, simple perceptrons~\citep{Gardner1988,GardnerDerrida1988}, supervised learning~\citep{PhysRevA.45.6056}, and unsupervised learning~\citep{barkai1994statistical}. These facilitated various related studies, leading a research area of statistical mechanics for information processing~\citep{nishimori2001statistical,mezard2009information}. This approach has gained renewed interests due to the recent upsurge in the field of machine learning, and some interesting findings have been accumulating. In this subsection, we review some contributions along this direction, highlighting their relevance to and distinction from our study.

  \citet{10.1214/009117905000000233} investigated behaviors of large eigenvalues of what are called the random non-central Wishart matrices in the infinite-dimensional limit and identified critical thresholds, which have implications for inference and estimation in high-dimensional settings. Although focusing on random matrix theory, their work shares similarities with ours in terms of utilizing infinite-dimensional analysis techniques. \citet{Lesieur_2017} applied tools from statistical physics to study low-rank matrix estimation, emphasizing phase transitions and the performance of approximate message passing algorithms. Although their use of the replica method in assessing the estimation performance is conceptually related to our approach, their focus was on matrix factorization and Bayes-optimal inference, which differs from our exploration of resampling and reweighting for imbalanced classification tasks.

  \citet{mignacco2020role} provided an analysis of the classification performance of the empirical risk minimization (ERM) with generic convex loss functions and the $\ell_2$ regularization when the data are generated from the two-class GCCMs as ours. Their setup is quite similar to ours, but they focused on the effects of regularization on classification performance, in contrast to our current aim to clarify the effect of resampling and reweighting on feature learning. A noteworthy point of their work is that it offers a mathematically rigorous proof for the formula derived using the replica method. In order to enable the rigorous proof, they restricted their loss function to convex ones, while the replica method can handle even non-convex loss functions and in fact we discuss general loss functions that may not necessarily be convex. \citet{loureiro2021learning} studied behaviors of generalized linear models (GLMs) when applied to data generated from the multiclass GCCMs in the ERM framework. Their analysis provided precise asymptotic results for training and generalization errors in high dimensions, and they examined how regularization affects performance. They also offered a proof of their formula, provided that both the loss and the regularizer are convex. One interesting point of this paper is that it pointed out that when the training dataset is generated by a Generative Adversarial Network (GAN), the corresponding learning curve shows a fairly good agreement with the theoretical curve derived assuming a GCCM as the data-generating distribution, implying a potential relevance to real data. However, this work did not deal with the problem of feature learning for imbalanced classification focused in this paper. \citet{loffredo2024restoring} tackled the problem of finding the optimal undersampling and oversampling strategies in imbalanced classification. To this end, they computed several quantities characterizing the classification performance in a systematic way using the replica method. Their problem setup is directly relevant to ours, as they analyzed the effects of under/oversampling on classification performance. However, their focus was primarily on optimizing sampling strategies for accuracy, whereas our study explores how resampling and reweighting affect the quality of feature learning.

\citet{Takahashi2022} investigated use of pseudo-labels in self-training for semi-supervised learning using the two-class GCCMs. Similarly to our work, this study employed the replica method to derive sharp asymptotic characterizations, focusing on iterative updates of model parameters and proposing heuristics for pseudo-label refinement that yield performance close to that of fully supervised learning. In another study \citep{Takahashi2024}, the same author provided a replica analysis of under-bagging (UB) for imbalanced classification, again using the two-class GCCMs, comparing it with undersampling and simple weighting methods. This study revealed similarity and difference among these three methods and established superiority of UB in terms of performance and simplicity of implementation. While these pieces of work examined the impact of label imbalance in classification performance and used the replica method as our current work, our study uniquely focuses on interaction between resampling/reweighting strategies and feature learning performance under class imbalance.

\citet{adomaityte2024classification} analyzed the impact of non-Gaussian, heavy-tailed data distributions on classification performance using the replica method. As a result, they demonstrated deviations from the Gaussian case, indicating the importance of considering the non-Gaussianity and heavy-tailedness for capturing real data in theory. This is an important contribution, but their emphasis on heavy-tailed distributions contrasts with our focus on the class imbalance. \citet{10.5555/3618408.3618821} investigated behaviors of gradient-based learning algorithms under the class imbalance, to find that the imbalance affects the convergence of the dynamics.  To mitigate this, they also introduced techniques of normalizing the gradient in a class-wise manner. While this study focused on class imbalance, it emphasized optimization dynamics and is rather different from our aim. Finally, the work by \citet{mannelli2024bias} explored the geometry of data and clarified how it introduces bias in machine learning models, via applying the replica method to a synthetic data-generation model called the teacher-mixture model. While they also examined the effect of the imbalance, their focus was on fairness and bias, and hence it does not align with the objective of the present paper. We investigate when the resampling and reweighting strategies enhance feature learning through a detailed analysis using the replica method.

%%%%%%%%%%%%%%%%%%%%%%%%%%%%%%%%%%%%%%%%%%%%%%%%%%%%%
%%%%%%%%%%%%%%%%%%%%%%%%%%%%%%%%%%%%%%%%%%%%%%%%%%%%%
%%%%%%%%%%%%%%%%%%%%%%%%%%%%%%%%%%%%%%%%%%%%%%%%%%%%%
\section{Theoretical Analysis}\Lsec{Theoretical}
\subsection{Overview}
In this section we derive a formula for the sample-averaged free energy $[\phi]_{D^M}=-(\beta N)^{-1}[\log Z]_{D^M}$ in the limit $N\to\infty$ by using the replica method under the RS ansatz. The formula is characterized by a small number of quantities which are called {\it order parameters}, and the order parameters satisfy a set of equations called {\it equations of state (EOS)}, both according to the physics terminology. The dependence of the order parameters on the parameters, especially on the reweighting factor $s_{\pm1}$, is of our special interest in the paper and is systematically studied on the basis of the EOS. A further simplified model inspired from the replica results will also be introduced later in this section to discuss the case with more than two classes.

For notational simplicity, we use the shorthand notation $s_+(s_-)$ to denote $s_{+1}(s_{-1})$ hereafter. The same shorthand rule applies to $\sigma_{y}$ and $r_{y}$ as well.

%%%%%%%%%%%%%%%%%%%%%%%%%%%%%%%%%%%%%%%%%%%%%%%%%%%%%
%%%%%%%%%%%%%%%%%%%%%%%%%%%%%%%%%%%%%%%%%%%%%%%%%%%%%
\subsection{Derivation of free energy and EOS under RS ansatz}\Lsec{Derivation of}
The computation starts from evaluating $\lsb Z^n \rsb_{D^M}$. If $n\in \mN$, %we have
we may consider $n$ ``replicas'' of the original system
which have distinct parameters $\{\V{w}_a\}_{a=1}^n$ and share the same dataset $D^M$,
and represent $\lsb Z^n\rsb_{D^M}$ using them as
\be
\lsb Z^n \rsb_{D^M}=
\lsb
\Tr{\{\V{w}_a\}_{a=1}^n}~
e^{
-\beta \sum_{\mu=1}^{M}\sum_{a=1}^{n}s_{y_{\mu}}\lossS(\V{w}_a^{\top}\V{x}_{\mu}/\sqrt{N}+b,y_{\mu})
}
\rsb_{D^M},
\Leq{Z^n}
\ee
where $\Tr{\{\V{w}_a\}_{a=1}^n}=\prod_{a=1}^{n}\Tr{\V{w}_a}$. The average over the dataset $D^M$ yields
\begin{equation}
  \label{eq:avgD}
  \lsb
\Tr{\{\V{w}_a\}_{a=1}^n}~
e^{
-\beta \sum_{\mu=1}^{M}\sum_{a=1}^{n}s_{y_{\mu}}\lossS(\V{w}_a^{\top}\V{x}_{\mu}/\sqrt{N}+b,y_{\mu})
}
\rsb_{D^M}
=\Tr{\{\V{w}_a\}_{a=1}^n} L^M,
\end{equation}
where we let
\begin{equation}
 L:=\sum_{y=\pm1}
r_y
\int d\V{x}\,
P_{\V{\Xi}\mid Y}\lb    \V{x}-y\frac{\V{w}_0}{\sqrt{N}} \remid{|}  \sigma^2_{y}\rb
e^{
-\beta \sum_{a=1}^{n}s_{y}\lossS(\V{w}_a^{\top}\V{x}/\sqrt{N}+b,y)
}.\Leq{ave1}
\end{equation}
 \if0
\be
&&
\lsb
\Tr{\{\V{w}_a\}_{a=1}^n}~
e^{
-\beta \sum_{\mu=1}^{M}\sum_{a=1}^{n}s_{y_{\mu}}\lossS(\V{w}_a^{\top}\V{x}_{\mu}/\sqrt{N}+b,y_{\mu})
}
\rsb_{D^M}
\no \\ &&
=
\Tr{\{\V{w}_a\}_{a=1}^n}~
\Biggl\{\sum_{y=\pm1}
r_y
\int d\V{x} 
P_{\V{\Xi}\mid Y}\lb    \V{x}-y\frac{\V{w}_0}{\sqrt{N}} \mid  \sigma^2_{y}\rb
e^{
-\beta \sum_{a=1}^{n}s_{y}\lossS(\V{w}_a^{\top}\V{x}/\sqrt{N}+b,y)
}
%\no \\ &&
%+
%(1-r_+)
%\int d\V{x} 
%P_{\V{\Xi}\mid Y}\lb    \V{x}+\frac{\V{w}_0}{\sqrt{N}} \mid  \sigma^2_{-}\rb
%e^{
%-\beta \sum_{a=1}^{n}s_{-}\lossS(\V{w}_a^{\top}\V{x}/\sqrt{N}+b,-1)
%}
\Biggr\}^M
\eqqcolon
\Tr{\{\V{w}_a\}_{a=1}^n}~
\Biggl\{
L
\Biggr\}^M.
\Leq{ave1}
\ee
\fi
The integral over $\V{x}$ in Eq.~\eqref{eq:ave1} is cumbersome. However, the integrand depends on $\V{x}$ only through the quantities $\left\{u_a=\V{w}_a^{\top}\V{x}/\sqrt{N}\right\}_{a=1}^n$.  Conditional on $y$, $\{u_a\}_{a=1}^n$ obeys a multivariate Gaussian, thanks to the Gaussianity assumption \NReq{Gaussian} on $\bm{\xi}(y)$ in the present setup or thanks to the central limit theorem in the large-$N$ limit in a more generic case.  Letting $\E \lsb \cdot \mid y \rsb$ denote the average over $\V{\xi}(y)$ given $y$, the conditional mean of $u_a$ given $y$ is 
\be
\E \lsb u_a \mid y \rsb
=\frac{\V{w}_a^\top\E \lsb \V{x} \mid y \rsb}{\sqrt{N}}
=ym_a,
%\left\{
%\begin{array}{cc}
%\V{w}^{\top}_0\V{w}_a/N  &  ({\rm 1st~term})   \\
%-\V{w}^{\top}_0\V{w}_a/N  &  ({\rm 2nd~term})    
%\end{array}
%\right.
%\eqqcolon
%\left\{
%\begin{array}{cc}
%m _a &  ({\rm 1st~term})   \\
%-m_a  &  ({\rm 2nd~term})    
%\end{array}
%\right..
\ee
where we let $m_a:=\V{w}_a^\top\V{w}_0/N$. 
The conditional covariance of $u_a$ and $u_b$ given $y$ is 
\be
&&
\E\lsb u_au_b\mid y\rsb-\E\lsb u_a\mid y\rsb\E\lsb u_b\mid y\rsb
=
Q_{ab}\sigma_y^2,
%\left\{
%\begin{array}{cc}
%\sigma_+^2  &  ({\rm 1st~term})   \\
%\sigma_-^2  &  ({\rm 2nd~term})    
%\end{array}
%\right.,
\ee
where
we let $Q_{ab}:=\V{w}_a^\top\V{w}_b/N$. 
%\be
%Q_{ab}=\frac{1}{N}\sum_{i=1}^{N}w_{ai}w_{bj}.
%\ee

To proceed further, we assume the RS as follows:
\be
m_a=m,\quad Q_{ab}=Q\delta_{ab}+q(1-\delta_{ab}).
\ee
This assumption allows us to express $u_a$ given $y$ as 
\be
u_a=\sigma_y\lb \sqrt{Q-q}t_a+\sqrt{q}z \rb+ym,
%\left\{
%\begin{array}{cc}
%\sigma_+\lb \sqrt{Q-q}t_a+\sqrt{q}z \rb+ m &  ({\rm 1st~term})   \\
%\sigma_-\lb \sqrt{Q-q}t_a+\sqrt{q}z \rb-m  &  ({\rm 2nd~term})    
%\end{array}
%\right.,
\ee
where $t_a$ and $z$ are independent standard Gaussian random variables. %with zero mean and unit variance.
Hence the integral over $\V{x}$ is recast into those over $z,\{t_a\}_{a=1}^{n}$. Using the shorthand notation
\be
\int_{-\infty}^{\infty} \frac{dz}{\sqrt{2\pi}}\,e^{-\frac{1}{2}z^2}
(\cdots)\eqqcolon \int Dz\,(\cdots),
\ee
we have
\begin{align}
  L&=\sum_{y=\pm1}r_y\int Dz\lb\int Dt\,
  e^{
-\beta 
s_y\lossS \lb \sigma_y \lb \sqrt{Q-q}t+\sqrt{q} z\rb + ym+b,y\rb
}
  \rb^n\nonumber\\
  &=\sum_{y=\pm1}r_y\int Dz\lb\int Dt\,
  e^{
-\beta 
s_y\lossS \lb h(t,z,Q,q,m,\sigma_y,yb)\rb
}
  \rb^n,\Leq{L}
\end{align}
%\be
%&&
%\hspace{-1.5cm}
%L=
%r_+
%\int Dz \lb \int Dt
%e^{
%-\beta 
%s_+\lossS \lb \sigma_+ \lb \sqrt{Q-q}t+\sqrt{q} z\rb + m+b,+1\rb
%}
%\rb^n
%\no \\ &&
%\hspace{-1.5cm}
%+
%(1-r_+)
%\int Dz \lb \int Dt 
%e^{
%-\beta 
%s_-\lossS \lb \sigma_- \lb \sqrt{Q-q}t+ \sqrt{q} z \rb - m+b,-1\rb
%}
%\rb^n
%\no \\ &&
%\hspace{-1.5cm}
%=
%r_+
%\int Dz \lb \int Dt
%e^{
%-\beta 
%s_+\lossS \lb h(t,z,Q,q,m,\sigma_{+},b)\rb
%}
%\rb^n
%+
%(1-r_+)
%\int Dz \lb \int Dt 
%e^{
%-\beta 
%s_-\lossS \lb h(t,z,Q,q,m,\sigma_{-},-b)\rb
%}
%\rb^n.
%\Leq{L}
%\ee
where we let
\be
h(t,z,Q,q,m,\sigma,b)=\sigma \lb \sqrt{Q-q}t+\sqrt{q}z \rb + m+b,
\ee
and in the last line of \Req{L} we used the invariance of the result with respect to (w.r.t.) $z\to -z,~t\to -t$ and the symmetry of the loss with the single-argument shorthand notation $\lossS(yh)=\lossS(h,y)$ introduced in \Rsec{Classifier and}.
Equation~\eqref{eq:L} reveals that $L$ depends on $\{\V{w}_a\}_{a=1}^n$
only through $Q=Q(\{\V{w}_a\}_{a=1}^n)$, $q=q(\{\V{w}_a\}_{a=1}^n)$,
and $m=m(\{\V{w}_a\}_{a=1}^n)$.
Hence the integral $\Tr{\{\V{w}_a\}_{a=1}^n}$ in Eq.~\eqref{eq:avgD} 
can be rewritten as that w.r.t.\ $Q,q,m$ as 
\be
\lsb Z^n \rsb_{D^M}=\Tr{\{\V{w}_a\}_{a=1}^n}L^M%(Q(\{\V{w}_a\}_{a=1}^n),q(\{\V{w}_a\}_{a=1}^n),m(\{\V{w}_a\}_{a=1}^n))
=\int dQ\,dq\,dm\,V(Q,q,m)L^M(Q,q,m),
\ee
where we introduced the notation $L(Q,q,m)$ to denote the dependence of $L$
on $Q,q,m$, and 
\if0
The dependence on $\{\V{w}_a\}_{a=1}^{n}$ only appears through $Q,q,m$. This naturally leads to the following transformation
\be
\lsb Z^n \rsb_{D^M}=\int dQ dq dm V(Q,q,m)L^M(Q,q,m),
\ee
\fi
where $V(Q,q,m)$ represents the volume of the subshell specified by $Q,q,m$ in the space of $\{\V{w}_a\}_{a=1}^{n}$:
\be
V(Q,q,m)\coloneqq
\Tr{\{\V{w}_a\}_{a=1}^n}
\lb 
\prod_{a=1}^{n}
\delta\lb NQ-\|\V{w}_a\|^2
\rb
\delta\lb Nm-\V{w}_0^{\top}\V{w}_a\rb
\rb
\prod_{a<b}\delta\lb Nq-\V{w}_a^{\top}\V{w}_b\rb.
\ee

We defer the details of its computation to \Rsec{Volume} and here only show the result: 
\be
\lim_{n\to 0,N\to\infty}\frac{1}{nN} \log V(1,q,m)
=
\Extr{\{\hat{Q},\hat{q},\hat{m}\}}
\Biggl\{
\frac{1}{2}\hat{Q} +\frac{1}{2}\hat{q}q -\hat{m}m
+\frac{1}{2}\log(2\pi)-\frac{1}{2}\log(\hat{Q}+\hat{q})
+
\frac{1}{2}\frac{\hat{m}^2+\hat{q}}{ \hat{Q}+\hat{q}}
\Biggr\},
\Leq{entropy}
\ee
where $\Extr{\{x\}}$ denotes the extremization w.r.t.~$x$. This extremization appears as the consequence of the saddle-point/Laplace method which is valid in the limit $N\to \infty$. The volume becomes finite only when $Q=1$, due to the normalization condition $\|\V{w}\|^2=N$. In the same way we can compute $\lim_{n\to 0,N\to\infty}\frac{1}{nN} \log L^M(Q,q,m)$.
%Combining these terms,
Substituting these expressions to Eq.~\eqref{eq:replica trick}, 
the free energy $\phi=-(\beta N)^{-1}\log Z$ averaged w.r.t.\ $D^M$
and in the limit $N\to\infty$ takes the following compact form:
\be
&&
-\beta \fe=\lim_{n\to 0,N\to\infty}\frac{1}{nN} \log \lsb Z^n \rsb_{D^M}
=\lim_{n\to 0,N\to\infty}\frac{1}{nN}\left[\log V(1,q,m)+\log L^M(1,q,m)\right]
\no \\ &&
=\Extr{\{\hat{Q},\hat{q},\hat{m},q,m\}}
\Biggl\{ 
\frac{1}{2}\hat{Q} +\frac{1}{2}\hat{q}q -\hat{m}m
+\frac{1}{2}\log(2\pi)-\frac{1}{2}\log(\hat{Q}+\hat{q})
+\frac{1}{2}\frac{\hat{m}^2+\hat{q}}{ \hat{Q}+\hat{q}}
\no \\ &&
+\alpha  
\sum_{y=\pm 1}r_y
\int Dz 
\log \lb \int Dt 
  e^{
-\beta 
s_y\lossS \lb h(t,z,1,q,m,\sigma_y,yb)\rb
}
\rb
\Biggr\},
\Leq{free energy}
\ee
where $\alpha=M/N$. The quantities $\{\hat{Q},\hat{q},\hat{m},q,m\}$ are the order parameters of the present system.

%%%%%%%%%%%%%%%%%%%%%
\paragraph{Zero-temperature limit $\beta \to \infty$}
Next we compute the zero-temperature limit $\beta \to \infty$.
\if0
We assume $\lim_{\beta\to\infty}|-\beta f|=\infty$:
Although whether this assumption 
%In this limit, we assume $|-\beta f| \to \infty$: whether this
holds or not depends on the choice of the loss and the value of $\alpha$, non-diverging $|-\beta f|$ implies that the energy $u$ becomes zero in the limit, leading to a possibility of multiplicity of the minimizer of $\mc{H}(\V{w}\mid D^M,b,s)$. Such a case is not suitable for the present purpose studying the effect of resampling/reweighing and we disregard it.  

Under this assumption, we assume the following scalings which are later shown to be consistent:
\fi
As discussed in \Rsec{Statistical mechanical},
the posterior of $\V{w}$ concentrates
on the set of minimizers of the loss in the limit $\beta\to\infty$.
We further expect that the minimizer is unique.
In view of this, we adopt the following ansatz
for the asymptotic behaviors of the order parameters around
the saddle points as $\beta$ becomes large:
\begin{align}
q&=1-\chi/\beta,&\chi&=O(1),&m&=O(1),
\no \\
\hat{Q}&=-\beta^2\T{\chi}+\beta\T{Q},&\hat{q}&=\beta^2\T{\chi},&\hat{m}&=\beta\T{m}.
\label{eq:scaling}
\end{align}
This ansatz will later turn out to be consistent. Inserting these relations into \Req{free energy}, performing the variable transform $v=t/\sqrt{\beta}$ and taking the limit $\beta \to \infty$ yield
\be
&&
u=\lim_{\beta \to \infty}\fe
=\Extr{\{\tilde{Q},\tilde{\chi},\tilde{m},\chi,m\}}
\Biggl\{ 
-\frac{1}{2}\T{Q} +\frac{1}{2}\T{\chi}\chi +\T{m}m
-\frac{1}{2}\frac{\T{m}^2+\T{\chi}}{ \T{Q}}
\no \\ &&
-\alpha %\lbb
\sum_{y=\pm1}r_y\int Dz\,G(v_y,h_y,s_y)
%r_+
%\int Dz  G(v_+,h_+,s_+)  
%+
%(1-r_+)
%\int Dz  G(v_-,h_-,s_-)
%\rbb
\Biggr\}.
\Leq{energy_RS}
\ee
where we let
\be
&&
G(v,h,s)=-\frac{1}{2}v^2-s\lossS \lb h\rb,
\Leq{G}
\\ &&
v_y=\argmax_{v}\lbb -\frac{1}{2}v^2-s_y\lossS\lb  \sigma_y \lb \sqrt{\chi}v + z\rb+m + yb \rb \rbb
\Leq{v_pm},
\\ &&
h_y=  \sigma_y \lb \sqrt{\chi}v_y + z\rb+m + yb
\Leq{h_pm}.
\ee
Here, $v_y$ denotes the maximum point of $G$ w.r.t.\ the variable $v$, %when taking the limit $\beta \to \infty$.
which appears in evaluating the inner integral w.r.t.\ $t$ in the last term of \Req{free energy} in the limit $\beta\to\infty$
via the saddle-point/Laplace method. When the loss function is nonconvex, the optimization problem in \Req{v_pm} may have multiple local optima,  and hence numerical solutions require careful consideration, such as exploring different initial conditions. Note also that the dependence of $v_y$ and $h_y$ on the integration variable $z$ is implicit in the above formulae. 

%%%%%%%%%%%%%%%%%%%
\paragraph{Equations of state (EOS)} 
The extremization condition in \Req{energy_RS} yields the following equations:
\subbe
\Leq{EOS}
\be
&&
\T{Q}^2=\T{m}^2+\T{\chi},
\Leq{Qt_sp}
\\ &&
m=\frac{\T{m}}{\T{Q}},
\Leq{m_sp}
\\ &&
\chi=\frac{1}{\T{Q}},
\Leq{chi_sp}
\\ &&
\T{m}=
\frac{\alpha}{\sqrt{\chi}}
\sum_{y=\pm1}\frac{r_y}{\sigma_y}\int Dz\,v_y,
%\lbb 
%\frac{r_+}{\sigma_+}\int Dz~v_+ + \frac{(1-r_+)}{\sigma_-}\int Dz~v_- 
%\rbb,
\Leq{mt_sp}
\\ &&
\T{\chi}=
\alpha
\sum_{y=\pm1}r_y\sigma_y^2\int Dz\,v_y^2.
%\lbb 
%r_+\sigma_+^2 \int Dz~v_+^2+(1-r_+)\sigma_-^2 \int Dz~v_-^2
%\rbb.
\Leq{chit_sp}
\ee
\subee
This set of equations is the EOS for the present problem.
An intuitive interpretation of the EOS is given in \Rsec{interp}.
The term $v_y$ appearing in \BReqs{mt_sp}{chit_sp}
is defined via \Req{v_pm} as a function of $m,\chi,b,y,s_y$.
Determining $v_y$ is easy if we are allowed to assume
differentiability of the loss $\lossS$ on $\mR$: %The derivation of the EOS is easy if we assume the differentiability of the loss in the whole $\mR$:
introducing the shorthand notation $\dloss=-\frac{d \lossS }{dh}$, we %have the equation to be satisfied by $v_y$ as
find that $v_y$ satisfies 
\be
v_y=s_y\sigma_y\sqrt{\chi} \dloss \lb h_y \rb.
\Leq{v_pm_diff}
\ee
It should be noted that $h_y$ appearing on the right-hand side depends on $v_y$ through \Req{h_pm}, so that the above equation is a (non-linear) equation on $v_y$.  If there are multiple solutions to \Req{v_pm_diff}, the one yielding the largest value of $G(v_y,h_y,s_y)$ should be selected. Under this differentiability assumption, the extremization condition w.r.t.\ $m$ can be computed from the partial derivative of $G$ w.r.t.\ $h$ as $\Part{G(v,h,s)}{h}{}\Part{h}{m}{}=s\dloss(h)$: this is because the partial derivative of $G$ w.r.t.\ $v$ vanishes since $v$ is fixed at the extremum value of $G$ given $h$ and $s$ as shown in \Req{v_pm}. The term $s\dloss(h)$ can be rewritten using \Req{v_pm_diff}, yielding the right-hand side of \Req{mt_sp}. A similar rewriting using \Req{v_pm_diff} is done for the extremization condition w.r.t.\ $\chi$, yielding the right-hand side of \Req{chit_sp}. Although the differentiability assumption does not hold for some losses having singularities such as the zero-one loss, those losses are usually representable as a limit of a certain differentiable function: for example, the zero-one loss can be expressed as
\be
\lossS_{01}(h)\coloneqq 
\left\{
\begin{array}{cc}
1  & (h<0)    \\
1/2  & (h=0)    \\
 0 & (h>0)     
\end{array}
\right\}
=\lim_{\gamma \to \infty}\frac{1}{2}\lb 1-\tanh(\gamma h)\rb
\eqqcolon \lim_{\gamma \to \infty}\lossS_{01,\gamma}(h).
\Leq{ell_01_smooth}
\ee
Hence, for such losses with singularities, the above discussion should be interpreted as that for such smoothed versions of the losses, and the smoothness-controlling parameter ($\gamma$ in \Req{ell_01_smooth}) is sent to an appropriate limit after the computation. The resultant formula \NReq{EOS} still holds after the limit even when the derivative $\dloss$ itself is not meaningful in the limit. 

 The EOS constitutes the basis of the following study. A special focus is on the overlap $m=\Ave{\V{w}}^{\top}\V{w}_0/N$, because it characterizes the performance of the feature learning as discussed in \Rsec{Classifier and}.

%%%%%%%%%%%%%%%%%%%%%%%%%%%%%%%%%%%%%%%%%%%%%%%%%%%%%
%%%%%%%%%%%%%%%%%%%%%%%%%%%%%%%%%%%%%%%%%%%%%%%%%%%%%
\subsection{Behaviors of quantities of interest}\Lsec{Behaviors of}

%%%%%%%%%%%%%%%%%%%%%%%%%%%%%%%%%%%%%%%%%%%%%%%%%%%%%
\subsubsection{Overview}
By specifying the functional form of the loss $\lossS$, one can numerically obtain values of the order parameters and the energy, via computing $v_{\pm}$ with the loss and numerically solving the EOS. 
We consider two choices for the loss $\lossS$ as representative examples: the zero-one loss with the perceptron $\lossS_{\rm 01pe}(h,y)=(1-\sgn{yh})/2$ and the CE loss with the logistic function $\lossS_{\rm CElo}(h,y)=-hy+\log (2\cosh(h))$. Hereafter these two losses are referred to as 01pe and CElo, respectively. The aim here is to investigate how the order parameters depend on the parameters in order to assess the feature learning performance. 
%As two representative examples, we analyze the two cases: the zero-one loss with the perceptron $\lossS_{01pe}(h,y)=(1-\sgn{yh})/2$ and the CE loss with the logit function $\lossS_{\rm CElo}(h,y)=-hy+\log (2\cosh(h))$.
In the following, we investigate two cases:
one is the equal-variance case $\sigma_{+}^2=\sigma_{-}^2$,
where the two classes $y=\pm1$ share the same variance,
and the other is the nonequal-variance case $\sigma_{+}^2\not=\sigma_{-}^2$,
where the variances of the classes $y=\pm1$ are different. 

%%%%%%%%%%%%%%%%%%%%%%%%%%%%%%%%%%%%%%%%%%%%%%%%%%%%%
\subsubsection{Equal-variance case $(\sigma_{+}^2=\sigma_{-}^2)$}\Lsec{Equivariance case}
We start from a canonical situation where the two class variances are equal: $\sigma_{+}^2=\sigma_{-}^2\eqqcolon \sigma^2$. As examples, we compare the balanced case $r_+=0.5$ and an imbalanced case $r_+=0.2$, with $\sigma=0.6$. For illustration of these cases, the probability density functions (PDF) of $\V{x}$ projected onto $\V{w}_0/\sqrt{N}$ are plotted in \Rfig{PDF_equivariance}. 
%%%%%%%%%%%%%%%%%%%%%%%
\begin{figure}[htbp]
\begin{center}
  %\vspace{0mm}\leavevmode
  \begin{minipage}{0.45\columnwidth}
    \centering
    $\sigma=0.6$, $r_+=0.5$\\
    \includegraphics[width=\columnwidth,clip,trim=0 48 0 38]{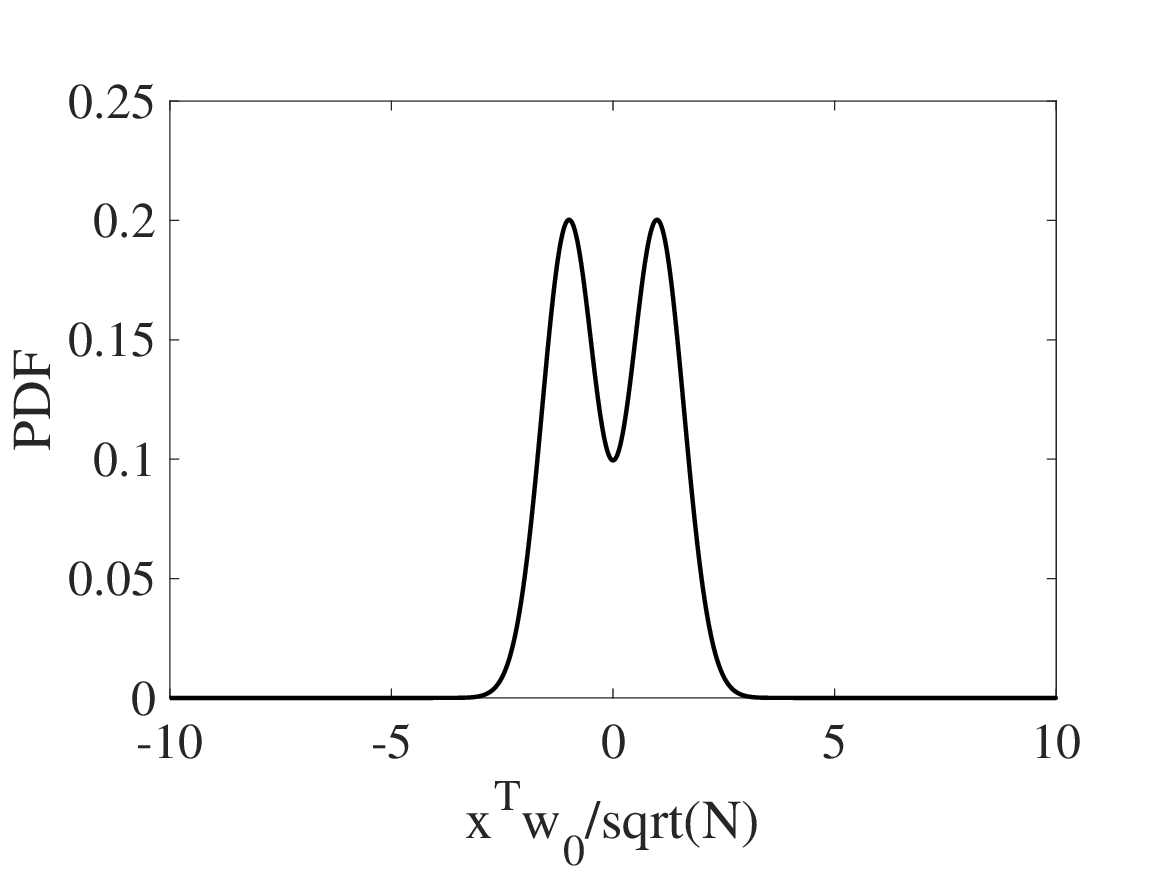}\\
    $\bm{w}_0^\top\bm{x}/\sqrt{N}$
  \end{minipage}
  \begin{minipage}{0.45\columnwidth}
    \centering
    $\sigma=0.6$, $r_+=0.2$\\
    \includegraphics[width=\columnwidth,clip,trim=0 48 0 38]{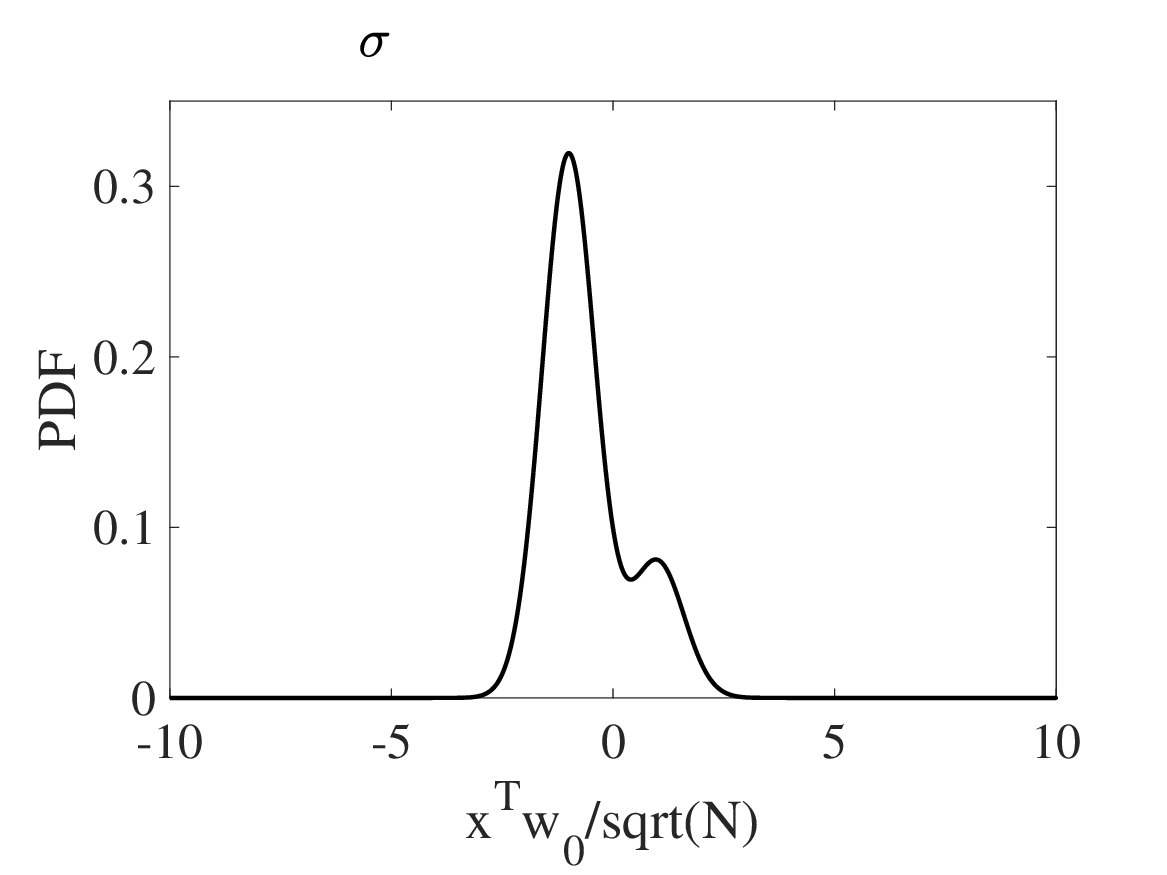}\\
    $\bm{w}_0^\top\bm{x}/\sqrt{N}$
  \end{minipage}
\caption{PDFs of $\V{x}$ projected onto $\V{w}_0$ with $\sigma=0.6$ for $r_+=0.5$ (left) and $0.2$ (right).} 
\Lfig{PDF_equivariance}
\end{center}
\end{figure}
%%%%%%%%%%%%%%%%%%%%%%%
Besides, we fix $\alpha=20$ in the following plots unless otherwise stated; the results for other values of $\alpha$, if not too small, were qualitatively similar as far as we have checked.  

Firstly, we show $m$ and $u$ plotted against $b$ for $s_+=0.1,0.5,0.9$: the results for $r_+=0.5$ and $0.2$ are shown in \Rfig{orderparam_al20_sig06_r05} and  \Rfig{orderparam_al20_sig06_r02}, respectively.
It should be noted that the class centers are located
at $\pm\V{w}_0/\sqrt{N}$ in our problem setting, so that a reasonable choice
of the parameter $b$ in view of the task of classification
would intuitively be $b\in[-1,1]$: indeed, if otherwise, the two class centers $\pm\V{w}_0/\sqrt{N}$ are located on the same side
of the decision plane $\V{w}^\top\V{x}/\sqrt{N}+b=0$.
In the following, we nevertheless investigate the behaviors of the models over
wider ranges of $b$. 
%%%%%%%%%%%%%%%%%%%%%%%
\begin{figure}[htbp]
\begin{center}
\vspace{0mm}
\includegraphics[width=0.34\columnwidth]{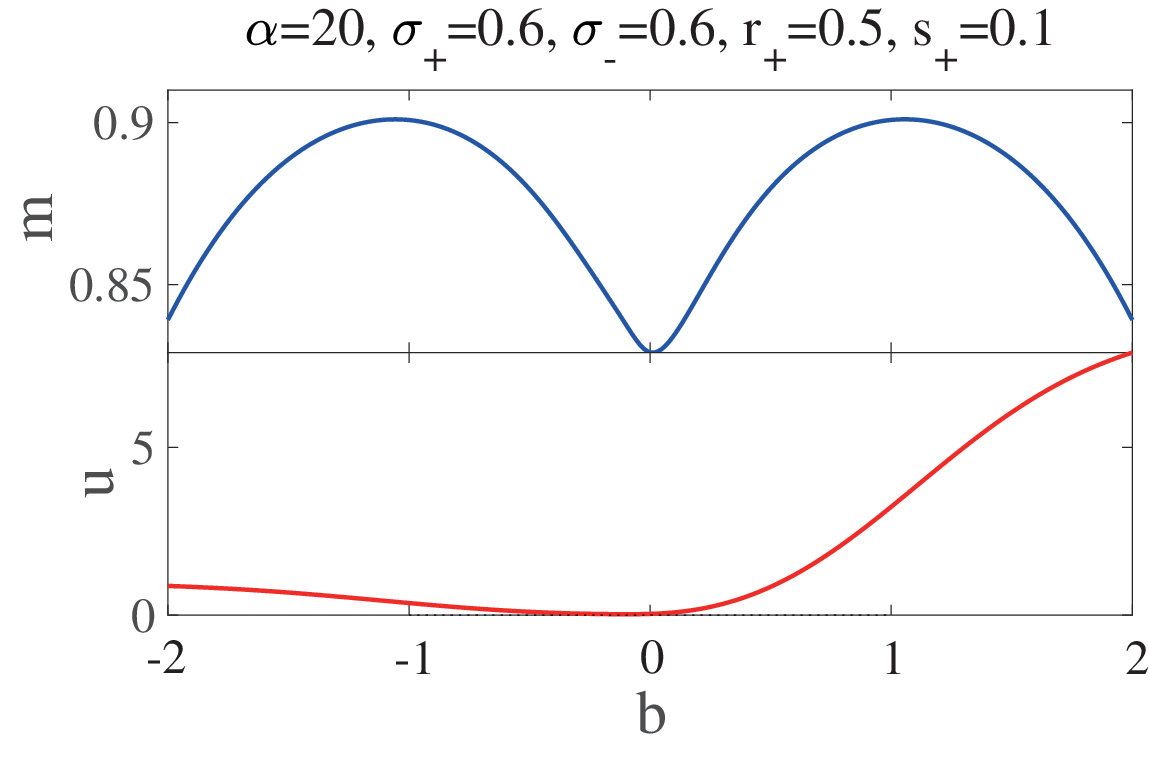}
\includegraphics[width=0.32\columnwidth]{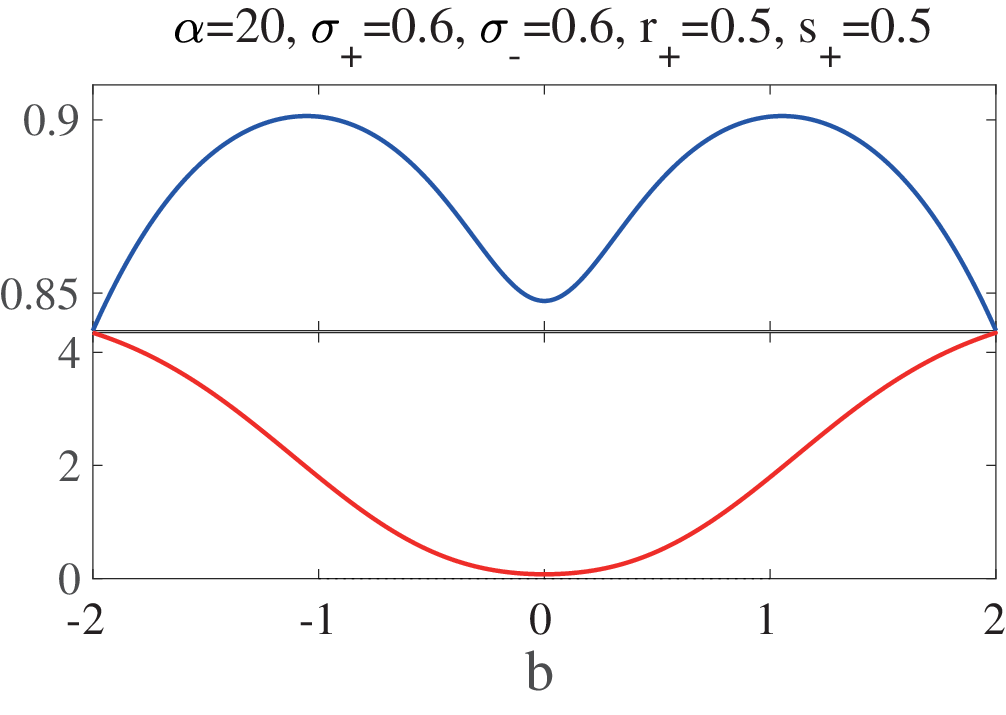}
\includegraphics[width=0.32\columnwidth]{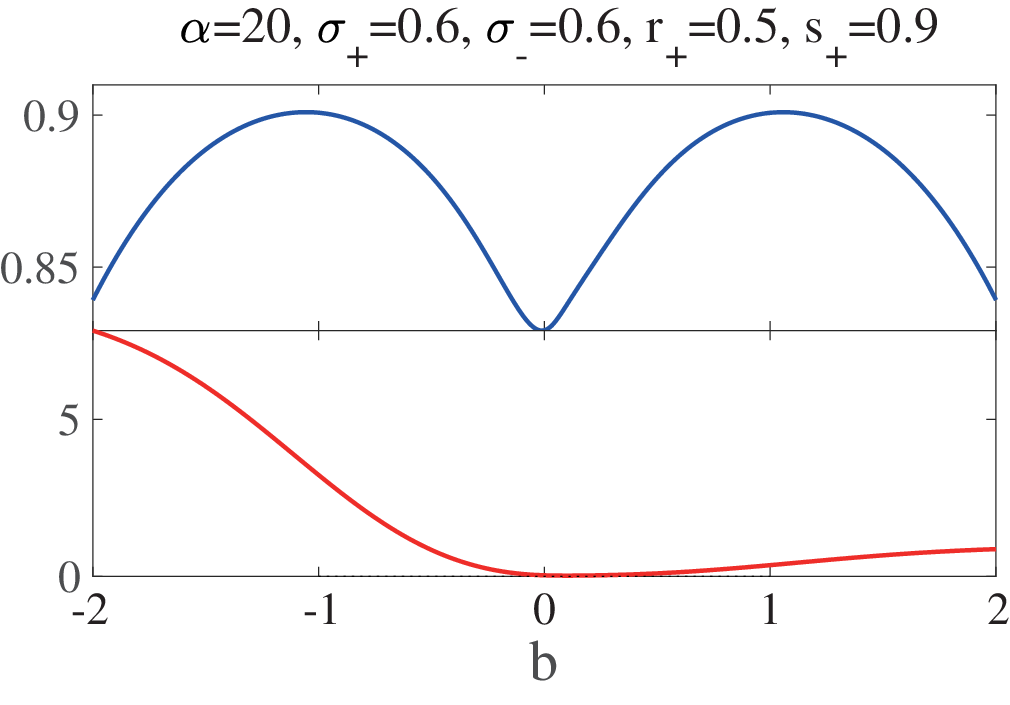}\\
(a) Zero-one loss with perceptron $\lossS_{\mathrm{01pe}}$.\\
\includegraphics[width=0.34\columnwidth]{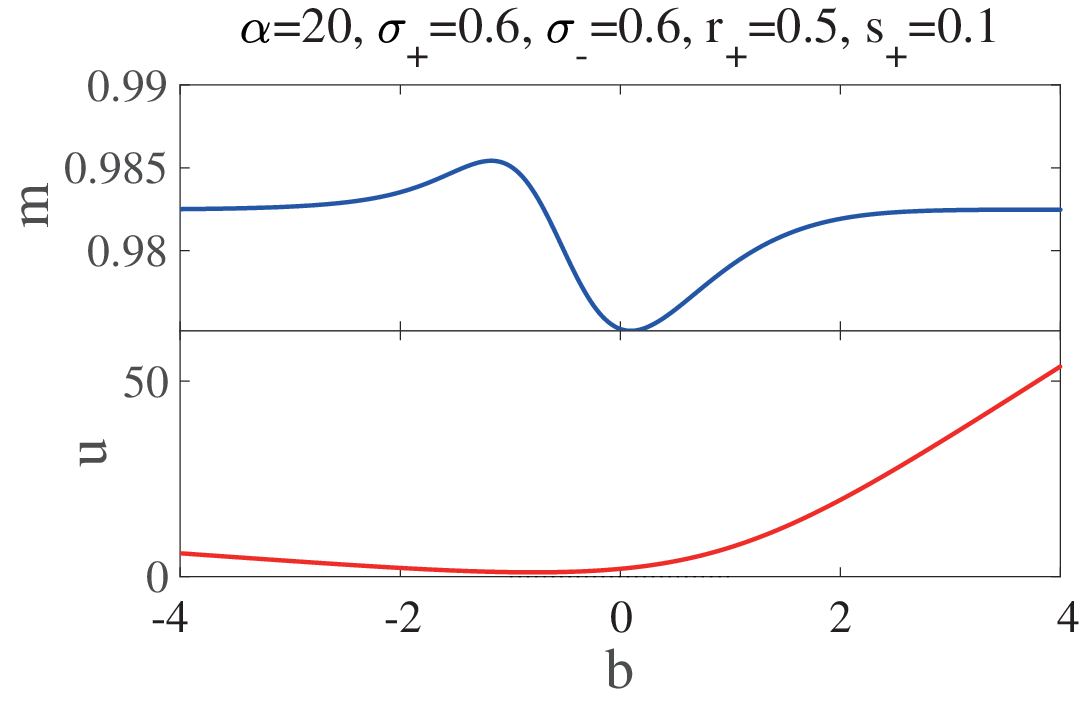}
\includegraphics[width=0.32\columnwidth]{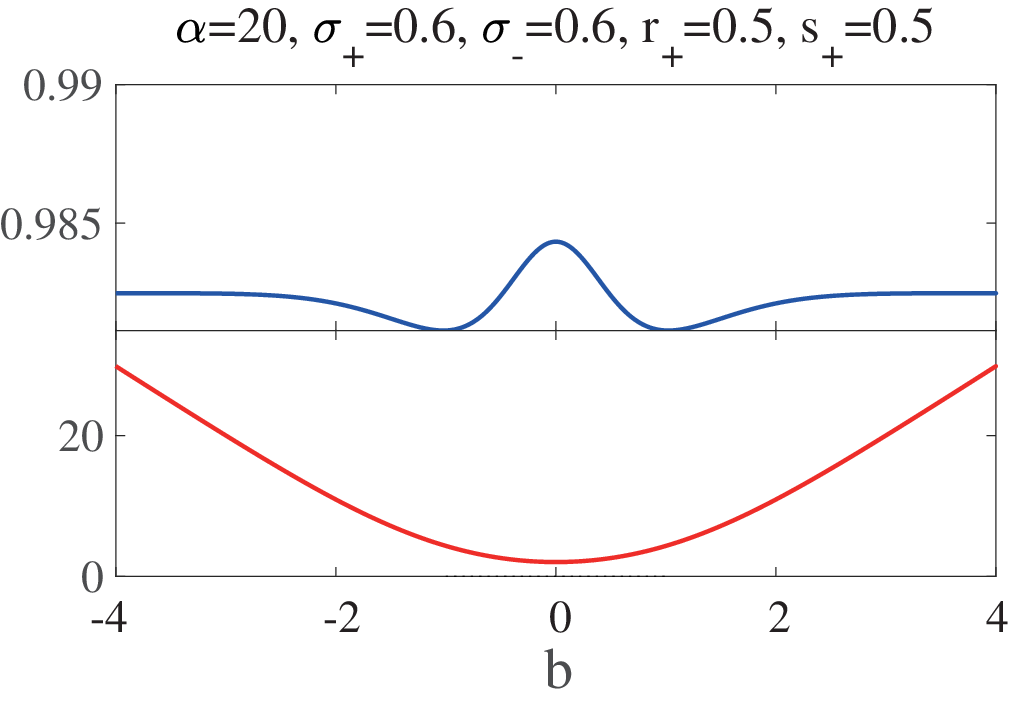}
\includegraphics[width=0.32\columnwidth]{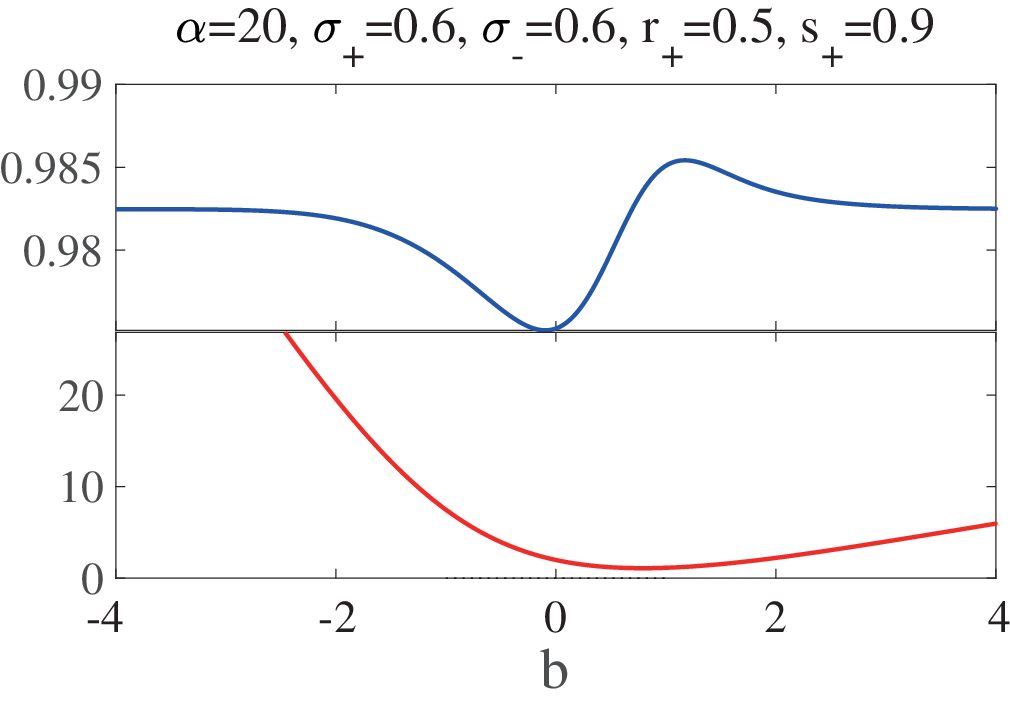}\\
(b) Cross-entropy loss with logistic function $\lossS_{\mathrm{CElo}}$.
\vspace{0mm}
\caption{Plots of $m$ and $u$ against $b$ in the balanced case $r_+=0.5$ for $s_+=0.1$ (left), $0.5$ (middle), and $0.9$ (right). (a) Zero-one loss with perceptron $\lossS_{\mathrm{01pe}}$. (b) Cross-entropy loss with logistic function $\lossS_{\mathrm{CElo}}$. } 
\Lfig{orderparam_al20_sig06_r05}
\end{center}
\end{figure}
%%%%%%%%%%%%%%%%%%%%%%%
%%%%%%%%%%%%%%%%%%%%%%%
\begin{figure}[htbp]
\begin{center}
\vspace{0mm}
\includegraphics[width=0.34\columnwidth]{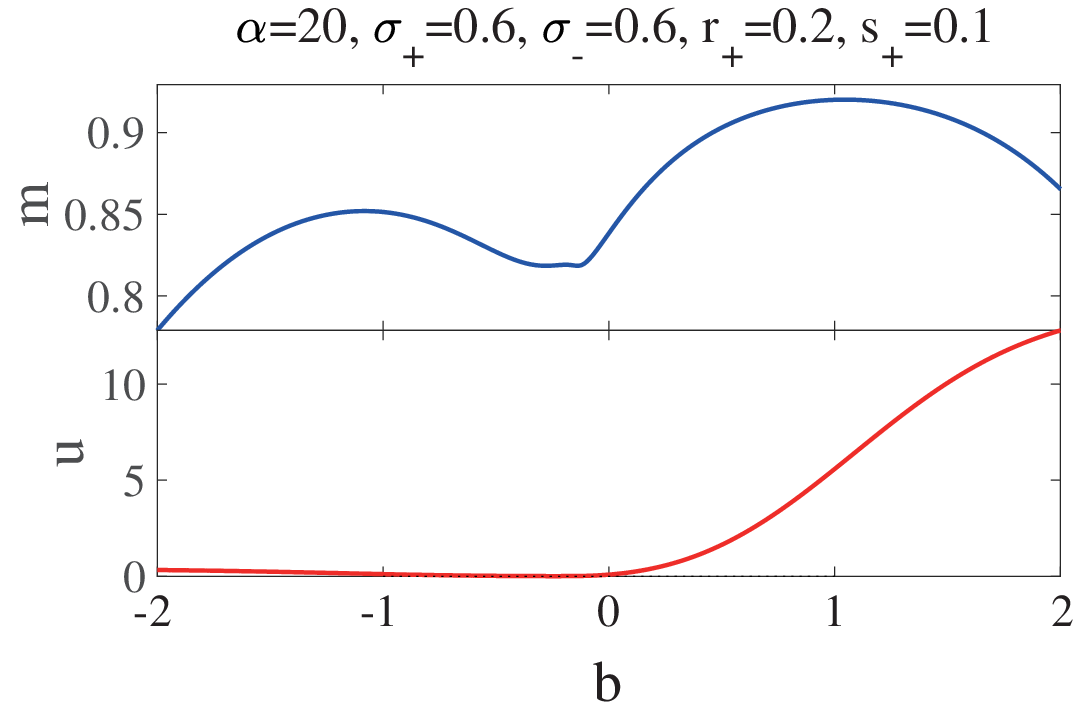}
\includegraphics[width=0.32\columnwidth]{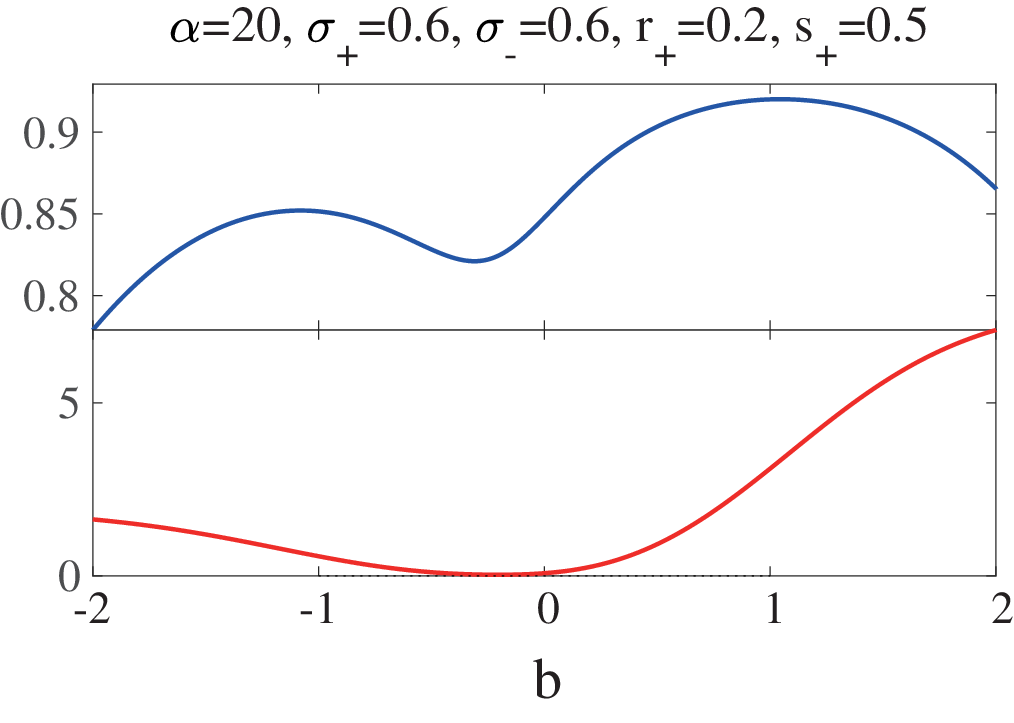}
\includegraphics[width=0.32\columnwidth]{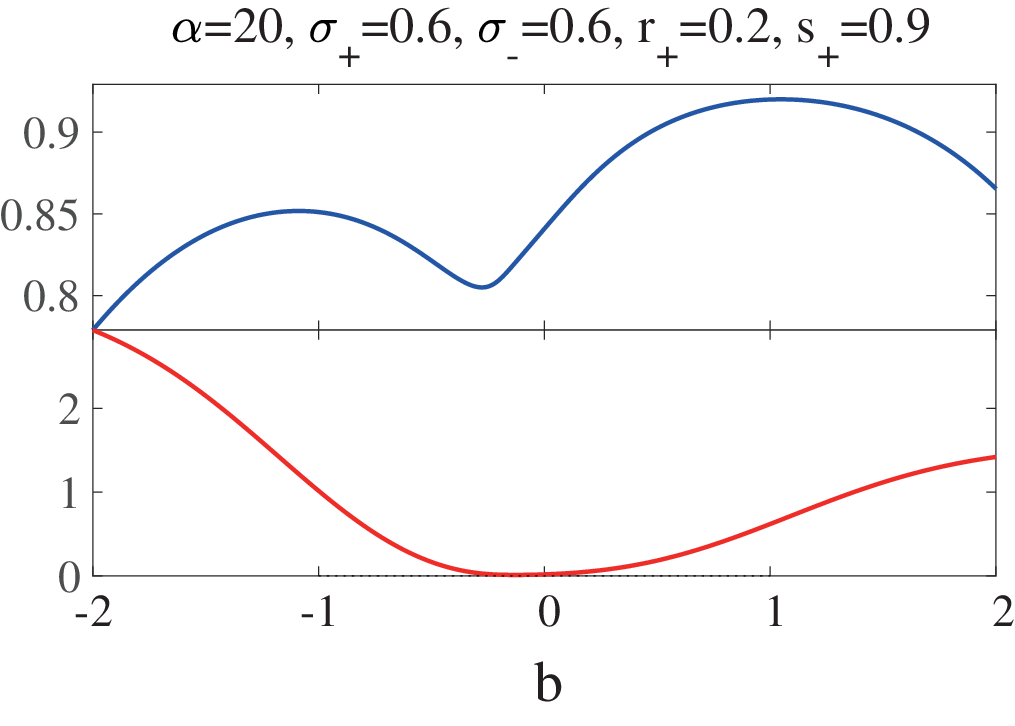}\\
(a) Zero-one loss with perceptron $\lossS_{\mathrm{01pe}}$.\\
\includegraphics[width=0.34\columnwidth]{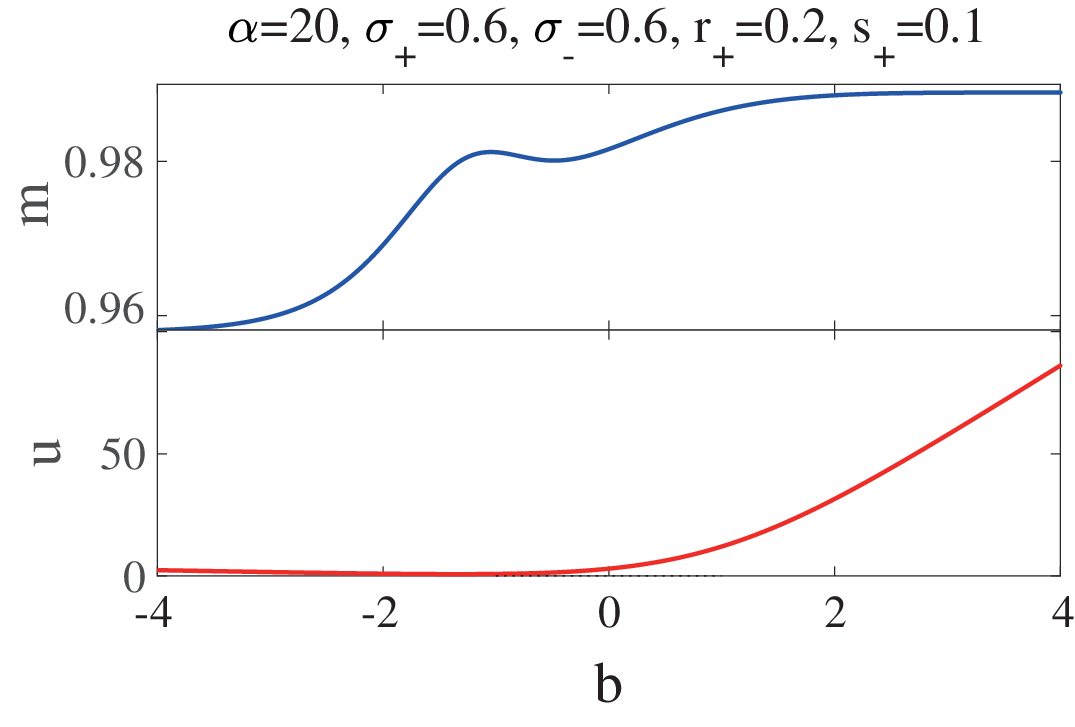}
\includegraphics[width=0.32\columnwidth]{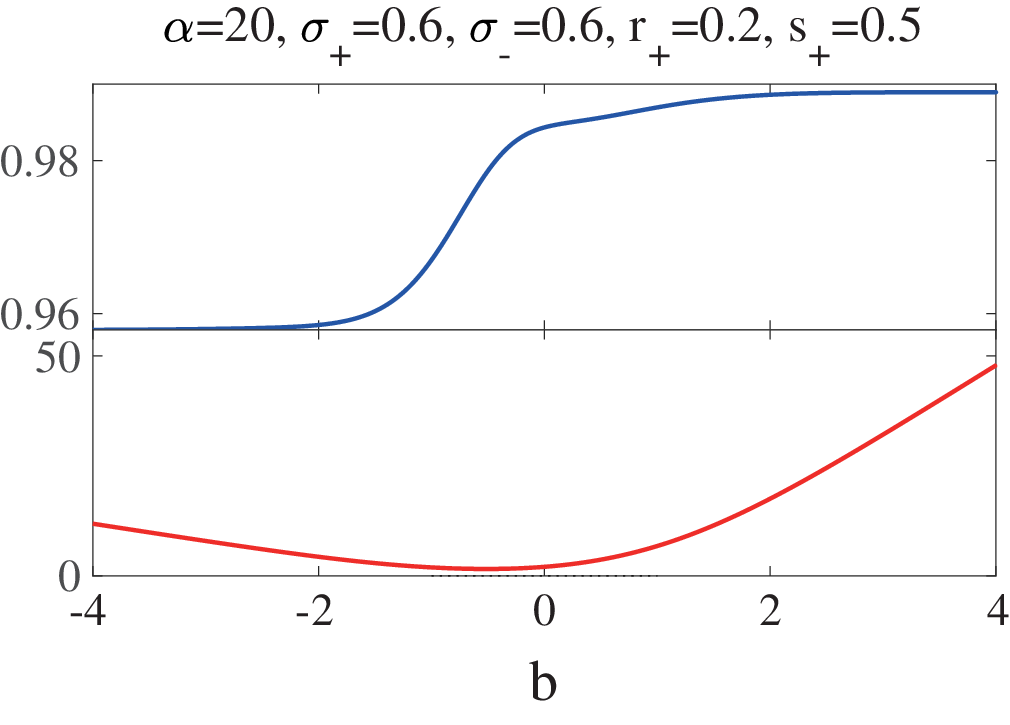}
\includegraphics[width=0.32\columnwidth]{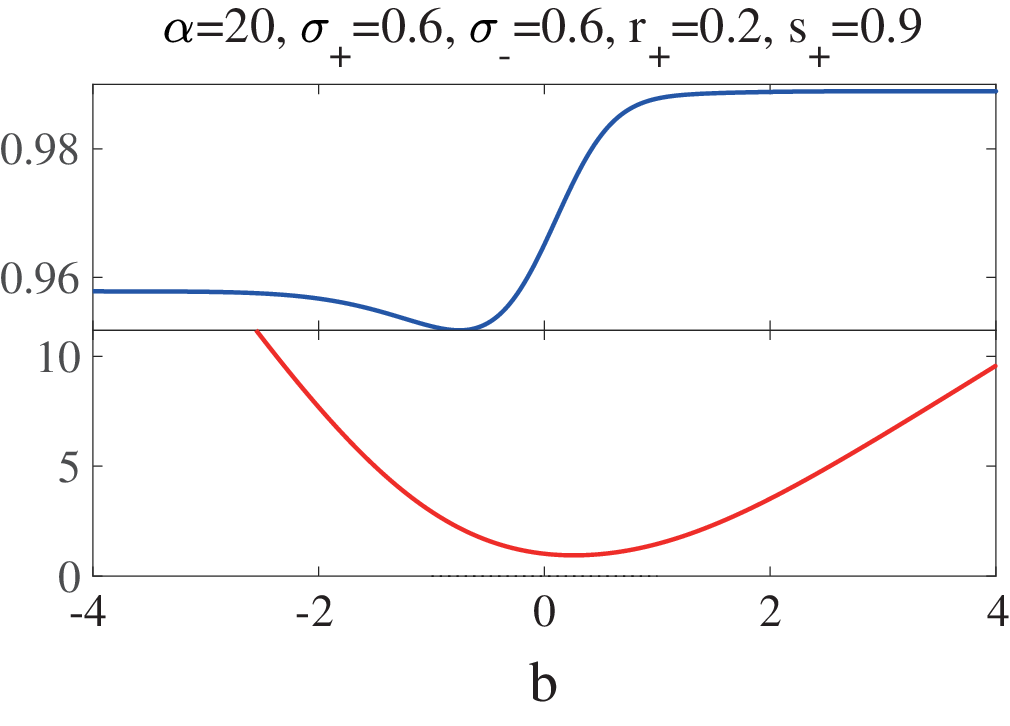}\\
(b) Cross-entropy loss with logistic function $\lossS_{\mathrm{CElo}}$.
\vspace{0mm}
\caption{Plots of $m$ and $u$ against $b$ in the imbalanced case $r_+=0.2$ for $s_+=0.1$ (left), $0.5$ (middle), and $0.9$ (right). (a) Zero-one loss with perceptron $\lossS_{\mathrm{01pe}}$. (b) Cross-entropy loss with logistic function $\lossS_{\mathrm{CElo}}$.} 
\Lfig{orderparam_al20_sig06_r02}
\end{center}
\end{figure}
%%%%%%%%%%%%%%%%%%%%%%%
\Rfig{orderparam_al20_sig06_r05} (a) and \Rfig{orderparam_al20_sig06_r02} (a)
are the results with 01pe, whereas 
\Rfig{orderparam_al20_sig06_r05} (b) and \Rfig{orderparam_al20_sig06_r02} (b)
are those with CElo.
%In both figures, the upper panels, each consisting of two plots corresponding to $m$ and $u$, are for 01pe  and the lower ones are for CElo.
As observed in \Rfig{orderparam_al20_sig06_r05}, the $b$-dependence of $m$ in the balanced case is very different between 01pe and CElo: with 01pe, it is almost symmetric w.r.t.\ $b$ even with the strong reweighting factors $s_+=0.1$ and $0.9$, while it exhibits clear asymmetry with CElo. The values of $b$ at which $m$ achieves its maximum are also different: $m$ achieves its maximum around $b\approx \pm 1$ with 01pe, while with CElo the maximum is achieved at $b$ less than $-1$ for $s_+=0.1$, at $b=0$ for $s_+=0.5$, and at $b$ larger than $1$ for $s_+=0.9$.
Meanwhile, as observed in \Rfig{orderparam_al20_sig06_r02} for the imbalanced case $r_+=0.2$, the maximum of $m$ seems to be obtained at a large positive $b$ in all the cases: this is considered to be natural since the positive bias enhances the probability of the minority class at $r_+=0.2$. As an overall tendency, the value of $m$ tends to be larger with CElo than that with 01pe, suggesting the superiority of the CE loss in the feature learning.

Secondly, to examine the maximum performance of feature learning, we compute the maximum of $m$ against $b$, $m_{\rm max}=\max_{b}m(b)$, and plot it against $s_+$ in \Rfig{mmax_al20_sig06}. \Rfig{mmax_al20_sig06} (a) and (b) are with 01pe and CElo, respectively; the left and right columns are the plots with $r_+=0.5$ and $0.2$, respectively. The maximum location $b(m_{\rm max}):=\argmax_{b}m(b)$ and the corresponding energy value $u(m_{\rm max}):=u(b(m_{\rm max}))$ are also plotted. 
%%%%%%%%%%%%%%%%%%%%%%%
\begin{figure}[htbp]
\begin{center}
\vspace{0mm}
\includegraphics[width=0.485\columnwidth]{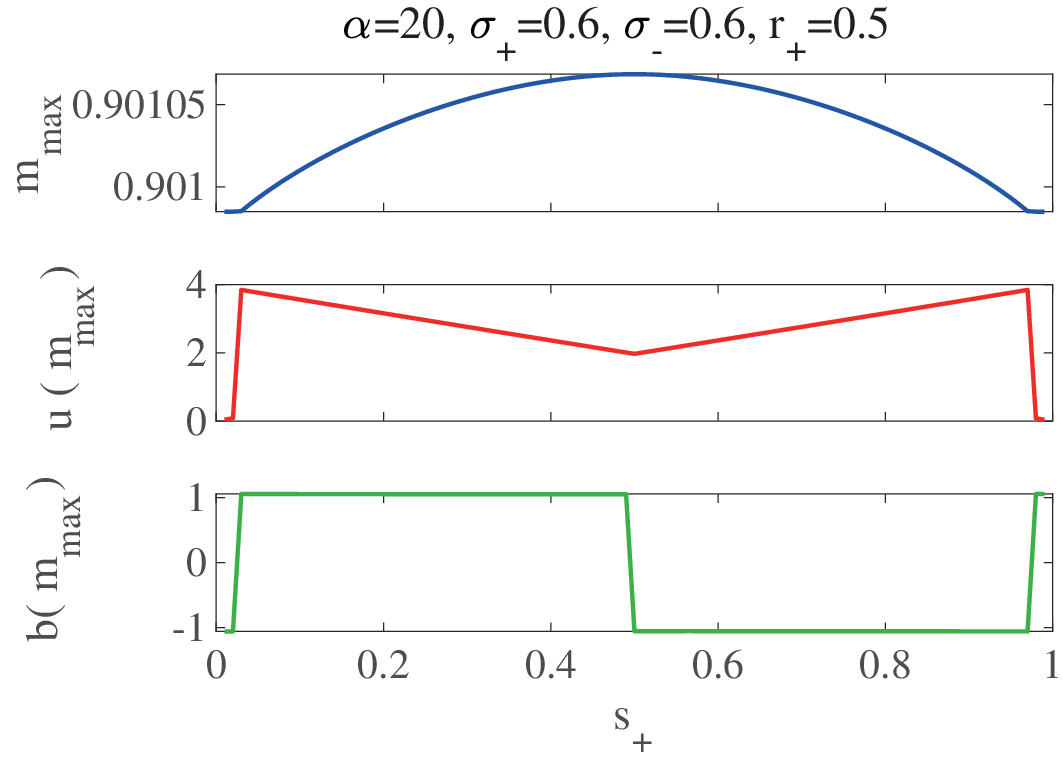}
\includegraphics[width=0.45\columnwidth]{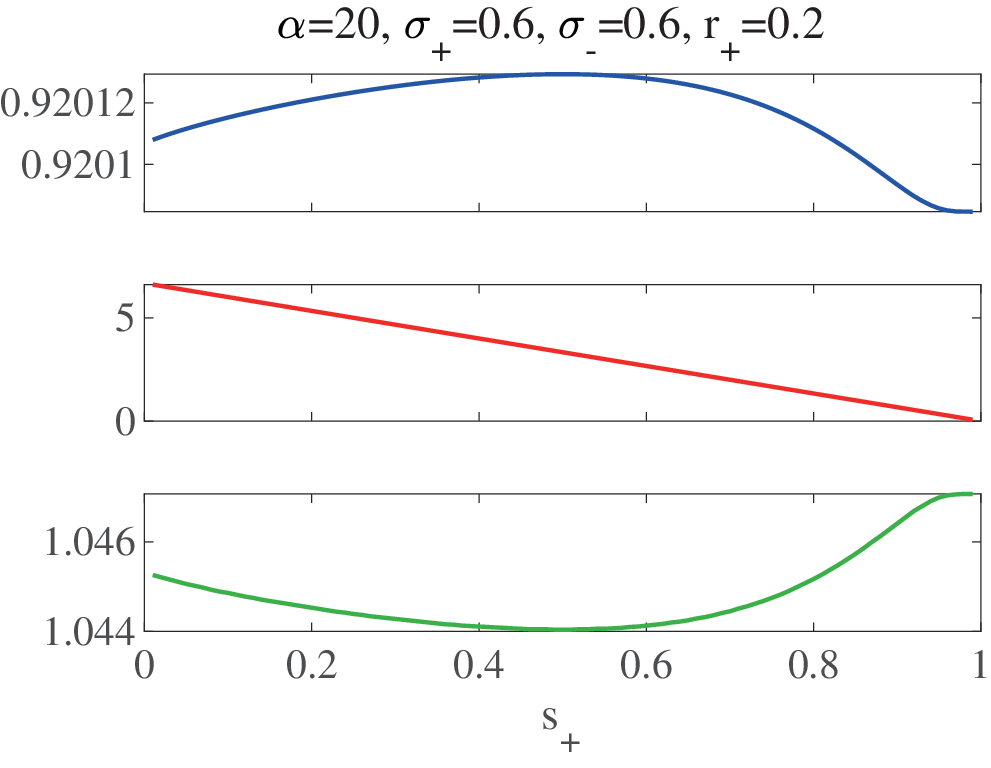}\\
(a) Zero-one loss with perceptron $\lossS_{\mathrm{01pe}}$.\\
\includegraphics[width=0.485\columnwidth]{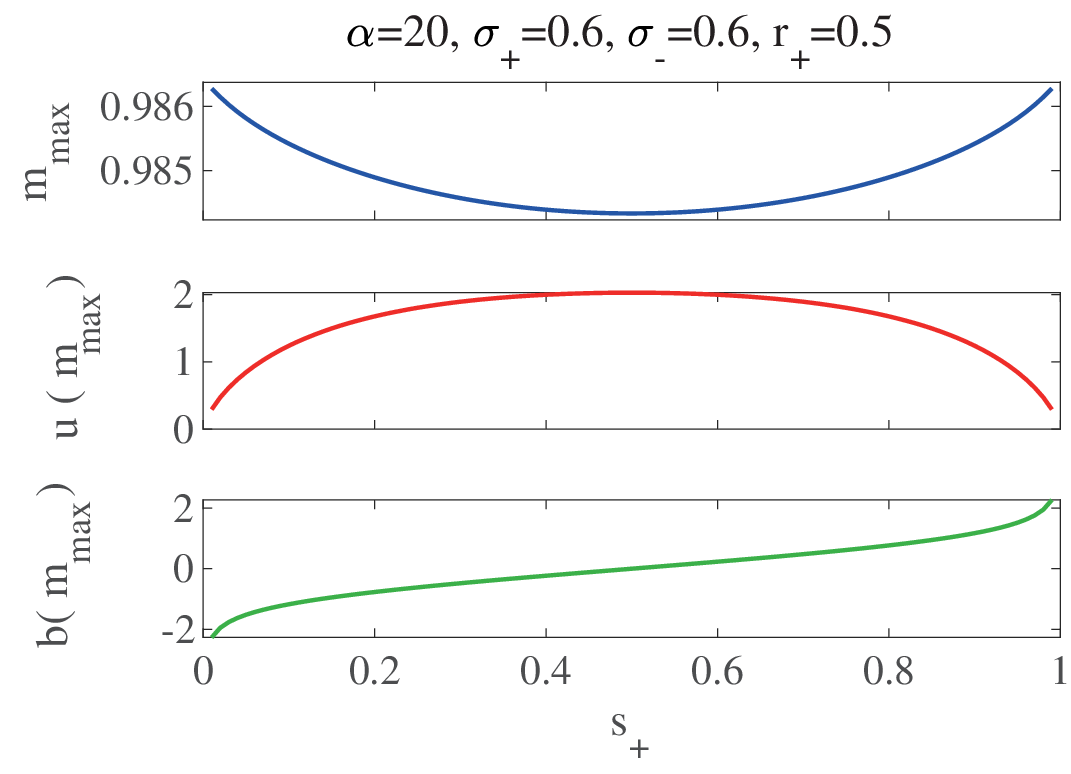}
\includegraphics[width=0.45\columnwidth]{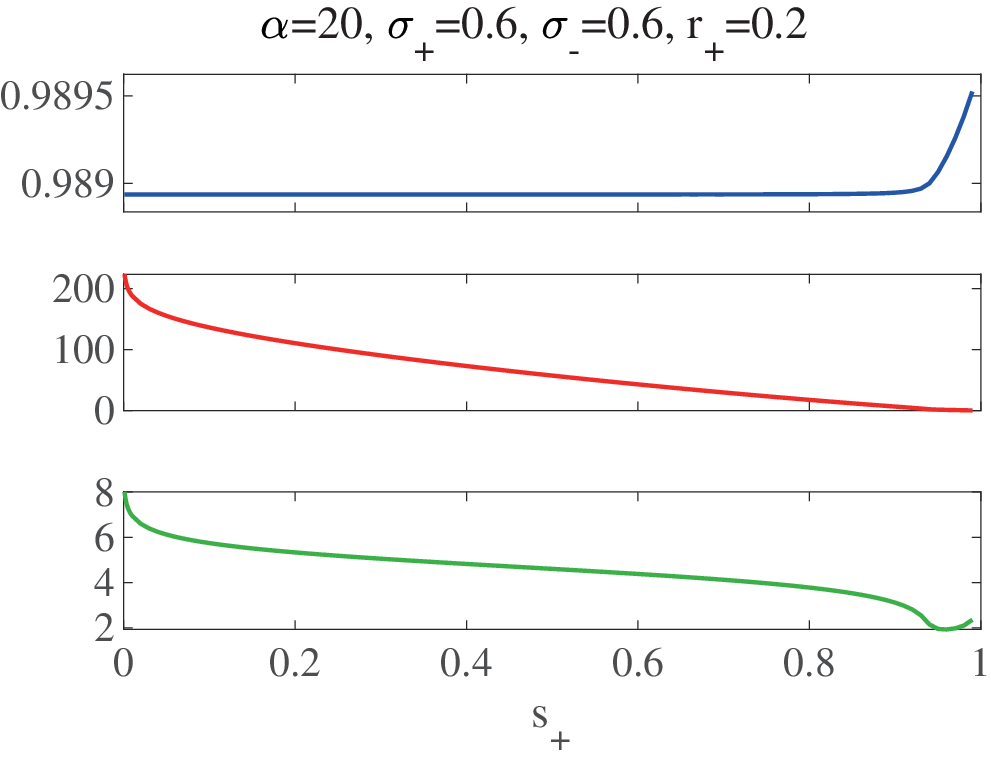}\\
(b) Cross-entropy loss with logistic function $\lossS_{\mathrm{CElo}}$.
\vspace{0mm}
\caption{Plots of $m_{\rm max},u(m_{\rm max}),b(m_{\rm max})$ against $s_+$ for $r_+=0.5$ (left) and $r_+=0.2$ (right). %The upper panels are for 01pe and the lower ones are for CElo. }
(a) Zero-one loss with perceptron $\lossS_{\mathrm{01pe}}$. 
(b) Cross-entropy loss with logistic function $\lossS_{\mathrm{CElo}}$.}
\Lfig{mmax_al20_sig06}
\end{center}
\end{figure}
%%%%%%%%%%%%%%%%%%%%%%%
As a general trend, $m_{\mathrm{max}}$ shows very weak dependence on $s_+$
in all the cases investigated,
as can be seen in the very small ranges of the plots.
An interesting observation is that with 01pe the maximum of $m_{\rm max}$ is obtained at the no-reweighting situation $s_+=1/2$ even in the imbalanced case (\Rfig{mmax_al20_sig06} (a), right), while with CElo the $m$'s maximum is located at some value different from $s_+=1/2$ even in the balanced case $r_+=0.5$ (\Rfig{mmax_al20_sig06} (b), left). This property of 01pe persisted as far as we have numerically investigated. This may be related to Kang et al.'s observation, although 01pe is not typically used in practical situations. 

In practical situations one cannot directly maximize the overlap $m$
since one does not know $\V{w}_0$. 
%As one cannot evaluate the maximum overlap $m_{\rm max}$ in realistic situations,
One will instead minimize the loss to obtain a reasonable estimator. From this viewpoint, thirdly, the minimum of $u$ w.r.t.\ $b$, $u_{\rm min}=\min_{b}u(b)$, is plotted against $s_+$ in \Rfig{umin_al20_sig06}, with the minimum location $b(u_{\rm min})=\argmin_{b}u(b)$ and the corresponding overlap value $m(u_{\rm min})=m(b(u_{\rm min}))$. 
%%%%%%%%%%%%%%%%%%%%%%%
\begin{figure}[htbp]
\begin{center}
\vspace{0mm}
\includegraphics[width=0.485\columnwidth]{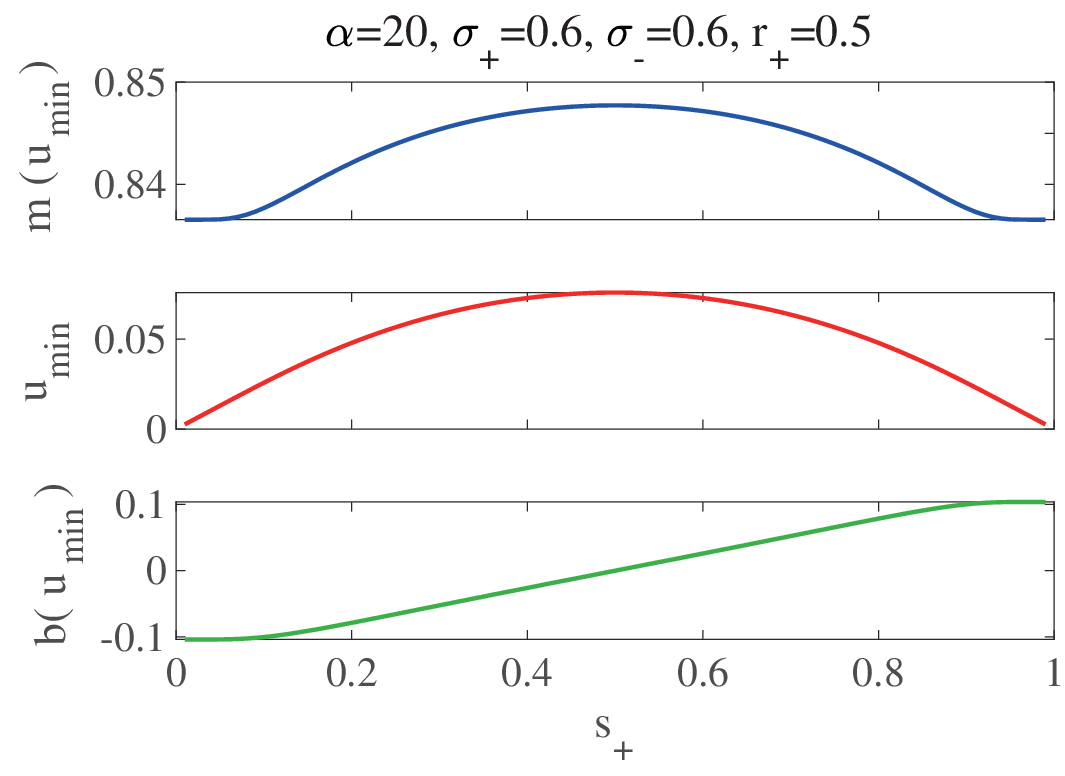}
\includegraphics[width=0.45\columnwidth]{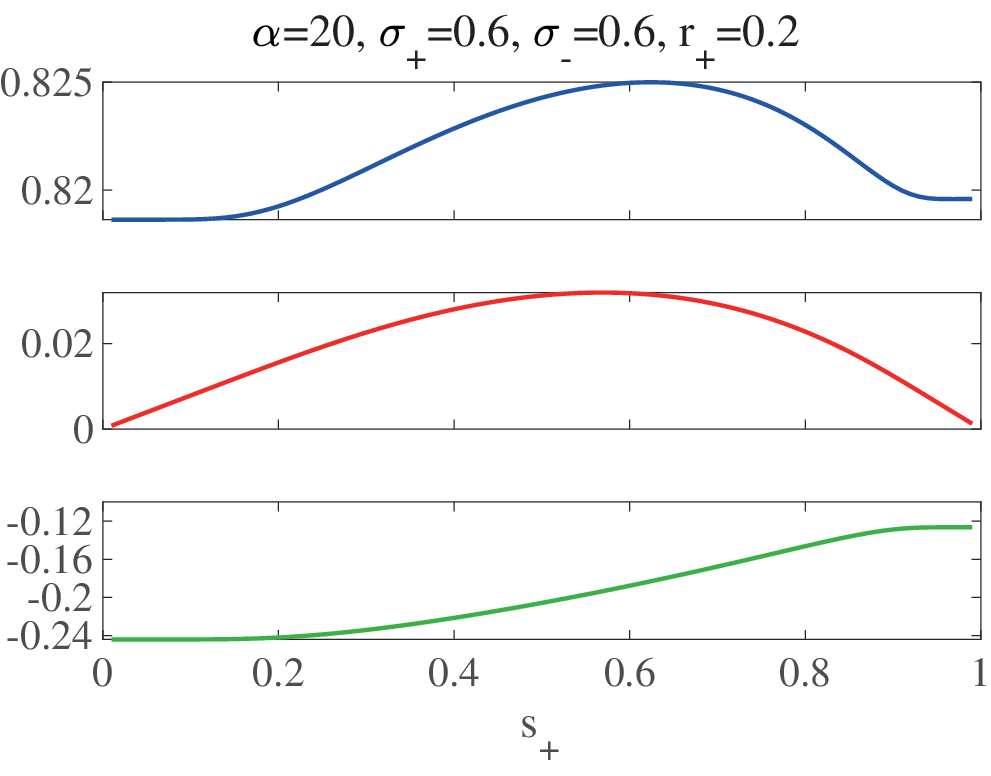}\\
(a) Zero-one loss with perceptron $\lossS_{\mathrm{01pe}}$. \\
\includegraphics[width=0.485\columnwidth]{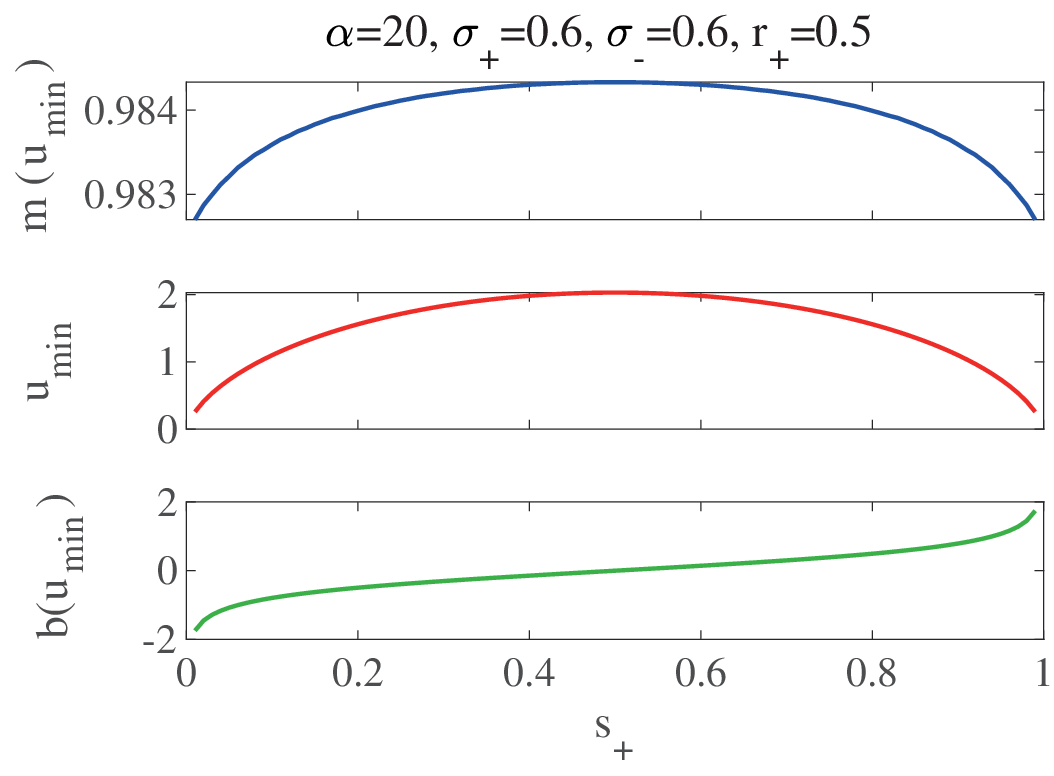}
\includegraphics[width=0.45\columnwidth]{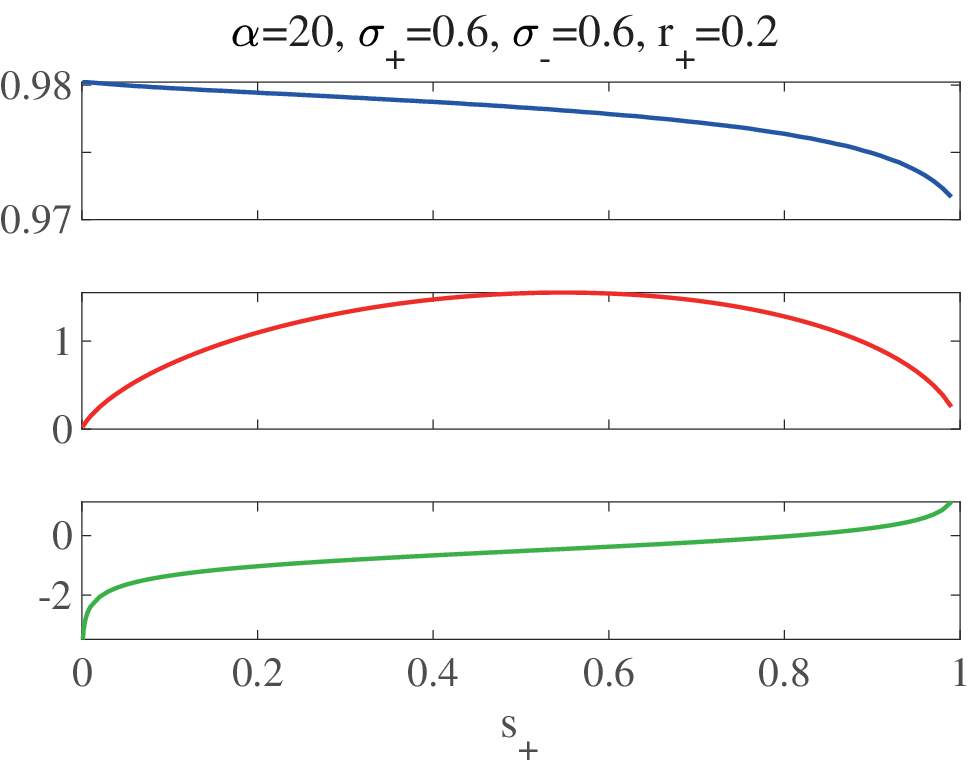}\\
(b) Cross-entropy loss with logistic function $\lossS_{\mathrm{CElo}}$.
\vspace{0mm}
\caption{Plots of $m(u_{\rm min}),u_{\rm min},b(u_{\rm min})$ against $s_+$ for $r_+=0.5$ (left) and $r_+=0.2$ (right). %The upper panels are for 01pe and the lower ones are for CElo. } 
(a) Zero-one loss with perceptron $\lossS_{\mathrm{01pe}}$. 
(b) Cross-entropy loss with logistic function $\lossS_{\mathrm{CElo}}$.}
\Lfig{umin_al20_sig06}
\end{center}
\end{figure}
%%%%%%%%%%%%%%%%%%%%%%%
This figure shows an intriguing behavior of $m(u_{\rm min})$: in the balanced case $r_+=0.5$, its curve is symmetric around $s_+=0.5$ and the maximum is obtained at $s_+=0.5$ as expected, with both 01pe and CElo. In the imbalanced case $r_+=0.2$, however, the tendency is different between 01pe and CElo: with 01pe, there is a maximum at a certain value of $s_+$ greater than $0.5$, which is natural because the values of $s_+$ greater than $0.5$ enhance the probability of the minority (positive here) class. With CElo, on the other hand, the opposite occurs and the overlap maximum is obtained at $s_+=0,$ meaning the complete disregard of the minority class, and we numerically confirmed that the corresponding minimum location $b(u_{\rm min})$ goes to $-\infty$. This is rather counterintuitive, and provides a lesson that it is not straightforward to predict how the interplay among the loss, classifier, and reweighting factor would influence the feature learning performance.

In recent practices, the bias $b$ is occasionally neglected~\citep{cao2019learning,menon2020long,kini2021label}. This is because DNNs learn feature vectors also from the data, and hence the adjustment of the feature space origin, that is the effect of the bias, can be incorporated by the learning even without the bias. Accordingly, we lastly examine the case $b=0$, which is actually a natural choice since the origin is located at the middle point of the two class centers in our setting. The overlap and energy in this case are plotted against $s_+$ in \Rfig{b0_al20_sig06}. 
%%%%%%%%%%%%%%%%%%%%%%%
\begin{figure}[htbp]
\begin{center}
\vspace{0mm}
\includegraphics[width=0.485\columnwidth]{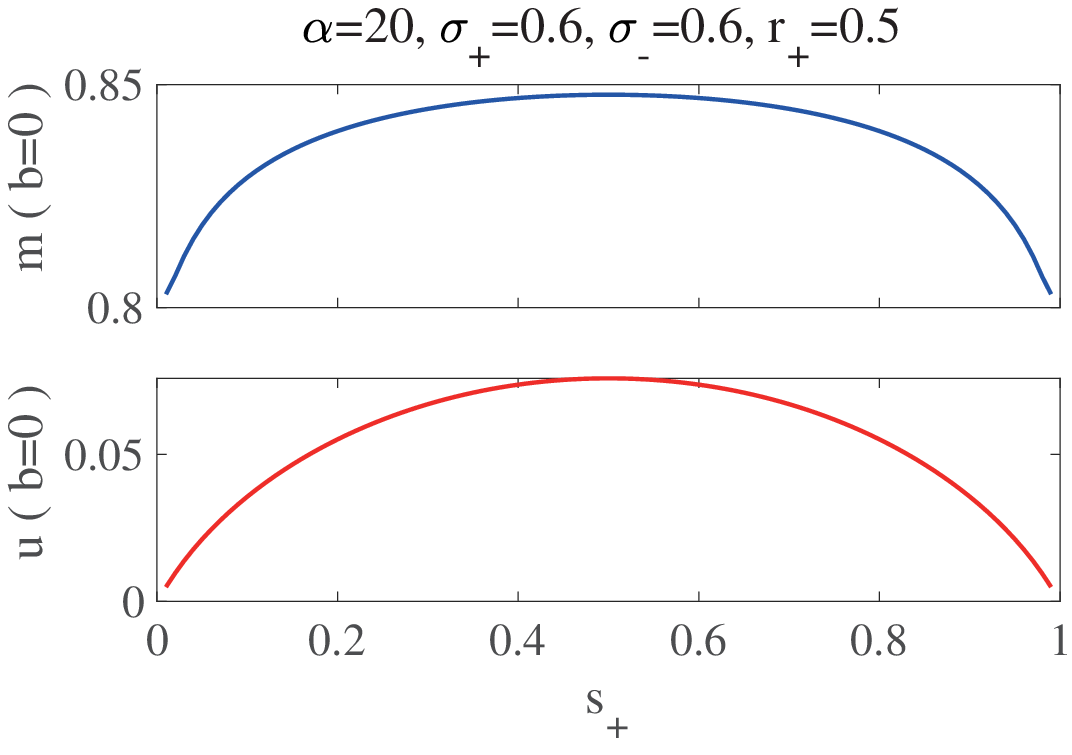}
\includegraphics[width=0.46\columnwidth]{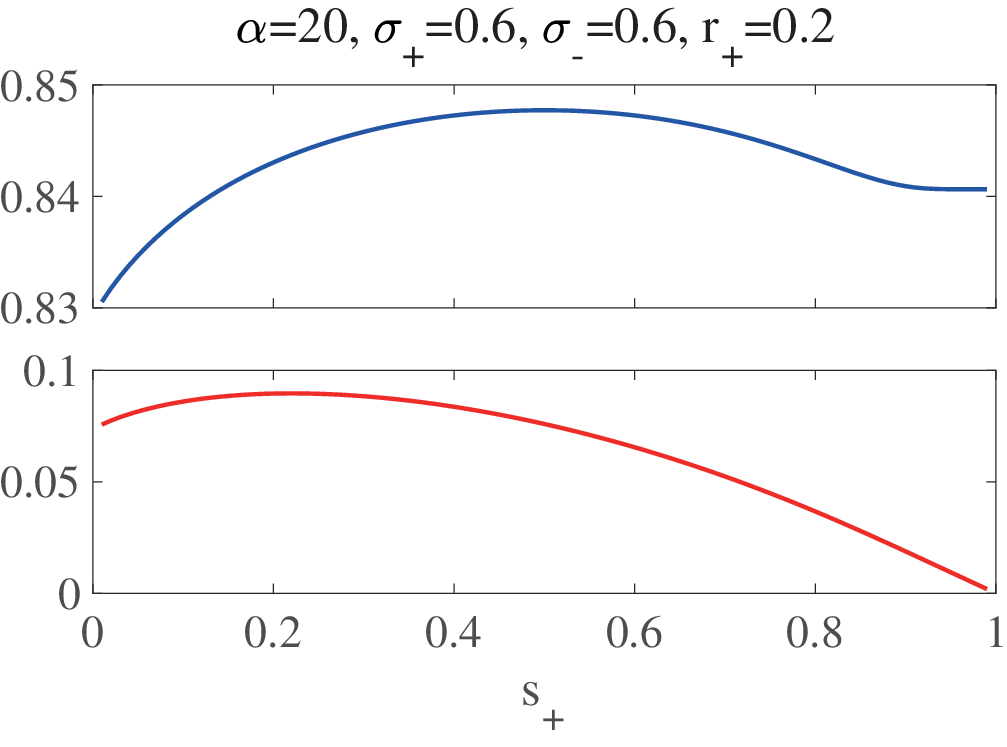}\\
(a) Zero-one loss with perceptron $\lossS_{\mathrm{01pe}}$.\\
\includegraphics[width=0.485\columnwidth]{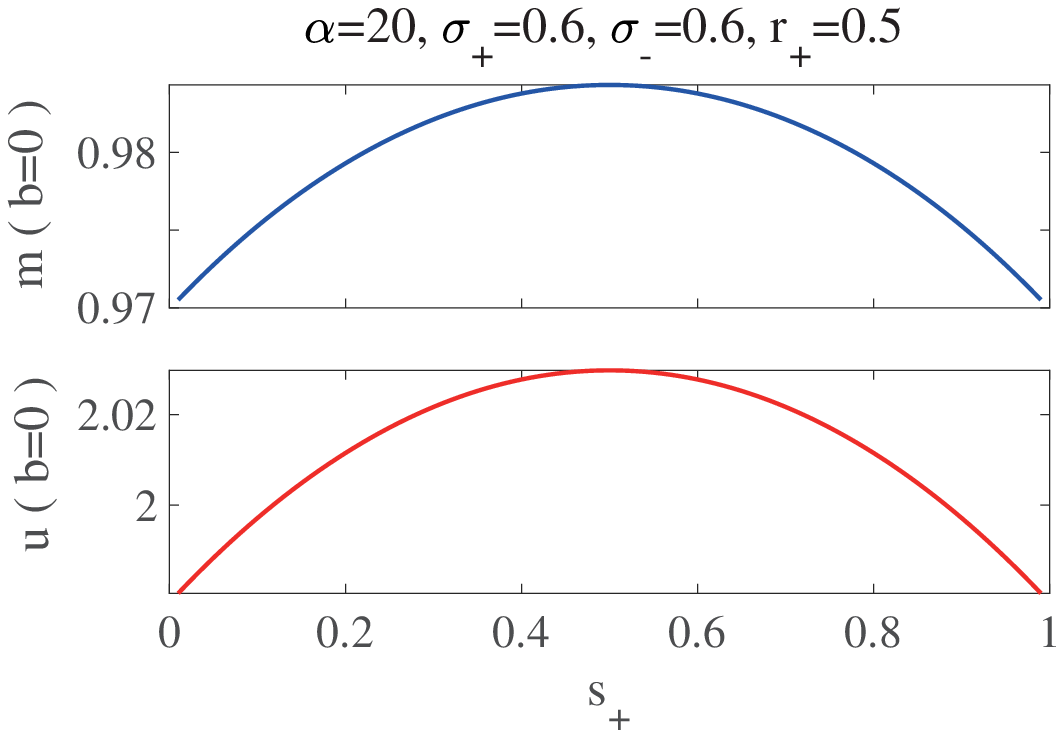}
\includegraphics[width=0.46\columnwidth]{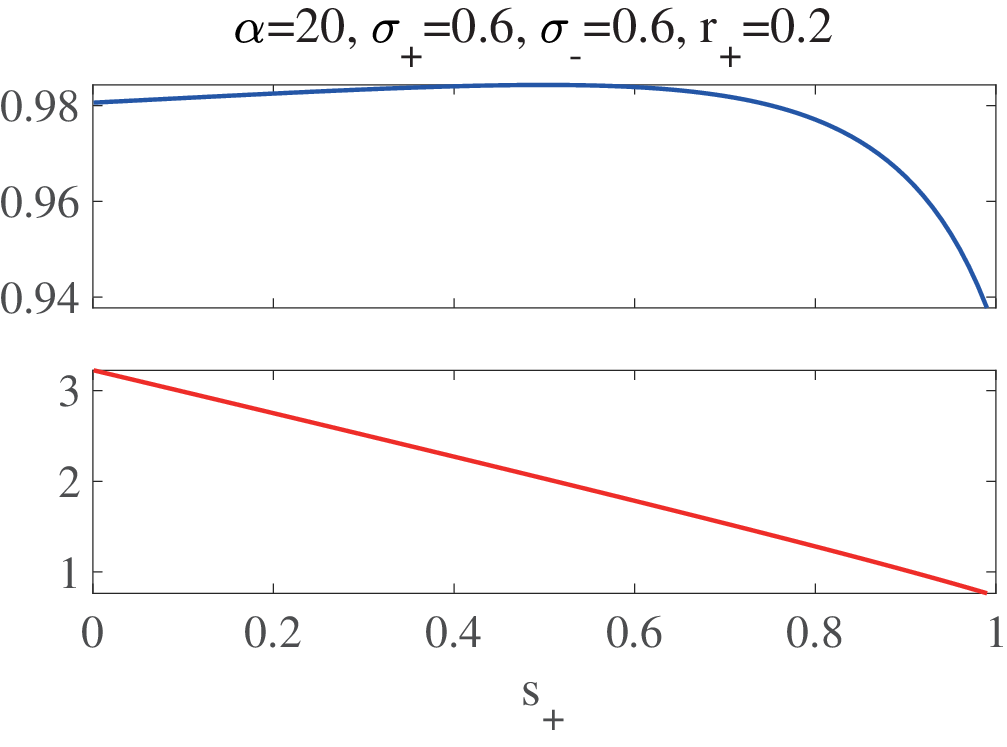}\\
(b) Cross-entropy loss with logistic function $\lossS_{\mathrm{CElo}}$.
\vspace{0mm}
\caption{Plots of $m$ and $u$ at $b=0$ against $s_+$ for $r_+=0.5$ (left) and $r_+=0.2$ (right). 
(a) Zero-one loss with perceptron $\lossS_{\mathrm{01pe}}$ and (b) Cross-entropy loss with logistic function $\lossS_{\mathrm{CElo}}$.
A crucial observation is that the maximum of $m$ is always obtained at the no reweighting situation $s_+=1/2$, irrespectively of the models, losses, and the degree of class imbalance.   }
\Lfig{b0_al20_sig06}
\end{center}
\end{figure}
%%%%%%%%%%%%%%%%%%%%%%%
A very interesting observation is that the maximum of $m$ is achieved at the no-reweighting situation $s_+=0.5$ in all the cases. Actually, this property holds irrespectively of the loss, model, or the degree of class imbalance: we will provide a strong analytical evidence of this fact later in \Rsec{Analytical evidence}. This property means that the feature learning has its best performance when no resampling/reweighting is applied if the bias is appropriately chosen to make the decision boundary to be located equidistantly from the two class centers. This provides an analytical support for Kang et al.'s observation and constitutes one of our main results in this paper. 

To summarize, we enumerate our findings in \Rsec{Equivariance case}: 
%%%%%%%%%
\begin{enumerate}
\item{The overlap value $m$ tends to be larger with CElo than with 01pe. }
\item{The values of $b$ with which maximum values of $m$ are achieved tend to be strongly dependent on the choice of the model and the loss.}
\item{The maximum overlap $m_{\rm max}=\max_bm(b)$ also shows a strong dependence on the choice of the model and the loss. With 01pe, $m_{\rm max}$ takes its largest value at the no-reweighting situation $s_+=0.5$, whereas with CElo it is obtained at some extreme values of $s_+$. }
\item{The overlap $m(u_{\mathrm{min}})$ at the point of minimizing the loss $u$ shows a moderate dependence on $s_+$. The dependence on $s_+$ is natural with 01pe but is counterintuitive with CElo. With CElo, the maximum is at $s_+=0.5$ in the balanced case but is at the extreme value of $s_+$ in the imbalanced case. }
\item{The overlap $m$ takes its maximum at the no-reweighting situation $s_+=0.5$ irrespectively of the loss, model, or degree of class imbalance when the bias is appropriately chosen to make the decision boundary to be located equidistantly from the two class centers.  }
\end{enumerate}
%%%%%%%%%
The assumption on the bias in item 5 of the above list is what is considered desirable in the standard view of classification. Hence, Kang et al.'s observation is a property that holds widely when features and bias are set to be in such a desirable situation, which conversely implies that their learning works well. Meanwhile, our other findings such as item 4 in the above suggest some other bias values different from the desirable one; the resultantly selected value tends to take an extraordinary value outside the reasonable range $ \lsb -1,1\rsb$ of the bias $b$ in the present setting. Presumably, this has prevented researchers from examining such extreme bias values in practical situations, and it may be an interesting future work to study such extreme biases in real-world datasets. 

%%%%%%%%%%%%%%%%%%%%%%%%%%%%%%%%%%%%%%%%%%%%%%%%%%%%%
\subsubsection{Nonequal-variance case}\Lsec{Non-equivariance case}
We turn to the nonequal-variance case. As an example, we examine $\sigma_+=1,\sigma_-=0.5$ with $r_+=0.5$ and $0.2$ as depicted in \Rfig{PDF_sigdiff}. 
%%%%%%%%%%%%%%%%%%%%%%%
\begin{figure}[htbp]
\begin{center}
  %\vspace{0mm}
  \begin{minipage}{0.45\columnwidth}
    \centering
    $\sigma_+=1$, $\sigma_-=0.5$, $r_+=0.5$\\
    \includegraphics[width=\columnwidth,clip,trim=0 48 0 38]{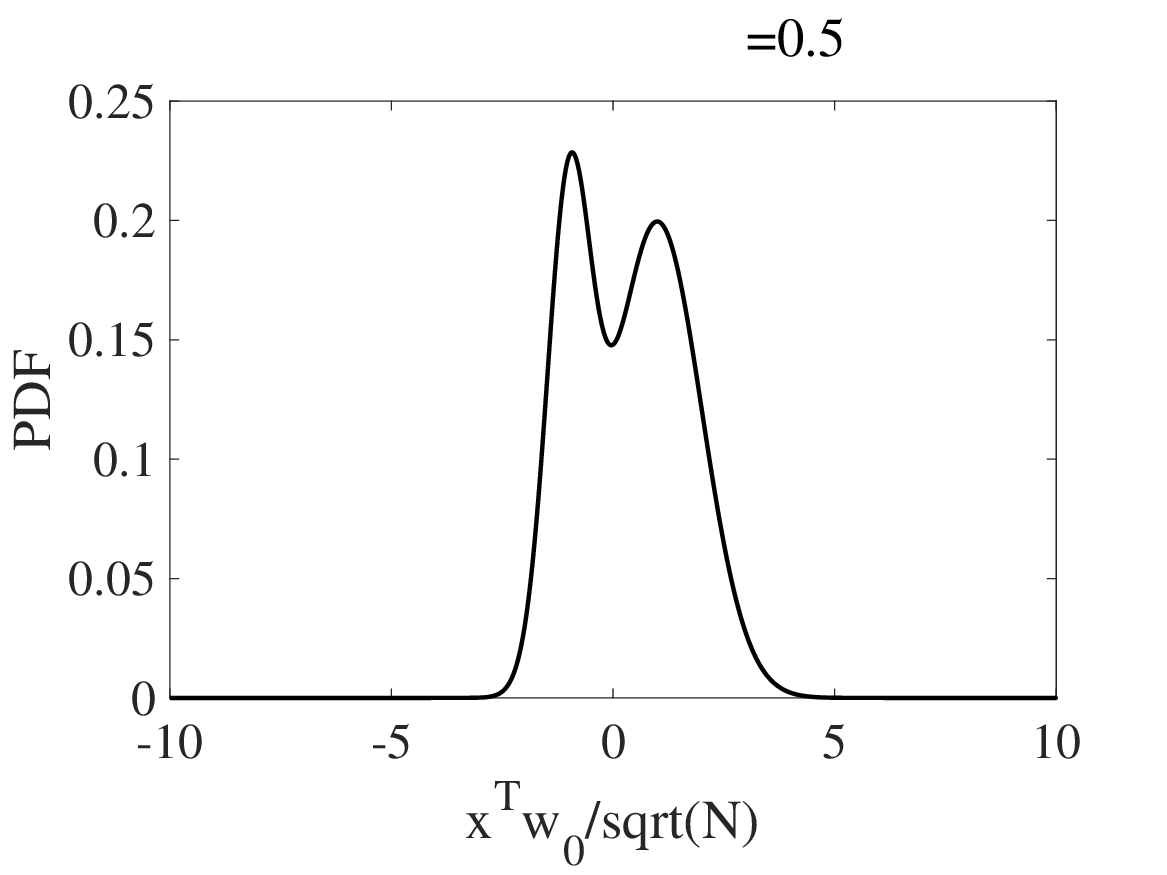}\\
    $\bm{w}_0^\top\bm{x}/\sqrt{N}$
  \end{minipage}
  \begin{minipage}{0.45\columnwidth}
    \centering
    $\sigma_+=1$, $\sigma_-=0.5$, $r_+=0.2$\\
    \includegraphics[width=\columnwidth,clip,trim=0 48 0 38]{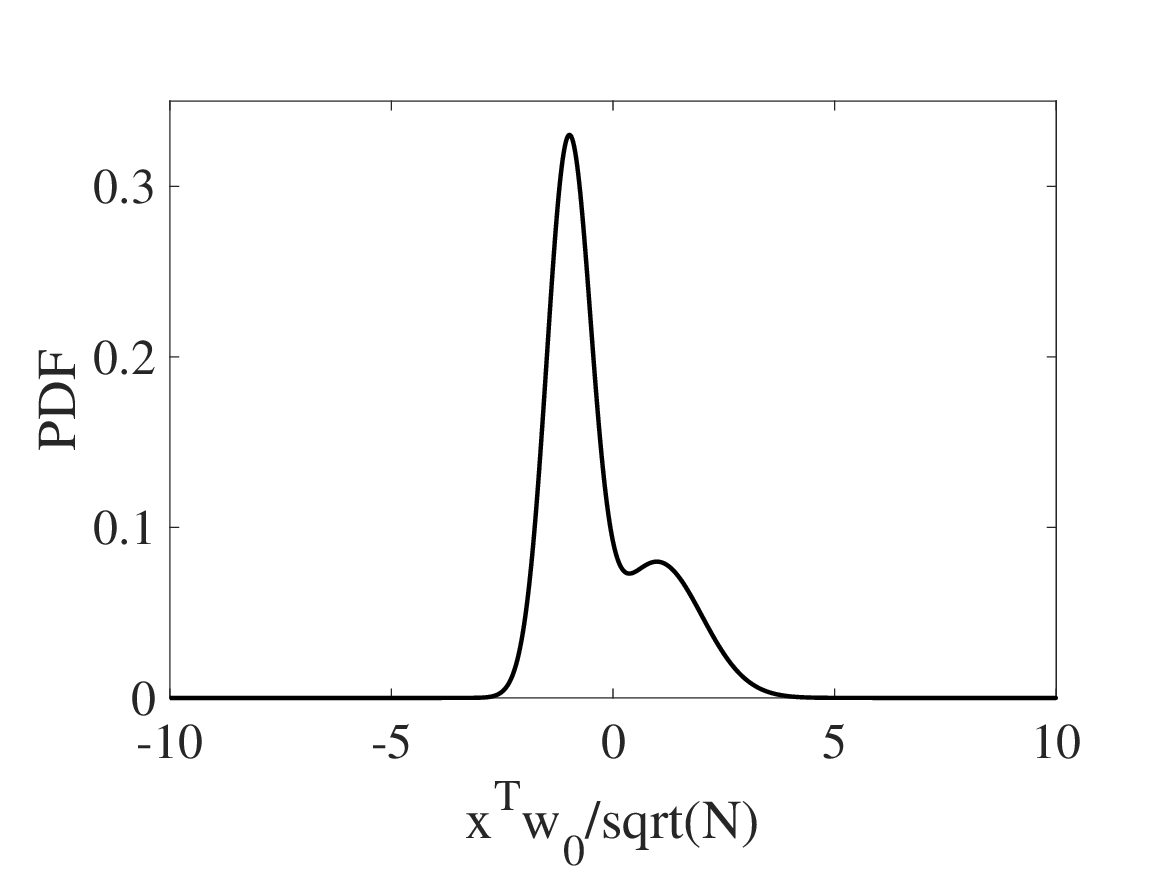}\\
    $\bm{w}_0^\top\bm{x}/\sqrt{N}$
  \end{minipage}
\vspace{0mm}
\caption{PDFs for $\sigma_{+}=1,\sigma_{-}=0.5$ at $r_+=0.5$ (left) and $r_+=0.2$ (right).} 
\Lfig{PDF_sigdiff}
\end{center}
\end{figure}
%%%%%%%%%%%%%%%%%%%%%%%
We compare the result for this case with the one in the equal-variance case, especially focusing on the reweighting factor dependence of $m$ after erasing the bias dependence as in \Rfigss{mmax_al20_sig06}{b0_al20_sig06}. 

Figure \NRfig{mmax_al20_sigp10_sigm05} is the counterpart of \Rfig{mmax_al20_sig06} in which $m_{\rm max}$ and the related quantities are plotted against $s_+$.
%%%%%%%%%%%%%%%%%%%%%%%
\begin{figure}[htbp]
\begin{center}
\vspace{0mm}
\includegraphics[width=0.485\columnwidth]{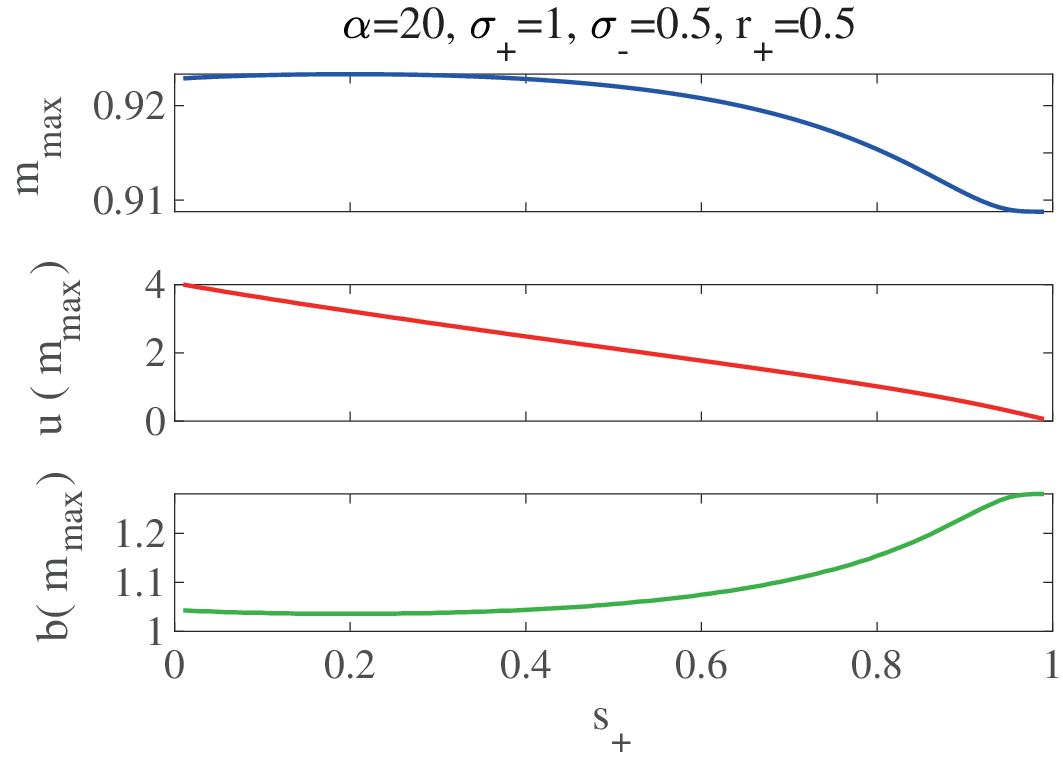}
\includegraphics[width=0.45\columnwidth]{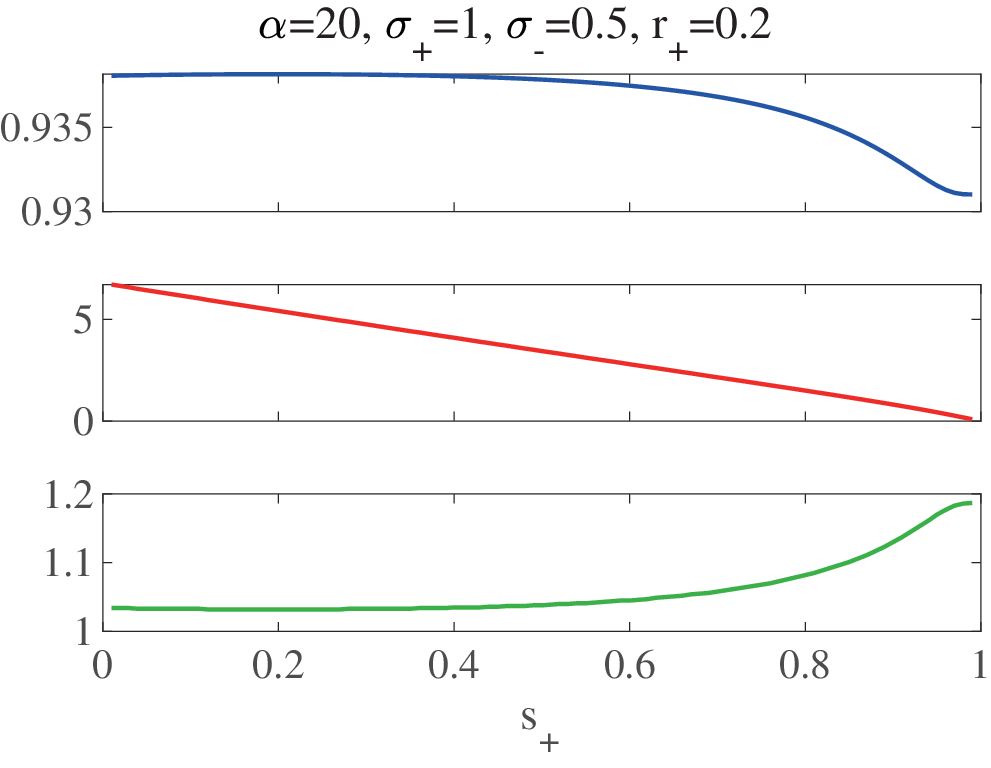}\\
(a) Zero-one loss with perceptron $\lossS_{\mathrm{01pe}}$.\\
\includegraphics[width=0.49\columnwidth]{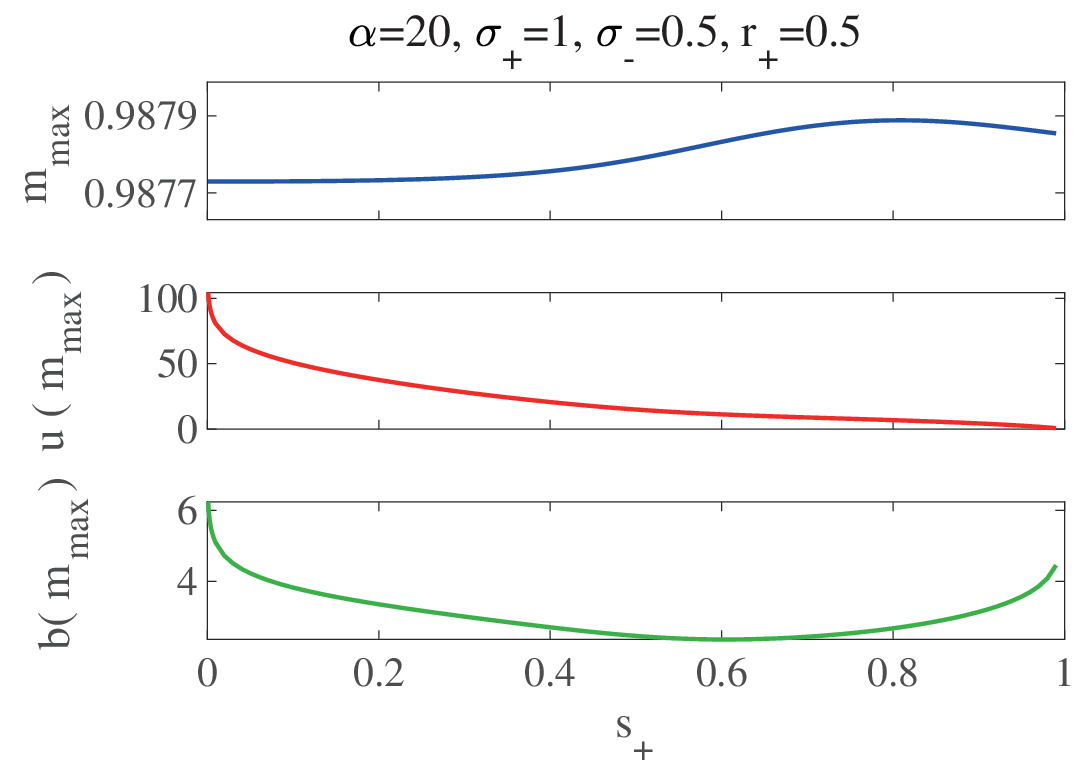}
\includegraphics[width=0.45\columnwidth]{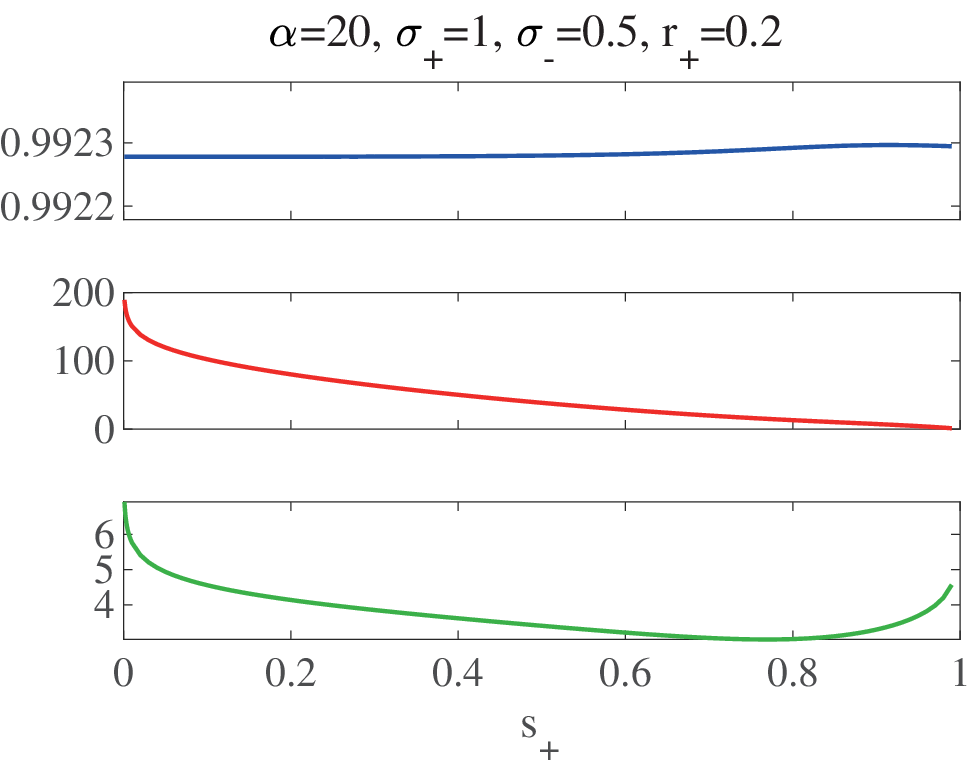}\\
(b) Cross-entropy loss with logistic function $\lossS_{\mathrm{CElo}}$.
\vspace{0mm}
\caption{Plots of $m_{\rm max},u(m_{\rm max}),b(m_{\rm max})$ against $s_+$ for $r_+=0.5$ (left) and $r_+=0.2$ (right) in the nonequal-variance case. (a) Zero-one loss with perceptron $\lossS_{\mathrm{01pe}}$. (b) Cross-entropy loss with logistic function $\lossS_{\mathrm{CElo}}$.} 
\Lfig{mmax_al20_sigp10_sigm05}
\end{center}
\end{figure}
%%%%%%%%%%%%%%%%%%%%%%%
As expected from the asymmetry between the classes, the maximum location of $m_{\rm max}$ is not $s_+=1/2$ anymore with either 01pe or CElo. Another interesting observation is that the maximum location is $s_+<1/2$ with 01pe and $s_+>1/2$ with CElo, showing that the effect of reweighting on the feature learning is not simple. 

Next, we study the loss-minimizing result in \Rfig{umin_al20_sigp10_sigm05}, which is the counterpart of \Rfig{umin_al20_sig06}.
%%%%%%%%%%%%%%%%%%%%%%%
\begin{figure}[htbp]
\begin{center}
\vspace{0mm}
\includegraphics[width=0.485\columnwidth]{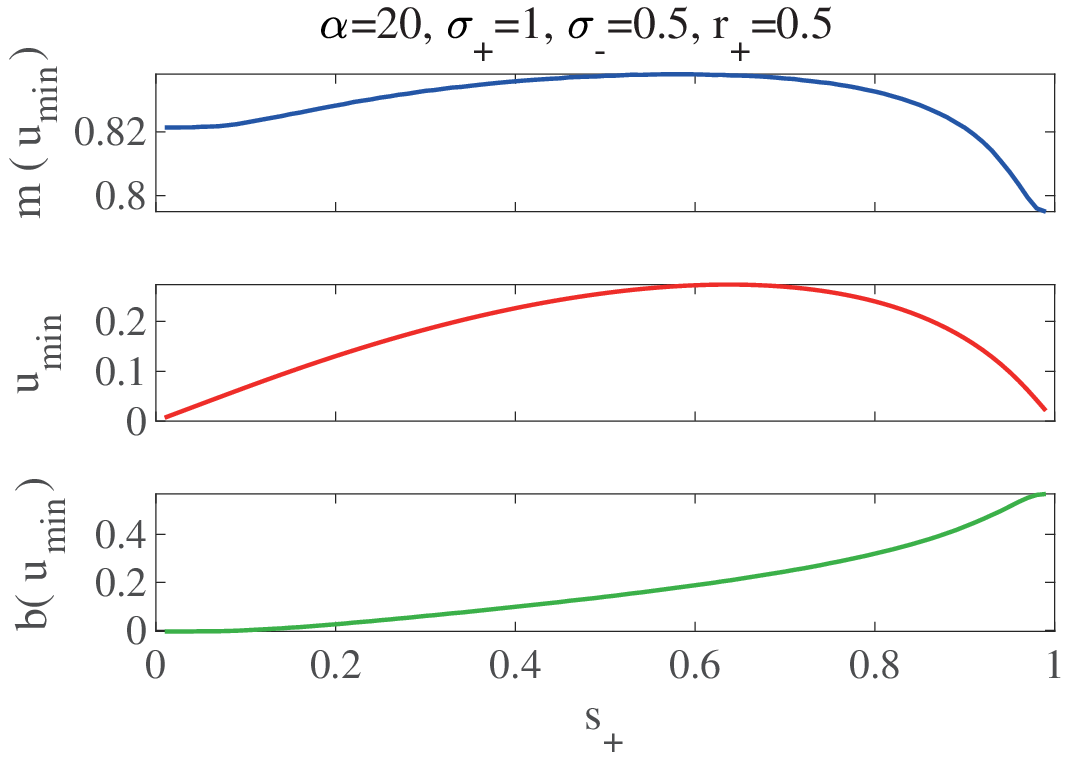}
\includegraphics[width=0.45\columnwidth]{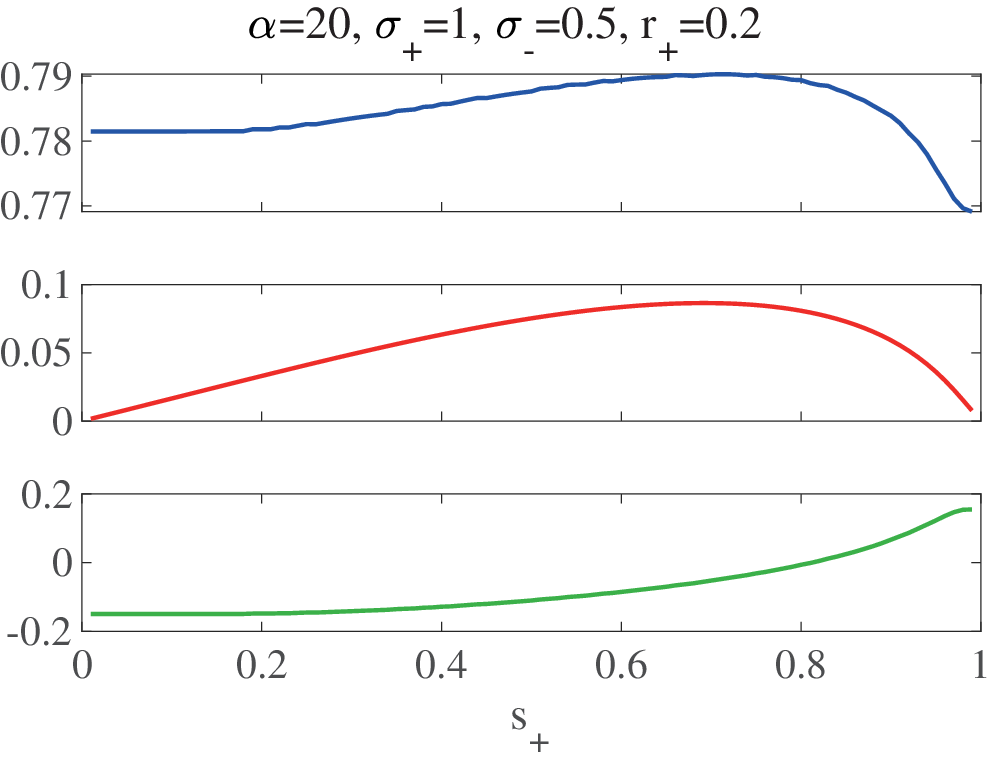}\\
(a) Zero-one loss with perceptron $\lossS_{\mathrm{01pe}}$.\\
\includegraphics[width=0.485\columnwidth]{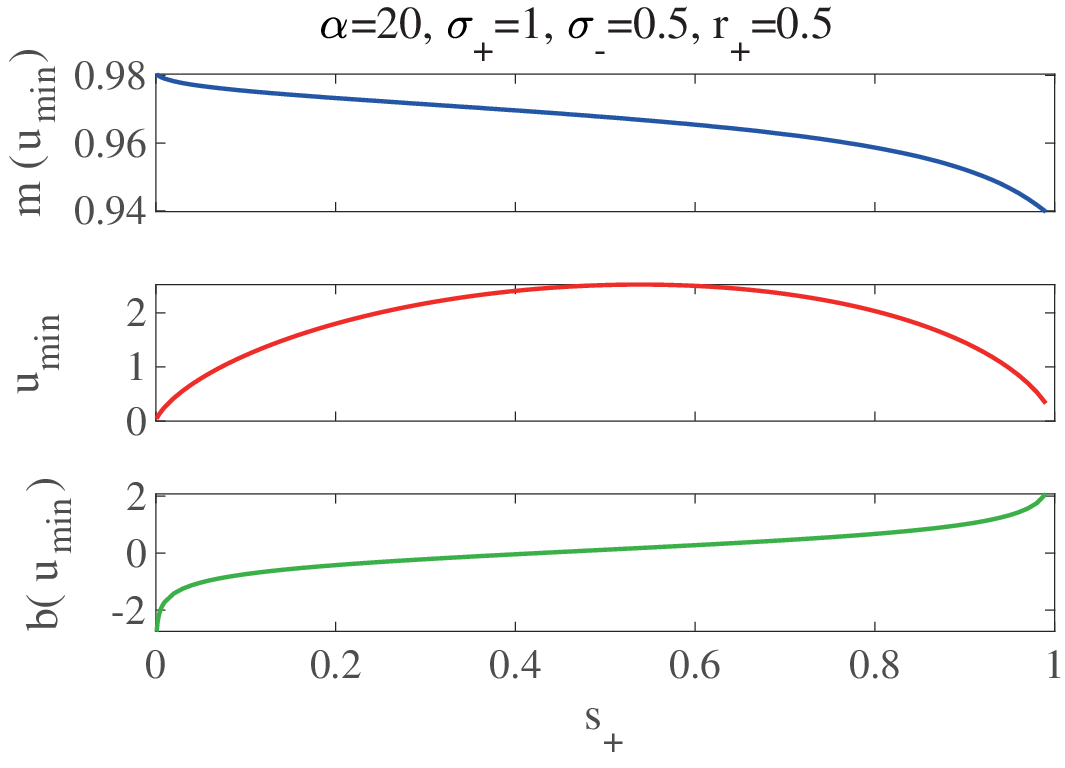}
\includegraphics[width=0.45\columnwidth]{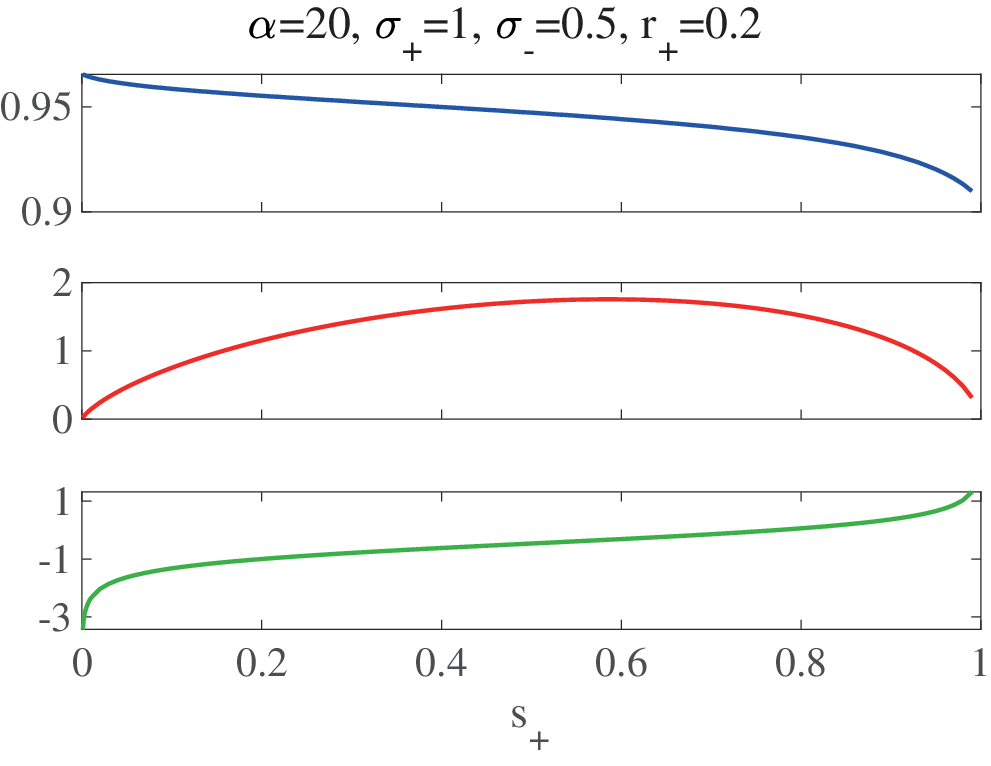}\\
(b) Cross-entropy loss with logistic function $\lossS_{\mathrm{CElo}}$.
\vspace{0mm}
\caption{Plots of $m(u_{\rm min}),u_{\rm min},b(u_{\rm min})$ against $s_+$ for $r_+=0.5$ (left) and $r_+=0.2$ (right) in the nonequal-variance case. (a) Zero-one loss with perceptron $\lossS_{\mathrm{01pe}}$. (b) Cross-entropy loss with logistic function $\lossS_{\mathrm{CElo}}$.} 
\Lfig{umin_al20_sigp10_sigm05}
\end{center}
\end{figure}
%%%%%%%%%%%%%%%%%%%%%%%
This time, again due to the asymmetry, the maximum location of $m$ is not at $s_+=1/2$ in all the cases. It is at $s_+>1/2$ with 01pe and at $s_+=0$ with CElo, which is in contrast to the overlap-maximizing result in \Rfig{mmax_al20_sigp10_sigm05}. Looking at both of the equal-variance and nonequal-variance results with CElo shown in \Rfigs{umin_al20_sig06}{umin_al20_sigp10_sigm05}, respectively, we see that the asymmetry due to the imbalance both in the number of examples and the variance magnitude commonly leads to extreme values of $s_+$ for the best feature learning. 

Finally, we examine the no-bias case $b=0$ in \Rfig{b0_al20_sigp10_sigm05}. 
%%%%%%%%%%%%%%%%%%%%%%%
\begin{figure}[htbp]
\begin{center}
\vspace{0mm}
\includegraphics[width=0.485\columnwidth]{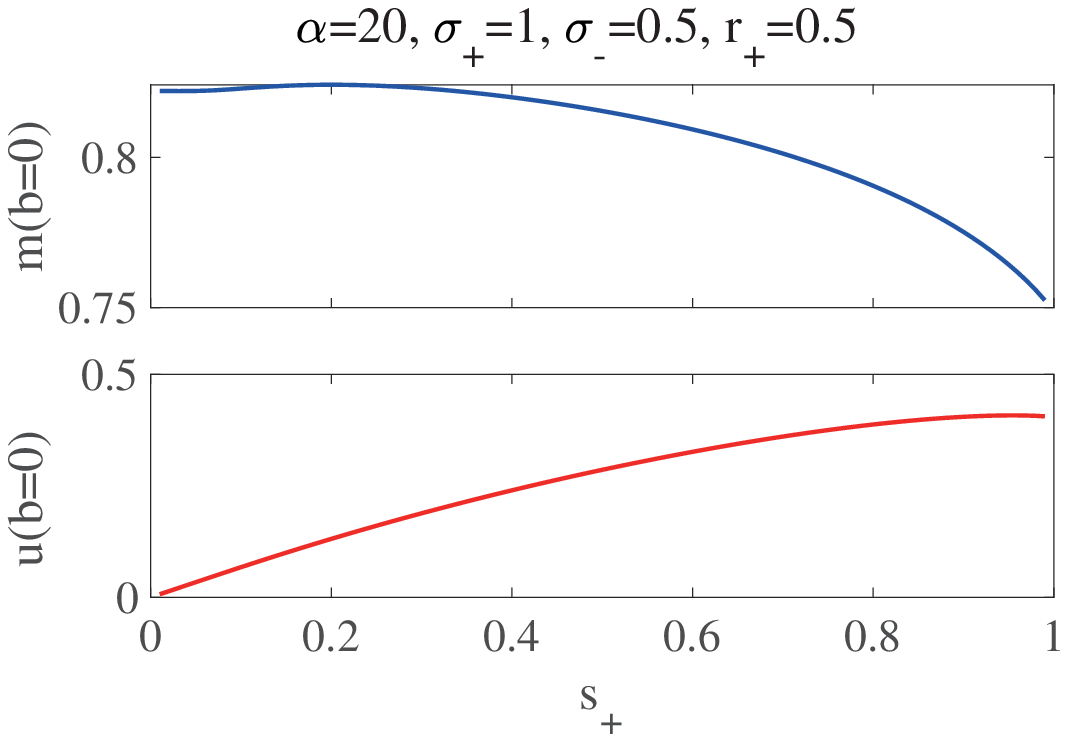}
\includegraphics[width=0.45\columnwidth]{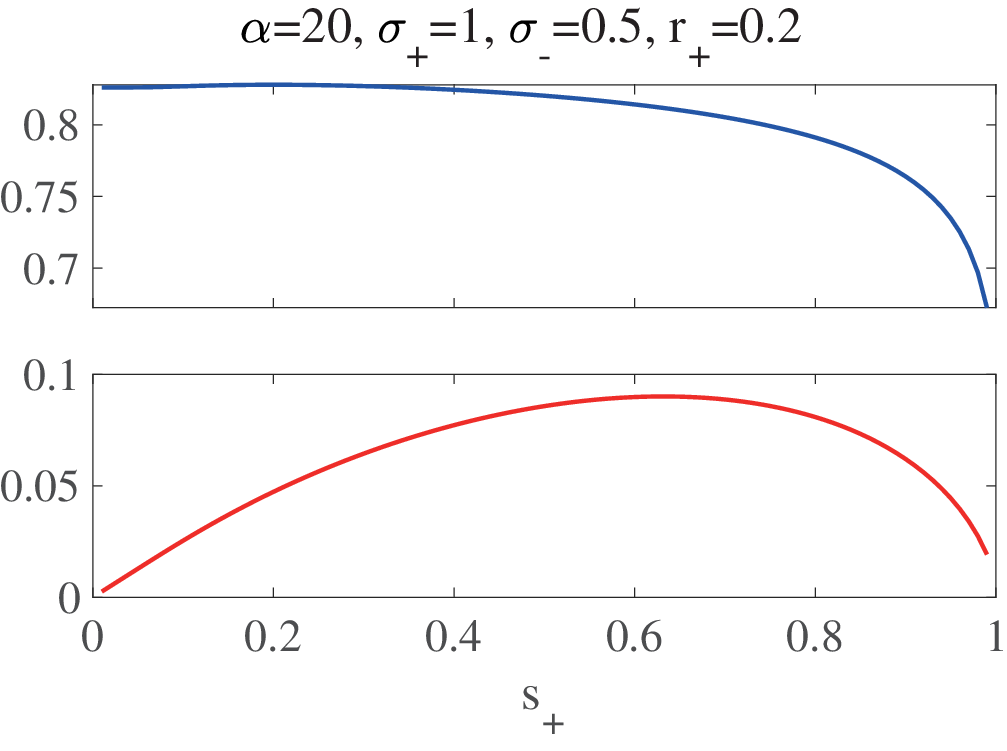}\\
(a) Zero-one loss with perceptron $\lossS_{\mathrm{01pe}}$.\\
\includegraphics[width=0.485\columnwidth]{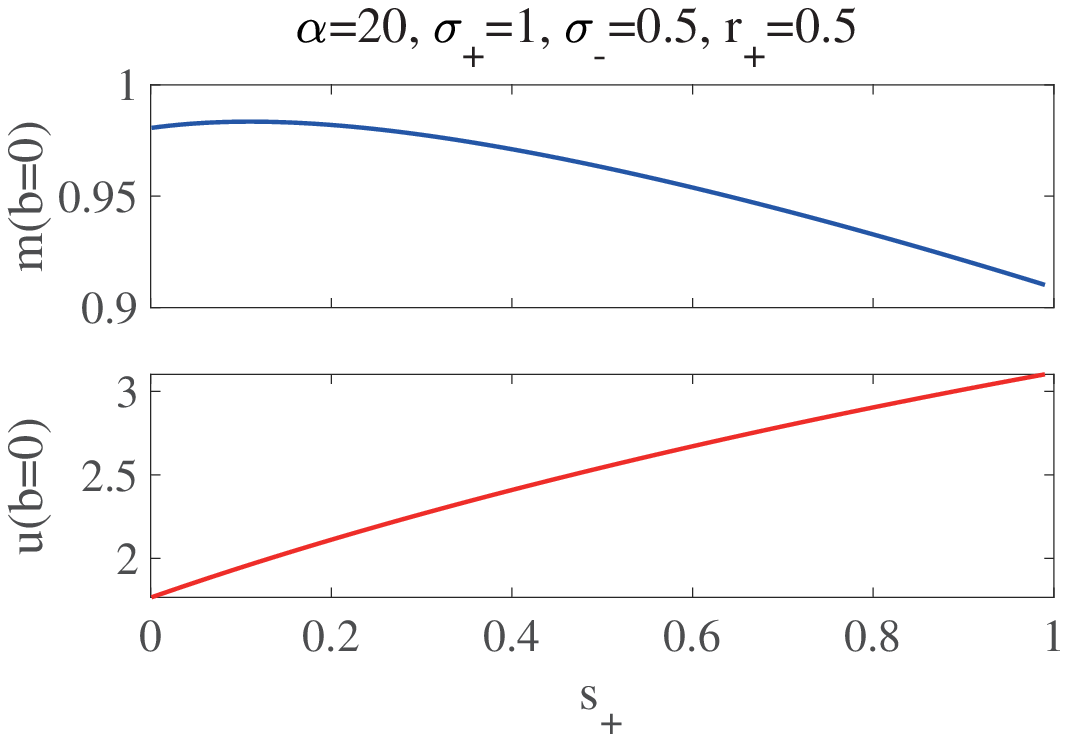}
\includegraphics[width=0.45\columnwidth]{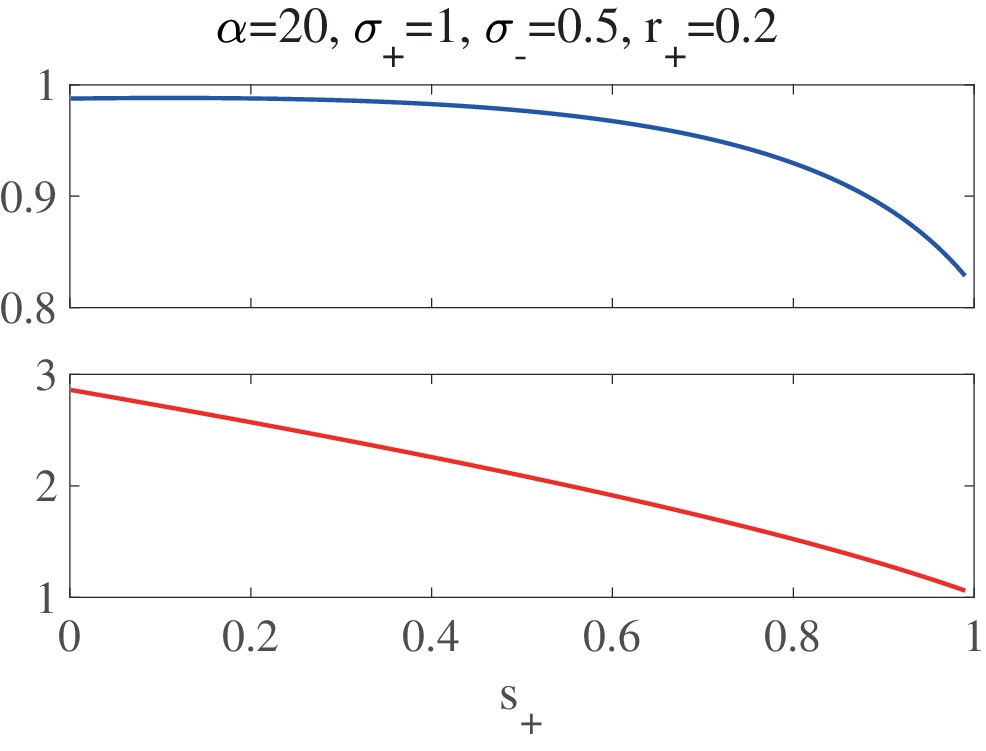}\\
(b) Cross-entropy loss with logistic function $\lossS_{\mathrm{CElo}}$.
\vspace{0mm}
\caption{Plots of $m(b=0),u(b=0)$ against $s_+$ for $r_+=0.5$ (left) and $r_+=0.2$ (right) in the nonequal-variance case. (a) Zero-one loss with perceptron $\lossS_{\mathrm{01pe}}$. (b) Cross-entropy loss with logistic function $\lossS_{\mathrm{CElo}}$. } 
\Lfig{b0_al20_sigp10_sigm05}
\end{center}
\end{figure}
%%%%%%%%%%%%%%%%%%%%%%%
Similarly to the above two cases, the maximum of $m$ is not at $s_+=1/2$. An interesting observation is that the maximum is located commonly in $s_+<1/2$. This may not be surprising since the class with $y=-1$ has a smaller variance and thus is considered to carry more information about the feature direction. 

Overall, in the nonequal-variance case, we found no empirical evidence supporting Kang et al.'s observation. This implies that the equal-variance condition of the feature across the two classes may be an important ingredient for their observation. 

%%%%%%%%%%%%%%%%%%%%%%%%%%%%%%%%%%%%%%%%%%%%%%%%%%%%%
%%%%%%%%%%%%%%%%%%%%%%%%%%%%%%%%%%%%%%%%%%%%%%%%%%%%%
\subsection{Analytical evidence that the maximum value of $m$ is achieved at $s_{+}=1/2$}\Lsec{Analytical evidence}
In this subsection, we provide an analytical evidence for the observation that $m$ takes its maximum at the no-reweighting situation $s_+=1/2$ irrespectively of the loss, model, or degree of class imbalance when the input distributions of the two classes are equal-variance and the bias is appropriately chosen to make the decision boundary to be located equidistantly from the two class centers. The last two conditions mean  $\sigma_+=\sigma_- (\eqqcolon \sigma)$ and $b=0$ in the present setting, and we assume them throughout this subsection.  

The idea is very simple. What we want to show is the following equality:
\be
\left. \frac{dm}{ds_+} \right|_{s_+=1/2}=0.
\Leq{m_extermum}
\ee
It means that no reweighting/resampling point $(s_+=1/2)$ is an extremum of $m$. In deriving this, we need to take the derivatives w.r.t.\ $s_+$ of the EOS \NReq{EOS}. Taking the derivative and rearranging the formulae using the simple relation between $\T{Q}$ and $(\T{m},\T{\chi})$ in \Req{Qt_sp}, we have the following set of linear equations among the first order derivatives:
\subbe
\be
&&
\frac{d}{ds_+}
\left(
\begin{array}{cc}
m
\\
\chi
\end{array}
\right)
=\T{T}
\frac{d}{ds_+}
\left(
\begin{array}{cc}
\T{m}
\\
\T{\chi}
\end{array}
\right),
\Leq{Ot2O}
\\ 
&&
\frac{d}{ds_+}
\left(
\begin{array}{cc}
\T{m}
\\
\T{\chi}
\end{array}
\right)
=
\V{c}
+
T
\frac{d}{ds_+}
\left(
\begin{array}{cc}
m
\\
\chi
\end{array}
\right).
\Leq{O2Ot}
\ee
\Leq{EOS_derivative}
\subee
The matrices $T,\T{T}$ and the vector $\V{c}$ have simple analytical expressions depending on the order parameters, the explicit forms of which are given in \BReqs{Ttilde}{EOS_diff_s} in \Rsec{Computational}. After lengthy but straightforward algebraic computations, we can show that \Req{EOS_derivative} implies \Req{m_extermum} under the above two conditions, without specifying details of the loss, the model, or the degree of class imbalance. This strongly supports the empirical observation that the maximum of $m$ is achieved when no resampling/reweighting is applied. %, achieving the purpose of this subsection.
The detailed computations are deferred to \Rsec{Computational}.

It can be seen from the algebras in \Rsec{Computational} that the symmetry $\ell(h,y)=\ell(-h,-y)$ explained in \Rsec{Problem} is important in deriving the above result, but not its details, implying the wide applicability of the obtained result.
%%%%%%%%%%%%%%%%%%%%%%%%%%%%%%%%%%%%%%%%%%%%%%%%%%%%%
%%%%%%%%%%%%%%%%%%%%%%%%%%%%%%%%%%%%%%%%%%%%%%%%%%%%%
\subsection{Relevance to DNNs}\Lsec{Relevance to}
Here, we discuss how our findings compare to those of Cao et al.\ or Kang et al.~\citep{cao2019learning,kang2019decoupling}.

First of all, we should keep in mind that in the experiments of Cao et al., as well as those of Kang et al., whatever backbone networks (bidirectional LSTM, variant of ResNet, etc.) were used as feature extractors, the final layer of the models are linear classifiers. Therefore, it is natural to consider that our input vector $\V{x}$ corresponds to the feature representation $\V{h}$ of the final layer in their models. Thus, if the distribution of $\V{h}$ has equal variance across classes, this would match with the assumptions underlying our findings. Although the class-wise distributions of $\V{h}$ have not been elucidated in their papers, it is not difficult to numerically check whether this point actually holds or not.
 
Moreover, a recent paper reported that an interesting phenomenon widely occurs in successful DNNs for classification~\citep{doi:10.1073/pnas.2015509117}. This phenomenon is called neural collapse (NC), and one of its important ingredients is  {\it within-class variation collapse}, meaning that the final-layer feature vectors $\V{h}$ of the same class converge to an identical vector as a result of learning. This implies that the equal variance assumption holds in the limit where the variance approaches zero. The mechanism why NC occurs is also understood from a simple model analysis~\citep{doi:10.1073/pnas.2103091118}, which suggests that this phenomenon universally occurs in sufficiently expressive DNNs with reasonable regularizations. Considering this, it is plausible that the equal-variance condition underlying our results may indeed hold in actual DNNs.

%%%%%%%%%%%%%%%%%%%%%%%%%%%%%%%%%%%%%%%%%%%%%%%%%%%%%
%%%%%%%%%%%%%%%%%%%%%%%%%%%%%%%%%%%%%%%%%%%%%%%%%%%%%
\subsection{A further simplified model with more than two classes}\Lsec{A further}
Can the results so far be extended to the case with multiple classes more than two? While extending our problem setup to multiple classes and analyzing it using the replica method is possible, such an analysis would be rather complicated and we avoid doing it in this paper. Instead, in this subsection, we consider a further simplified model allowing us to analyze the multiclass problem easily. 

From the discussion of the previous subsection, we understand that the symmetry in the loss and the problem setup is important to find the overlap maximum in the no-reweighting situation. On the basis of this understanding, reflecting the symmetry argument, we propose the following ``one-dimensional'' feature generation process for multiclass classification:
\be
\V{x}_{\mu}=t_{\mu}\frac{\V{w}_0}{\sqrt{N}}+\V{\xi}_{\mu},\quad \mu=1,2,\ldots,M,
\ee
where the true feature is again normalized as $\|\V{w}_0\|^2=N$. The label $t_{\mu}\in\mathbb{R}$ is assumed to take one of $K$ values, where $K$ corresponds to the number of classes, and $\V{\xi}_{\mu}$ is assumed to be a random variable following $\mc{N}(\V{0},\sigma^2 I_N)$ where $I_N$ is the identity matrix of dimension $N$. We further assume that $t_1,t_2,\ldots,t_M,\V{\xi}_1,\V{\xi}_2,\ldots,\V{\xi}_M$ are independent. Here, only one true feature vector $\V{w}_0$ governs the feature space, and thus the feature learning performance can be quantified in terms of the accuracy of estimating $\V{w}_0$ as in the case of binary classification discussed so far.
For the sake of simplicity of the analysis,
we consider, instead of a classification loss,
a simple \emph{unsupervised} loss for feature learning. 
Concretely, we consider $h_{\mu}=\V{w}^{\top}\V{x}_{\mu}$ and obtain the estimator by maximizing the variance of $\{h_{\mu}\}_{\mu=1}^M$ under the presence of a set of sample-wise reweighting factors $\{ s_{\mu}\}_{\mu=1}^{M}$ satisfying $s_{\mu}\geq 0$ and $\sum_{\mu=1}^{M}s_{\mu}=1$. The loss is formulated as 
\be
\mc{H}(\V{w})=-\sum_{\mu=1}^{M}s_{\mu}(h_{\mu}-\overline{h})^2=-\V{w}^{\top}A\V{w},
\Leq{H_simple}
\ee
where $\overline{\cdot}$ denotes the weighted mean with the reweighting factors $\{s_\mu\}_{\mu=1}^M$ (i.e., $\overline{f}=\sum_{\mu=1}^{M}s_{\mu}f_{\mu}$), and where 
\be
A
=\sum_{\mu=1}^{M}s_{\mu}(\V{x}_{\mu}-\overline{\V{x}})(\V{x}_{\mu}-\overline{\V{x}})^{\top}
=\overline{\V{xx}^\top}-\overline{\V{x}}\,\overline{\V{x}}^\top.
\ee
The minimizer of \Req{H_simple} under the normalization condition $\|\V{w}\|^2=N$ becomes our estimator of $\V{w}_0$. Namely, the estimator is given by the eigenvector of the largest eigenvalue of $A$. The matrix $A$ can be rewritten as
\be
A&=&\overline{\left(t\frac{\V{w}_0}{\sqrt{N}}+\V{\xi}\right)
  \left(t\frac{\V{w}_0}{\sqrt{N}}+\V{\xi}\right)^\top}
-\overline{\left(t\frac{\V{w}_0}{\sqrt{N}}+\V{\xi}\right)}\,
\overline{\left(t\frac{\V{w}_0}{\sqrt{N}}+\V{\xi}\right)}^\top
\\
&=&\tau\frac{\V{w}_0\lb\V{w}_0\rb^{\top}}{N}+\hat{\Sigma}+R,
\Leq{defA}
\ee
where
\be
&&
\tau=\overline{t^2}-(\overline{t})^2,
%\sum_{\mu=1}^{M}s_{\mu}t_{\mu}^2-\lb \sum_{\mu=1}^{M}s_{\mu}t_{\mu} \rb^2
\\ &&
\hat{\Sigma}=\overline{\V{\xi}\V{\xi}^{\top}}-\overline{\V{\xi}}\, \overline{\V{\xi}}^{\top},
\\ &&
R=
\lb \overline{t\V{\xi}}-\overline{t}\, \overline{\V{\xi}} \rb\frac{\V{w}_0^{\top}}{\sqrt{N}}
+
 \frac{\V{w}_0}{\sqrt{N}}\lb \overline{t\V{\xi}}-\overline{t}\, \overline{\V{\xi}} \rb^{\top}
.
%\lbb 
%\sum_{\mu=1}^{M}s_{\mu}t_{\mu}\V{\xi}_{\mu}
%-
%\lb \sum_{\mu=1}^{M}s_{\mu}t_{\mu}\rb 
%\lb \sum_{\mu=1}^{M}s_{\mu}\V{\xi}_{\mu} \rb
%\rbb\V{w}_0^{\top}+(\rm transpose).
\ee
The last term $R$ on the right-hand side of \Req{defA} is mean zero and is negligible if $M$ is large enough. Neglecting this term, we have our estimator as
\be
\hat{\V{w}} = \argmax_{\V{w}:\|\V{w}\|^2=N}\lbb \V{w}^{\top}\lb \tau \frac{\V{w}_0}{\sqrt{N}}\lb\frac{\V{w}_0}{\sqrt{N}}\rb^{\top}+\hat{\Sigma}\rb\V{w} \rbb.
\ee
We furthermore assume that the reweighting factors $\{s_\mu\}_{\mu=1}^M$
are given depending on $M$ and satisfy $\lim_{M\to\infty}\sum_{\mu=1}^Ms_\mu^2=0$. 
The empirical covariance $\hat{\Sigma}$ then converges to $\sigma^2 I$ as $M \to \infty$. Hence, in the large-$M$ limit, the maximizer $\hat{\V{w}}$, which is the eigenvector associated with the leading eigenvalue of $A$, approaches $\V{w}_0$ irrespectively of the choice of $\{s_{\mu} \}_{\mu=1}^{M}$, yielding $m\to1$. Meanwhile, for large but finite $M$, stochastic fluctuations of $\hat{\Sigma}$ make $\hat{\V{w}}$ to deviate from $\V{w}_0$, resulting in decrease of $m$ from 1. A detailed analysis reveals that minimizing the fluctuations leads to the best estimate of $\V{w}_0$, which is achieved at the no resampling/reweighting situation $s_{\mu}=1/M, \forall\mu$ on average. This can be easily shown by computing the mean squared deviation of the diagonal $\hat{\Sigma}_{ii}$ from $\sigma^2$, as 
\be
\E_{\V{\xi}}\lb \hat{\Sigma}_{ii}-\sigma^2\rb^2=\sigma^4
\lbb 
3\lb \sum_{\mu=1}^{M}s_{\mu}^2\rb^2 
-4\sum_{\mu=1}^{M}s_{\mu}^3 
+
2\sum_{\mu=1}^{M}s_{\mu}^2
\rbb.
\ee
Minimization of the right-hand side under the constraint $\sum_{\mu=1}^{M}s_{\mu}=1$ yields the uniform weights $s_{\mu}=1/M, \forall\mu$, implying that no resampling/reweighing leads to the best feature learning performance. This provides a simple demonstration supporting Kang et al.'s observation in the multiclass case. 

%%%%%%%%%%%%%%%%%%%%%%%%%%%%%%%%%%%%%%%%%%%%%%%%%%%%%
%%%%%%%%%%%%%%%%%%%%%%%%%%%%%%%%%%%%%%%%%%%%%%%%%%%%%
%%%%%%%%%%%%%%%%%%%%%%%%%%%%%%%%%%%%%%%%%%%%%%%%%%%%%
\section{Numerical experiments}\Lsec{Numerical}
In this section, we conduct numerical experiments to verify the correctness of the theoretical analysis presented in \Rsecss{Derivation of}{Behaviors of}. Our theoretical framework is built upon a method that assumes $N \to \infty$. Thus, if our numerical results, computed with sufficiently large $N$, align well with the theoretical findings, this would support the validity of our analysis. For this purpose, we only examined the case with CElo in this section since the CE loss is convex and thus the numerical optimization is relatively easy; such a good property is absent in the zero-one loss. The standard interior-point method was used for the optimization. In the following results, we conducted simulations with $N=400$ and took a sample average over $100$ different realizations of the dataset; the error bar is the standard error in the average. The parameter $\alpha$ was fixed to $\alpha=20$, which is identical to the value used in \Rsec{Behaviors of}.

%%%%%%%%%%%%%%%%%%%%%%%%%%%
\paragraph{Equal-variance case} We first experimented the equal-variance case $\sigma_+=\sigma_-=0.6$. We start from plotting $m$ and $u$ against $b$ for $r_+=0.5,0.2$ and $s_+=0.1,0.5$ in \Rfig{comp_orderparam_al20_sig06}. 
%%%%%%%%%%%%%%%%%%%%%%%
\begin{figure}[htbp]
\begin{center}
\includegraphics[width=0.45\columnwidth]{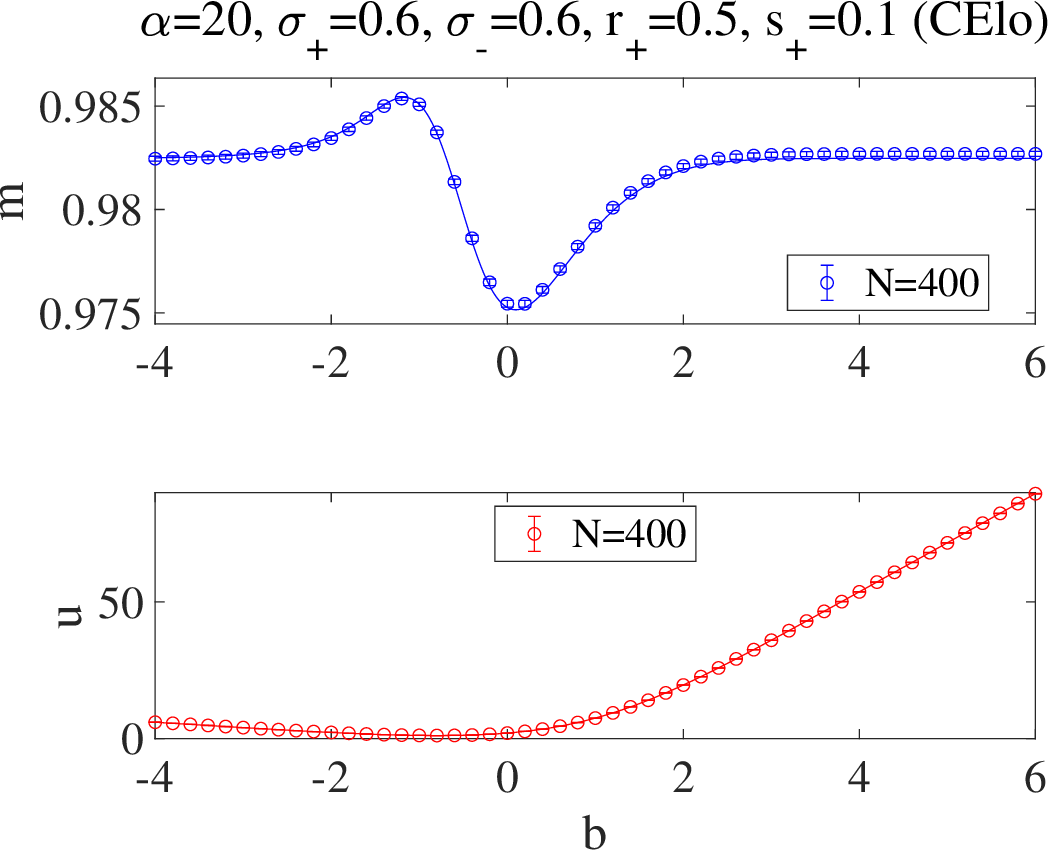}
\includegraphics[width=0.45\columnwidth]{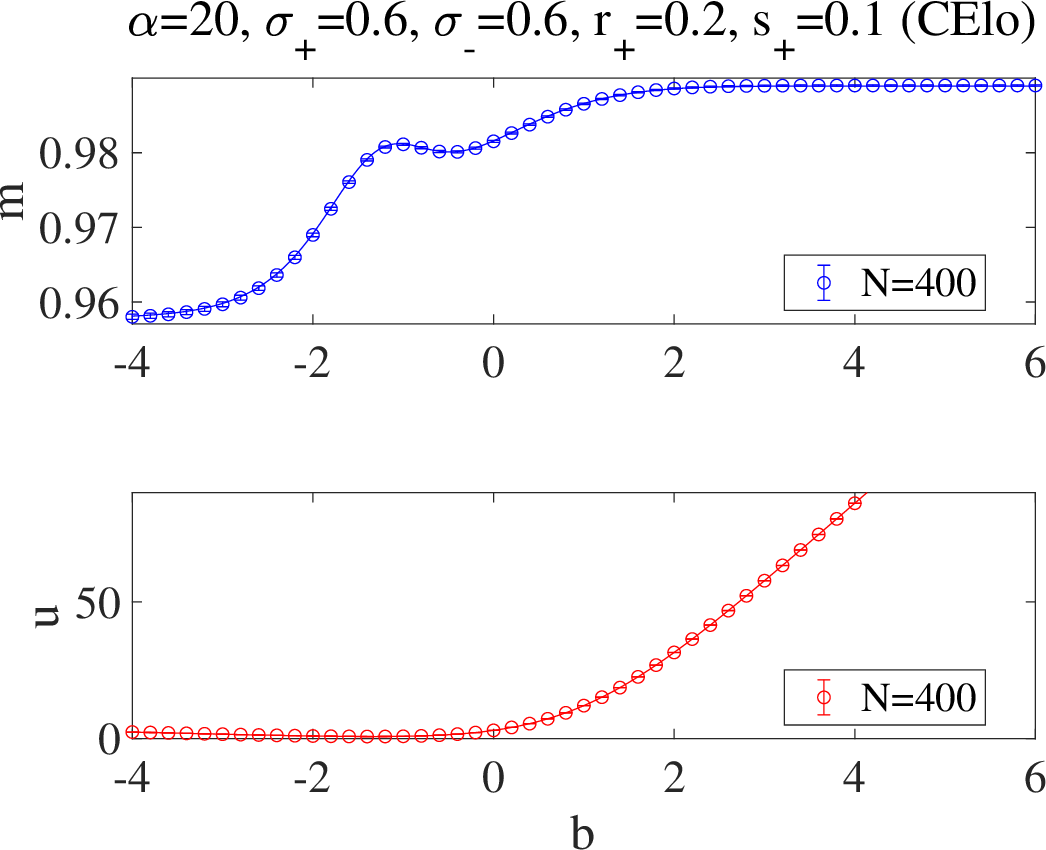}\\
(a) $s_+=0.1$ case.\\ \vspace{2mm}
\includegraphics[width=0.45\columnwidth]{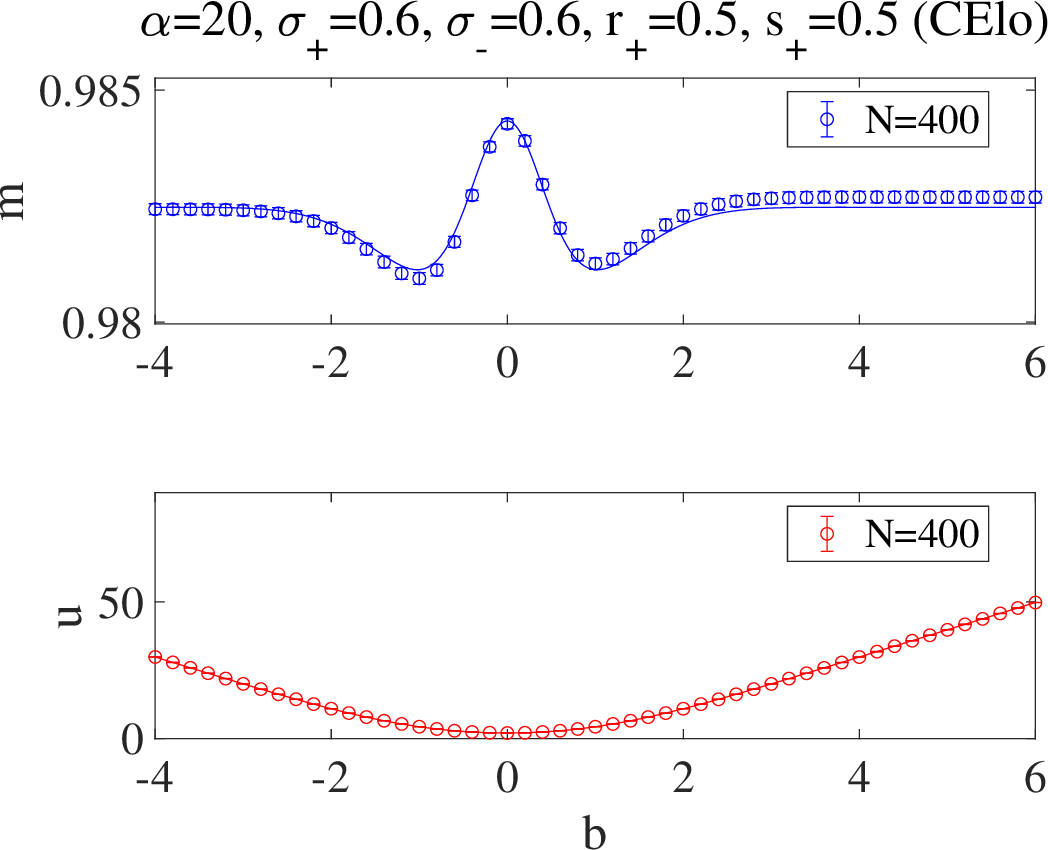}
\includegraphics[width=0.45\columnwidth]{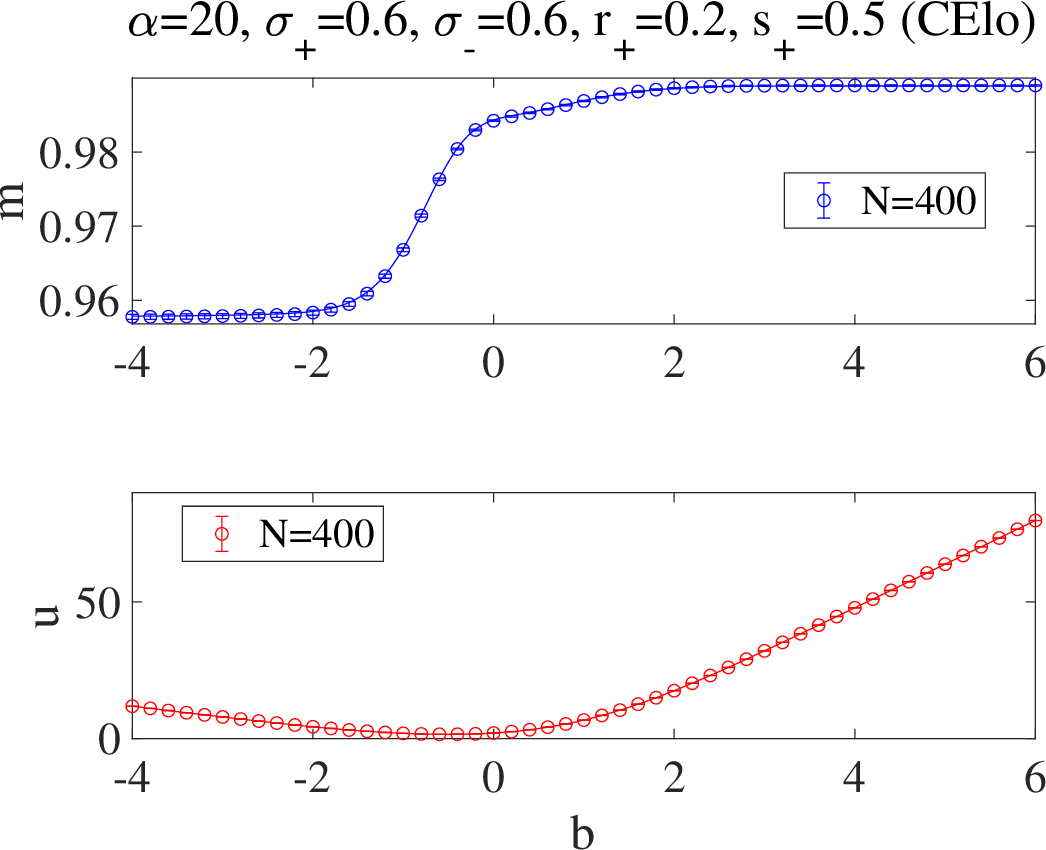}\\
(b) $s_+=0.5$ case.
\caption{Plots of $m$ and $u$ against $b$ for $r_+=0.5$ (left) and $0.2$ (right) at the equal-variance case $\sigma_+=\sigma_-=0.6$ for (a) $s_+=0.1$ and (b) $s_+=0.5$. The markers denote the numerical result and the lines represent the theoretical one. The agreement between them is fairly good. 
 } 
\Lfig{comp_orderparam_al20_sig06}
\end{center}
\end{figure}
%%%%%%%%%%%%%%%%%%%%%%%
The agreement between the numerical results (markers) and the theoretical ones (lines) is excellent, justifying our theoretical treatment. Although our theoretical analysis assumes the high-dimensional limit $N\to \infty$, this numerical result indicates that several hundreds of $N$ can be regarded large enough. 

For further quantification, we evaluated the loss-minimizing bias value and the corresponding $u_{\rm min}$ and $m(u_{\rm min})$ from the numerical experiments. %A simultaneous plot of them
Their plots, along with the theoretically-evaluated curves, are 
given in \Rfig{comp_umin_al20_sig06}.  
%%%%%%%%%%%%%%%%%%%%%%%
\begin{figure}[htbp]
\begin{center}
\vspace{0mm}
\includegraphics[width=0.485\columnwidth]{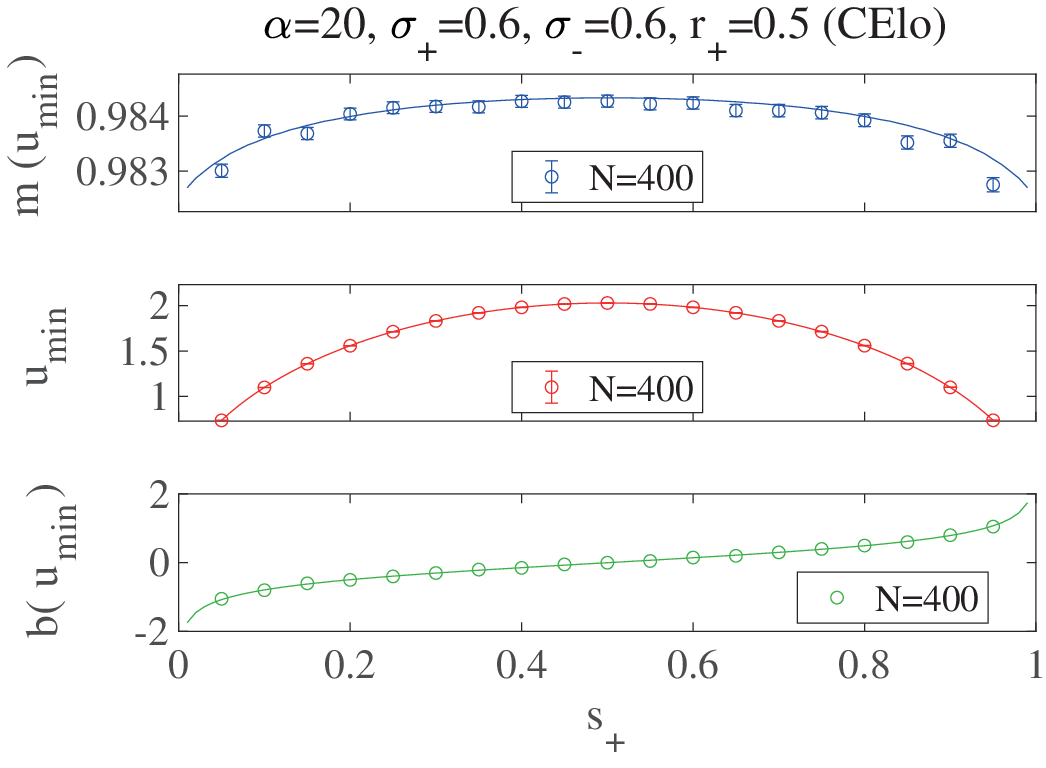}
\includegraphics[width=0.46\columnwidth]{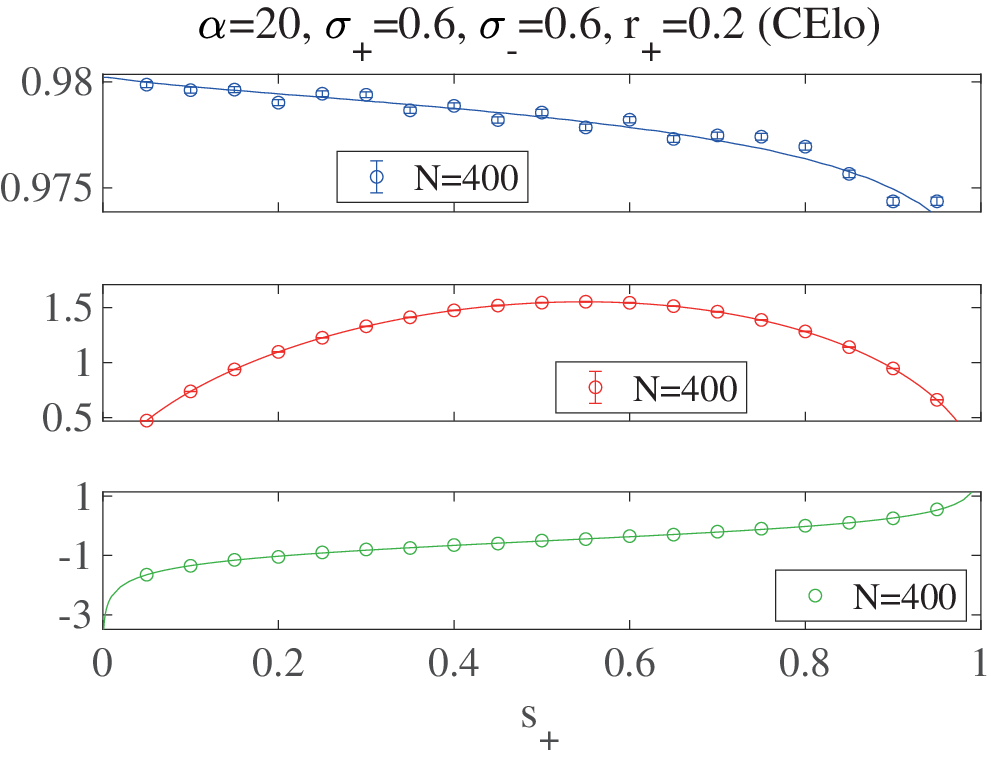}
\vspace{0mm}
\caption{Plots of $m(u_{\rm min})$ and $u_{\rm min}$ against $s_+$ for $r_+=0.5$ (left) and $0.2$ (right) at the equal-variance case $\sigma_+=\sigma_-=0.6$. The agreement between the theoretical and numerical results is again very good.  } 
\Lfig{comp_umin_al20_sig06}
\end{center}
\end{figure}
%%%%%%%%%%%%%%%%%%%%%%%
The agreement between the numerical and theoretical results is again very good. Even the nontrivial location of the maximum point of $m(u_{\rm min})$ is reproduced by the numerical experiments. 

The last result for the equal-variance case is the plot at $b=0$: $m(b=0)$ and $u(b=0)$ are plotted against $s_+$ with the theoretical curves in \Rfig{comp_b0_al20_sig06}. 
%%%%%%%%%%%%%%%%%%%%%%%
\begin{figure}[htbp]
\begin{center}
\vspace{0mm}
\includegraphics[width=0.486\columnwidth]{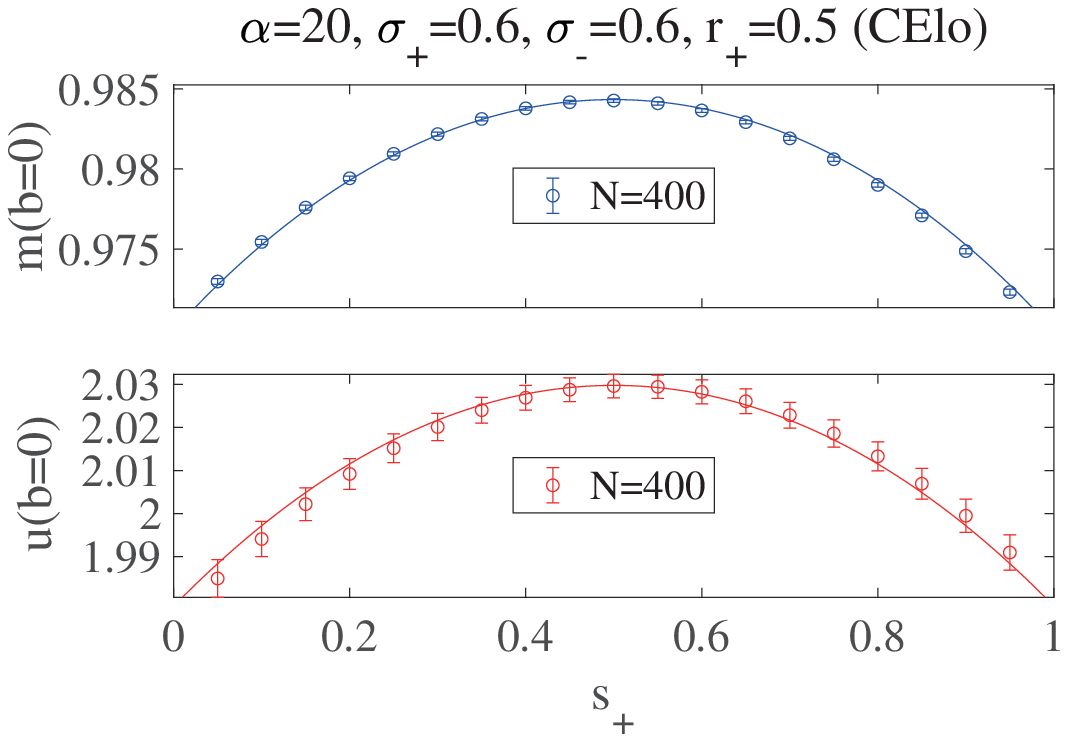}
\includegraphics[width=0.46\columnwidth]{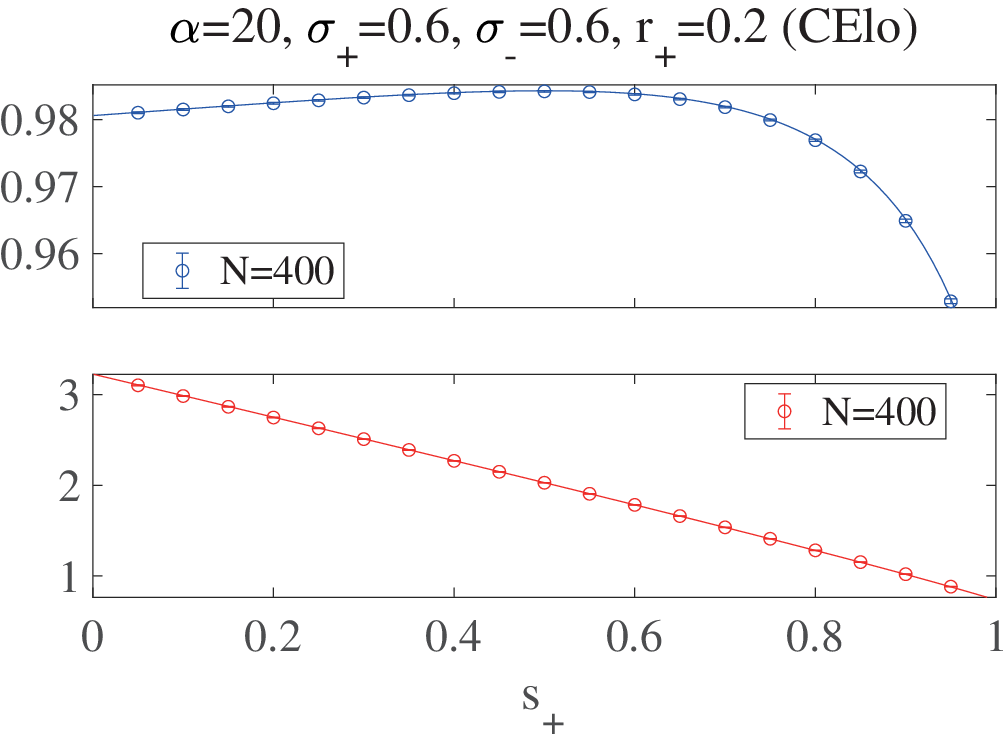}
\vspace{0mm}
\caption{Plots of $m(b=0)$ and $u(b=0)$ against $s_+$ for $r_+=0.5$ (left) and $0.2$ (right) at the equal-variance case $\sigma_+=\sigma_-=0.6$. } 
\Lfig{comp_b0_al20_sig06}
\end{center}
\end{figure}
%%%%%%%%%%%%%%%%%%%%%%%
The agreement is again good, and the maximum of $m(b=0)$ is approximately obtained at the no resampling/reweighting case $s_+=1/2$, numerically supporting our main result in this paper. 

%%%%%%%%%%%%%%%%%%%%%%%%%%%
\paragraph{Nonequal-variance case} We next experimented the nonequal-variance case $\sigma_+=1,\sigma_-=0.5$, and show the results briefly. The plots of $m$ and $u$ against $b$ for $r_+=0.5$ and $0.2$ are shown in \Rfig{comp_orderparam_al20_sigp10_sigm05}. 
%%%%%%%%%%%%%%%%%%%%%%%
\begin{figure}[htbp]
\begin{center}
\vspace{0mm}
\includegraphics[width=0.45\columnwidth]{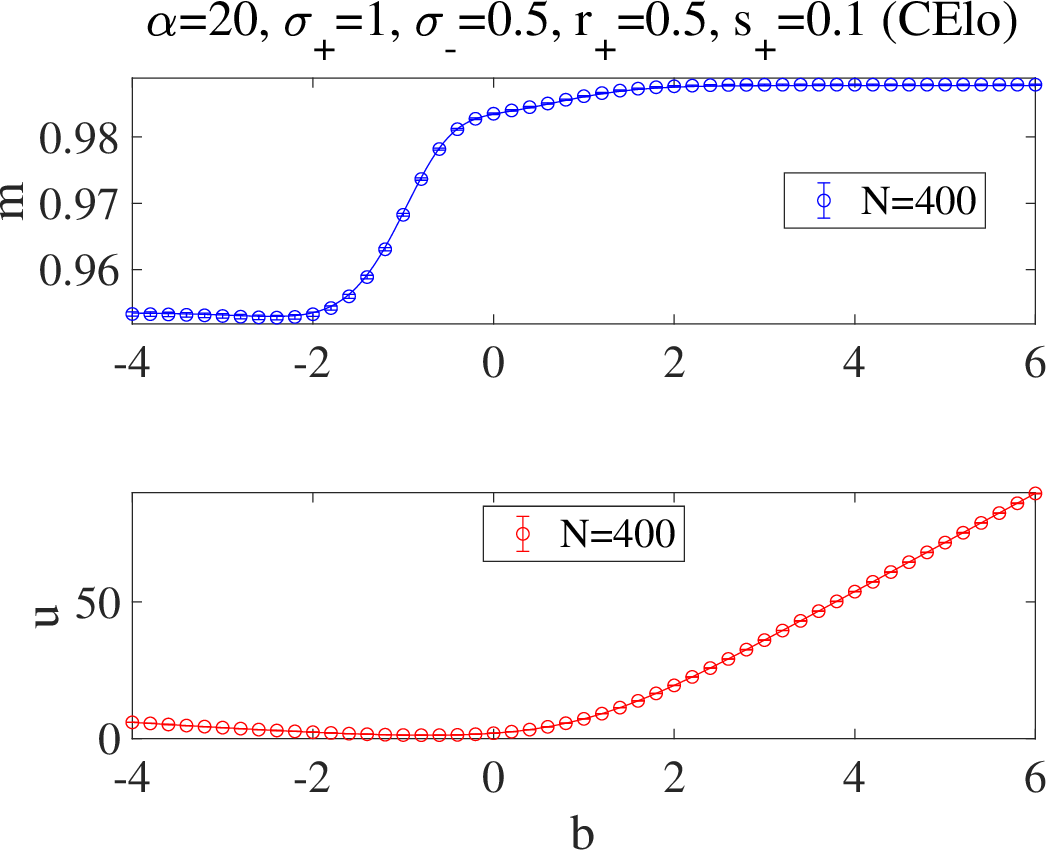}
\includegraphics[width=0.45\columnwidth]{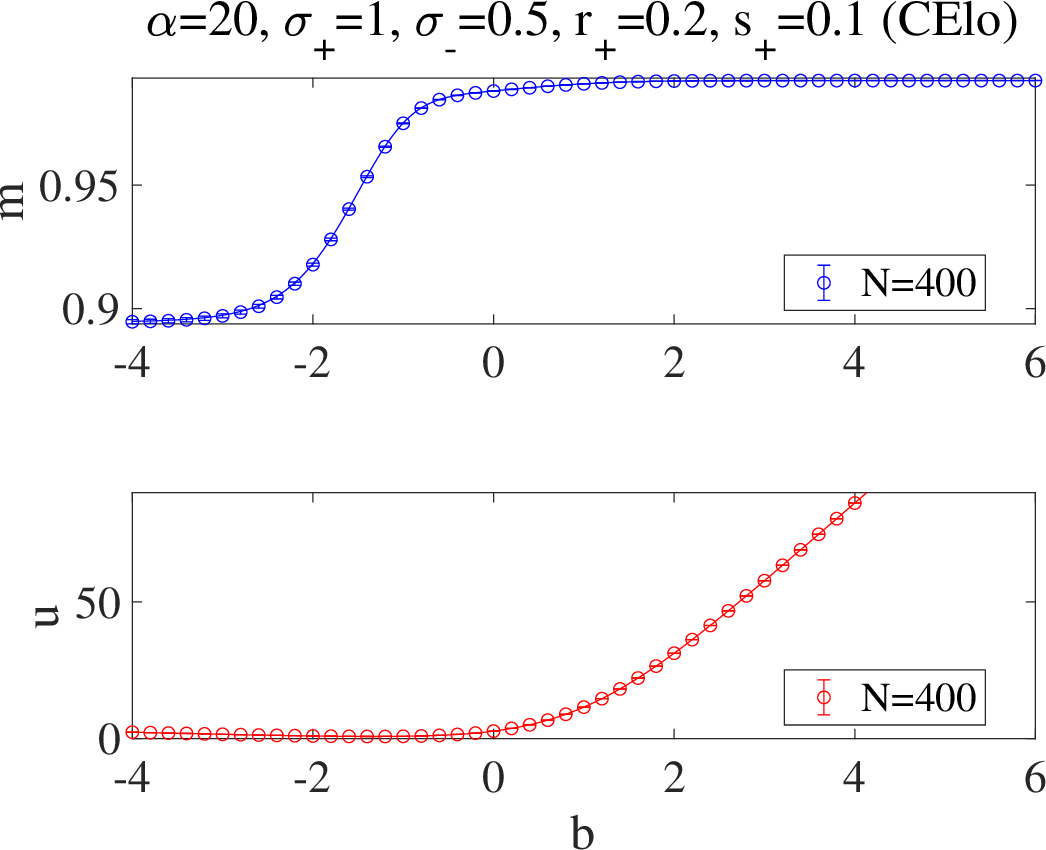}\\
(a) $s_+=0.1$ case.\\ \vspace{2mm}
\includegraphics[width=0.45\columnwidth]{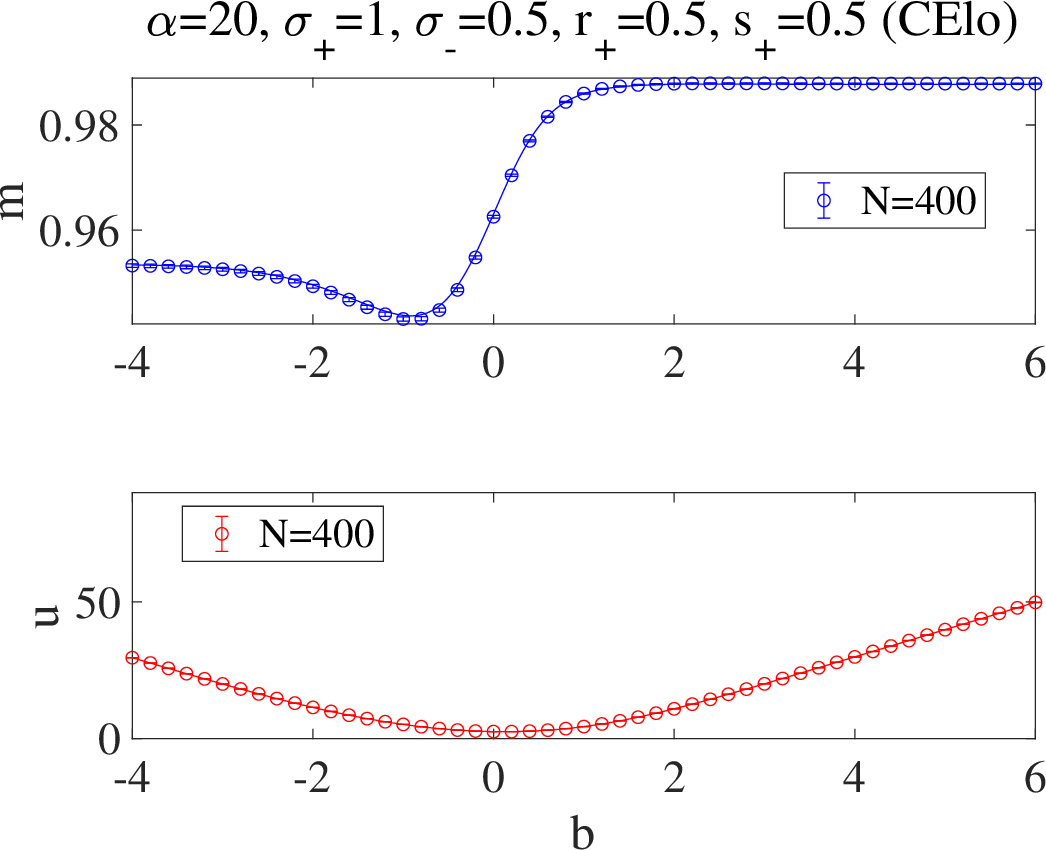}
\includegraphics[width=0.45\columnwidth]{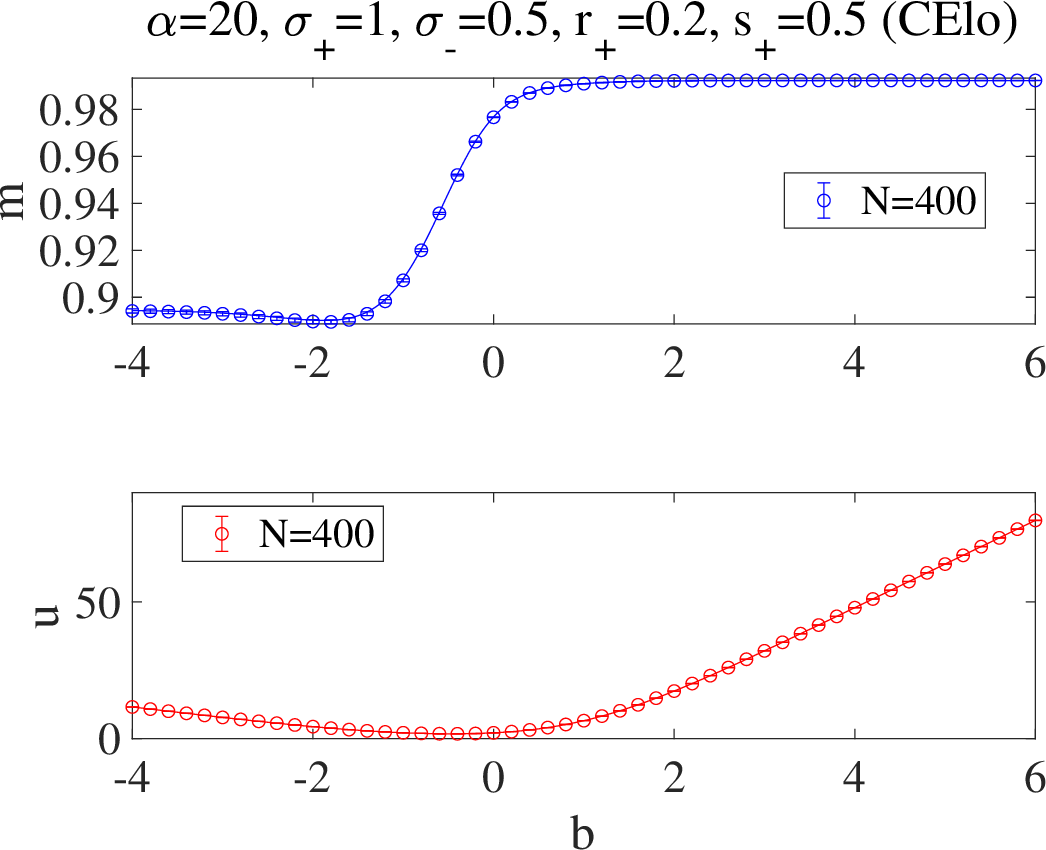}\\
(a) $s_+=0.5$ case.
\caption{Plots of $m$ and $u$ against $b$ for $r_+=0.5$ (left) and $0.2$ (right) for the nonequal-variance case $\sigma_+=1,\,\sigma_-=0.5$. Two cases  (a) $s_+=0.1$ and (b) $s_+=0.5$ are shown.
} 
\Lfig{comp_orderparam_al20_sigp10_sigm05}
\end{center}
\end{figure}
%%%%%%%%%%%%%%%%%%%%%%%
Yet again, the agreement between the numerical and the theoretical results is fairly good. Our theoretical result is thus validated even for the nonequal-variance case.

%%%%%%%%%%%%%%%%%%%%%%%%%%%%%%%%%%%%%%%%%%%%%%%%%%%%%
%%%%%%%%%%%%%%%%%%%%%%%%%%%%%%%%%%%%%%%%%%%%%%%%%%%%%
%%%%%%%%%%%%%%%%%%%%%%%%%%%%%%%%%%%%%%%%%%%%%%%%%%%%%
\section{Conclusion} 
In this paper we have studied a toy model of binary classification for investigating the effect of the resampling/reweighting on the feature learning performance in the imbalanced classification; a special aim is at providing a theoretical basis for Kang et al.'s observation that the best performance of feature learning is achieved without any resampling/reweighting. The model's feature space is $\mR^{N}$ and the class centers are assumed to be described by $\pm \V{w}_{0}/\sqrt{N}(\in \mR^N)$. The data generation is on the standard i.i.d.\ assumption with a label distribution having a control parameter of class imbalance and input distributions being allowed to have different variances on the two classes. In this setting, the feature learning performance can be quantified as the accuracy of estimating $\V{w}_0$. The analysis of this and related quantities is conducted in the high-dimensional limit $N\to \infty$ keeping the dataset size ratio $\alpha=M/N$ finite, using the replica method from statistical mechanics. 

Our theoretical analysis has revealed that the best performance of feature learning is actually achieved with no resampling/reweighting for a fairly wide range of losses and classifiers when input distributions are of equal-variance and the bias is set so that the decision boundary is located equidistantly from the two class centers: this is a desirable situation from the standard viewpoint of classification. The key of the derivation is the symmetry of the loss and the problem setup, explaining the wide applicability of the result. This provides a theoretical basis for Kang et al.'s observation through the connection between their last layer feature representation and our input vector $\V{x}$, and also implies that their learning result achieved the desirable situation. The emergence of such desirable situations as a result of learning may be understood from the viewpoint of NC~\citep{doi:10.1073/pnas.2015509117,doi:10.1073/pnas.2103091118}. More quantitative analysis has been conducted in two cases: the combinations of the cross-entropy loss and the logistic function and of the zero-one loss and the perceptron. Although we have found that optimization of the energy or the overlap over the bias $b$ yields the better overlap than the above desirable situation, the selected bias values tend to take extreme values and also be highly dependent on the choice of loss, classifier, and the parameters. The practicality of such sensitive results should be tested by experiments in more realistic situations, which constitutes interesting future work. 

Numerical simulations on the $N=400$-dimensional systems have been also performed to check the validity of our theoretical results. The result showed good consistency with the theoretical one, which verifies our theoretical results and also reinforces the practicality of our theory derived in the limit $N\to \infty$. 

As a future direction, it will be interesting to extend the analysis in this paper to multiple classes. Although we have proposed a further simplified model to treat multiclass classification on the basis of the insight from the analysis, the model is too simple in that it is essentially a one-dimensional problem with an unsupervised loss: this is incompatible with the standard treatment of multiclass classification. Such an extension would involve some difficulties related to the arrangement of the cluster centers. Some reasonable assumptions simplifying the analysis would be necessary, and the accumulated practical knowledge in machine learning communities would help find appropriate assumptions, just as Kang et al.'s observation which inspired the present study. Another interesting extension is to the semi-supervised setting with both labeled and unlabeled data. The effect of unlabeled data for performance is debatable from a theoretical viewpoint~\citep{yang2020rethinking} while the benefit is evident in some applications~\citep{jing2020self}. A toy-model study would be helpful to resolve this puzzle and to find conditions under which unlabeled data improve the performance.

%%%%%%%%%%%%%%%%%%%%%%%%%%%%%%%%%%%%%%%%%%%%%%%%%%%%%
%%%%%%%%%%%%%%%%%%%%%%%%%%%%%%%%%%%%%%%%%%%%%%%%%%%%%
%%%%%%%%%%%%%%%%%%%%%%%%%%%%%%%%%%%%%%%%%%%%%%%%%%%%%
%\subsubsection*{Acknowledgements} 
%This work is partially supported by JSPS KAKENHI Nos. 19H01812, 18K11463, 17H00764, 22K12179, and 22H05117 (TO). 

%%%%%%%%%%%%%%%%%%%%%%%%%%%%%%%%%%%%%%%%%%%%%%%%%%%%%
%%%%%%%%%%%%%%%%%%%%%%%%%%%%%%%%%%%%%%%%%%%%%%%%%%%%%
%%%%%%%%%%%%%%%%%%%%%%%%%%%%%%%%%%%%%%%%%%%%%%%%%%%%%
\appendix

%%%%%%%%%%%%%%%%%%%%%%%%%%%%%%%%%%%%%%%%%%%%%%%%%%%%%
%%%%%%%%%%%%%%%%%%%%%%%%%%%%%%%%%%%%%%%%%%%%%%%%%%%%%
%%%%%%%%%%%%%%%%%%%%%%%%%%%%%%%%%%%%%%%%%%%%%%%%%%%%%
\section{Volume computation}\Lsec{Volume}
To handle the Dirac measures, we employ the following identities and trick:
\be
&&
1
=C_1
\int dQ\delta\lb NQ-\|\V{w}_a\|^2\rb
=C_2\int dQd\hat{Q}
e^{\frac{1}{2}N\hat{Q}Q-\frac{1}{2}\hat{Q}\|\V{w}_a\|^2},
\\ &&
1=C_1
\int dq\delta\lb Nq-\V{w}_a^{\top}\V{w}_b\rb
=C_3\int dqd\hat{q}e^{-N\hat{q}q+\hat{q}\V{w}_a^{\top}\V{w}_b},
\\ &&
1=C_1
\int dm\delta\lb Nm-\V{w}_0^{\top}\V{w}_a\rb
=C_3\int dmd\hat{m}e^{-N\hat{m}m+\hat{m}\V{w}_0^{\top}\V{w}_a},
\ee
where the Fourier expression of the delta function is used to obtain the expression on the rightmost side of each equation, and where $C_1,C_2,C_3$ are normalization constants that are irrelevant in the following computations and will be discarded hereafter. Using these, we obtain 
\be
&&
V=
\int dQdqdm d\hat{Q}d\hat{q}d\hat{m}
~e^{N\lb \frac{1}{2}n\hat{Q}Q -\frac{1}{2}n(n-1)\hat{q}q -n\hat{m}m
\rb}
\no \\ &&
\times 
\Tr{\{\V{w}_a\}_{a=1}^n}
e^{
-\frac{1}{2}\hat{Q}\sum_{a=1}^{n}\|\V{w}_a\|^2
+\hat{m}\sum_{a=1}^{n}\V{w}_0^{\top}\V{w}_a
+\hat{q}\sum_{a<b}\V{w}_a^{\top}\V{w}_b
}
\no \\ &&
\eqqcolon
\int dQdqdm d\hat{Q}d\hat{q}d\hat{m}
~e^{N\lb \frac{1}{2}n\hat{Q}Q -\frac{1}{2}n(n-1)\hat{q}q -n\hat{m}m
\rb}\mc{I}.
\ee
The factor $\mc{I}$ is computed as follows:
\be
&&
\mc{I}=
\int \lb \prod_{a}d\V{w}_a \delta(N-\|\V{w}_a\|^2) \rb
e^{
-\frac{1}{2}\hat{Q}\sum_{a=1}^{n}\|\V{w}_a\|^2
+\hat{m}\sum_{a=1}^{n}\V{w}_0^{\top}\V{w}_a
+\hat{q}\sum_{a<b}\V{w}_a^{\top}\V{w}_b
}
\no \\ &&
=
\int_{-i\infty}^{i\infty} \lb \prod_{a}\frac{i d\Lambda_a}{2\pi} \rb
e^{\frac{N}{2}\sum_a\Lambda_a}
\no \\ && \times 
\int \lb \prod_{a}d\V{w}_a \rb
e^{
-\frac{1}{2}\sum_{a=1}^{n} \Lambda_a \|\V{w}_a\|^2
-\frac{1}{2}\hat{Q}\sum_{a=1}^{n}\|\V{w}_a\|^2
+\hat{m}\sum_{a=1}^{n}\V{w}_0^{\top}\V{w}_a
+\hat{q}\sum_{a<b}\V{w}_a^{\top}\V{w}_b
}
\no \\ &&
=
\int_{-i\infty}^{i\infty} \lb \prod_{a}\frac{i d\Lambda_a}{2\pi} \rb
e^{N\lb \frac{1}{2}\sum_a\Lambda_a
+\frac{1}{N}\sum_{i=1}^{N}\log 
\int Dz
\prod_{a=1}^n
\lb 
\int dw
e^{
-\frac{1}{2}\Lambda_a w^2
-\frac{1}{2}(\hat{Q}+\hat{q})w^2
+\hat{m}w_{0i}w
+\sqrt{\hat{q}}zw
}
\rb
\rb
}.
\label{eq:tmp1}
\ee
Here, the integration w.r.t.\ $\Lambda_a$ can be replaced by its extremization condition thanks to the saddle-point/Laplace method. This yields the symmetric solution $\Lambda_a^*=\Lambda^*$. Hence,
\be
&&
\mc{I}=
e^{N\lb \frac{1}{2}n\Lambda^*
+\frac{1}{N}\sum_{i=1}^{N}\log 
\int Dz
\lb 
\int dw
e^{
-\frac{1}{2}(\Lambda^* +\hat{Q}+\hat{q})w^2
+(\hat{m}w_{0i}+\sqrt{\hat{q}}z)w
}
\rb^n
\rb
}
\no \\ &&
=
e^{N\lb \frac{1}{2}n\Lambda^*
+\frac{1}{N}\sum_{i=1}^{N}\log 
\int Dz
\lb 
\sqrt{\frac{2\pi}{\Lambda^*+\hat{Q}+\hat{q}}}e^{\frac{1}{2}\frac{(\hat{m}w_{0i}+\sqrt{\hat{q}}z)^2}{\Lambda^*+\hat{Q}+\hat{q}}}
\rb^n
\rb
}.
\ee
The last term in the exponent can be rewritten in the limit $n\to 0$ as
\be
&&
\lim_{n\to 0}
\frac{1}{nN}\sum_{i=1}^{N}
\log 
\int Dz
\lb 
\sqrt{\frac{2\pi}{\Lambda^*+\hat{Q}+\hat{q}}}e^{\frac{1}{2}\frac{(\hat{m}w_{0i}+\sqrt{\hat{q}}z)^2}{\Lambda^*+\hat{Q}+\hat{q}}}
\rb^n
\no \\ &&
=
\frac{1}{2}\log(2\pi)-\frac{1}{2}\log(\Lambda^*+\hat{Q}+\hat{q})
+\frac{1}{N}\sum_{i=1}^{N}
\int Dz
\frac{1}{2}\frac{(\hat{m}w_{0i}+\sqrt{\hat{q}}z)^2}{\Lambda^*+\hat{Q}+\hat{q}}
\no \\ &&
=
\frac{1}{2}\log(2\pi)-\frac{1}{2}\log(\Lambda^*+\hat{Q}+\hat{q})
+
\frac{1}{2}\frac{\hat{m}^2+\hat{q}}{ \Lambda^*+\hat{Q}+\hat{q}},
\ee
where the relation $\sum_{i}w_{0i}^2=N$ is used in the last equality. Overall, the limit $n\to 0$ yields
\be
&&
\hspace{-10mm}
\lim_{n\to 0,N\to\infty}\frac{1}{nN} \log V(Q,q,m)
\no \\ &&
\hspace{-10mm}
=\Extr{\Lambda^*,\hat{Q},\hat{q},\hat{m} } 
\Biggl\{ 
\frac{1}{2}\hat{Q}Q +\frac{1}{2}\hat{q}q -\hat{m}m
+\frac{1}{2}\Lambda^*
+\frac{1}{2}\log(2\pi)-\frac{1}{2}\log(\Lambda^*+\hat{Q}+\hat{q})
+
\frac{1}{2}\frac{\hat{m}^2+\hat{q}}{ \Lambda^*+\hat{Q}+\hat{q}}
\Biggr\}.
\ee
The extremization conditions w.r.t.\ $\hat{Q}$ and $\Lambda^*$ are degenerating. Rewriting $\hat{Q} \to \hat{Q}-\Lambda^*$ erases this degeneracy and makes the $\Lambda^*$-dependence very simple as $(-(1/2)Q+1/2)\Lambda^*$ in the equation. The extremization condition w.r.t.\ $\Lambda^*$ thus yields $Q=1$, leading to \Req{entropy}.

\section{Interpretation of EOS}\Lsec{interp}
In this appendix, we provide an intuitive interpretation
of the EOS \eqref{eq:EOS} derived in \Rsec{Derivation of}.
We start with the first line of \eqref{eq:tmp1} for the factor $\mathcal{I}$. 
One can rewrite the exponent of the integrand as 
\begin{equation}
  -\frac{1}{2}\hat{Q}\sum_{a=1}^n\|\bm{w}_a\|^2
  +\hat{m}\sum_{a=1}^n\bm{w}_0^\top\bm{w}_a
  +\hat{q}\sum_{a<b}\bm{w}_a^\top\bm{w}_b
  =\frac{1}{2}\left\|
  \frac{\hat{m}}{\sqrt{\hat{q}}}\bm{w}_0
  +\sqrt{\hat{q}}\sum_{a=1}^n\bm{w}_a\right\|^2
  -\frac{\hat{m}^2}{2\hat{q}}\|\bm{w}_0\|^2
  -\frac{\hat{Q}+\hat{q}}{2}\sum_{a=1}^n\|\bm{w}_a\|^2.
\end{equation}
Using the Hubbard-Stratonovich transform
\begin{equation}
  e^{c\|\bm{a}\|^2/2}
  =\left(\frac{c}{2\pi}\right)^{N/2}
  \int e^{-c\|\bm{z}\|^2/2+c\bm{a}^\top\bm{z}}\,d\bm{z}
\end{equation}
with $c=\frac{1}{\hat{q}}$ and
$\bm{a}=[(\hat{m}/\sqrt{\hat{q}})\bm{w}_0+\sqrt{\hat{q}}\sum_{a=1}^n\bm{w}_a]/\sqrt{c}$,
one can rewrite the integrand as 
\begin{align}
  &e^{-\frac{\hat{Q}}{2}\sum_{a=1}^n\|\bm{w}_a\|^2
    +\hat{m}\sum_{a-1}^n\bm{w}_0^\top\bm{w}_a+\hat{q}\sum_{a<b}\bm{w}_a^\top\bm{w}_b}
  \nonumber\\
  &=e^{\frac{c}{2}\|\bm{a}\|^2-\frac{\hat{m}^2}{2\hat{q}}\|\bm{w}_0\|^2
    -\frac{\hat{Q}+\hat{q}}{2}\sum_{a=1}^n\|\bm{w}_a\|^2}
  \nonumber\\
  &=\left(\frac{1}{2\pi\hat{q}}\right)^{N/2}
  \int e^{-\frac{1}{2\hat{q}}\|\bm{z}\|^2
    -\frac{\hat{m}^2}{2\hat{q}}\|\bm{w}_0\|^2
    +\hat{m}\bm{w}_0^\top\bm{z}
    +\sum_{a=1}^n\bm{w}_a^\top\bm{z}
    -\frac{\hat{Q}+\hat{q}}{2}\sum_{a=1}^n\|\bm{w}_a\|^2}
  \,d\bm{z}
  \nonumber\\
  &=\left(\frac{1}{2\pi\hat{q}}\right)^{N/2}
  \int e^{-\frac{1}{2\hat{q}}\|\bm{z}-\hat{m}\bm{w}_0\|^2}
  \left(
  \prod_{a=1}^ne^{-\frac{1}{2\hat{q}}\|\bm{z}-\hat{q}\bm{w}_a\|^2-\frac{\hat{Q}}{2}\|\bm{w}_a\|^2}\right)
  e^{\frac{n}{2\hat{q}}\|\bm{z}\|^2}\,d\bm{z}.
\end{align}
This formula, after taking the limit $n\to0$,
implies that the problem of estimating $\bm{w}_0$ is
equivalent, in the limit $N\to\infty$, to estimating it 
from its scaled and noisy version $\bm{z}=\hat{m}\bm{w}_0+\bm{n}$
with Gaussian noise $\bm{n}\sim N(\mathbf{0},\hat{q}I_N)$
by assuming the likelihood $\propto e^{-\frac{1}{2\hat{q}}\|\bm{z}-\hat{q}\bm{w}\|^2}$
and the prior $\propto e^{-\frac{\hat{Q}}{2}\|\bm{w}\|^2}$.
The posterior distribution of $\V{w}$ given $\V{z}$
turns out to be $N\left(\frac{1}{\hat{Q}+\hat{q}}\V{z},\frac{1}{\hat{Q}+\hat{q}}I_N\right)$, and the posterior average with respect to this model
turns out to be corresponding to the average $\langle(\cdots)\rangle$
over the Boltzmann distribution defined in \Req{BAve}.

Let $\hat{\V{w}},\hat{\V{w}}'$ be independent samples
from the posterior distribution $p(\V{w} \mid \V{z})$.
One then has
\begin{align}
  \frac{\E[\hat{\V{w}}^\top\V{w}_0]}{N}
  &=\frac{1}{\hat{Q}+\hat{q}}\frac{\E[\V{w}_0^\top\V{z}]}{N}
  =\frac{\hat{m}}{\hat{Q}+\hat{q}},
  \\
  \frac{\E[\|\hat{\V{w}}\|^2]}{N}
  &=\frac{1}{(\hat{Q}+\hat{q})^2}\frac{\E[\|\V{z}\|^2]}{N}
  +\frac{1}{\hat{Q}+\hat{q}}
  =\frac{\hat{Q}+2\hat{q}+\hat{m}^2}{(\hat{Q}+\hat{q})^2},
  \label{eq:Ew^2N}
  \\
  \frac{\E[\hat{\V{w}}^\top\hat{\V{w}}']}{N}
  &=\frac{1}{(\hat{Q}+\hat{q})^2}
  \frac{\E[\|\V{z}\|^2]}{N}
  =\frac{\hat{m}^2+\hat{q}}{(\hat{Q}+\hat{q})^2},
\end{align}
which should be equal to $m=[\langle\V{w}\rangle^\top\bm{w}_0]_{D^M}/N$,
$Q=[\langle\|\V{w}\|^2\rangle]_{D^M}/N$,
and $q=[\|\langle\V{w}\rangle\|^2]_{D^M}/N$, respectively.
These provide an interpretation of the EOS:
in the limit $N\to\infty$,
estimation of $\V{w}$ may be regarded as being performed
with the Gaussian model defined as above,
whose parameters $\hat{m},\hat{Q},\hat{q}$ are to be determined
via a scalar estimation problem defined in terms of the loss $\lossS$. 
The parameters $\hat{m},\hat{Q},\hat{q}$ should be taken so that the estimate
of $\bm{w}$ has length $\sqrt{N}$, which implies that $Q=1$ holds
and hence 
$\hat{Q}+2\hat{q}+\hat{m}^2=(\hat{Q}+\hat{q})^2$ from Eq.~\eqref{eq:Ew^2N}.
Under this condition one may forget the constraints
on $\|\V{w}_a\|^2$ in \Req{tmp1}, as they will be satisfied
automatically in the limit $N\to\infty$. 

We have assumed the scaling of the order parameters
in the limit $\beta\to\infty$ as in \Req{scaling}.
In this limit, the signal-to-noise ratio of $\V{z}=\hat{m}\V{w}_0+\V{n}$
with $\V{n}\sim N(\mathbf{0},\hat{q}I_N)$ is
$\hat{m}^2/\hat{q}=\tilde{m}^2/\tilde{\chi}$, which is finite, whereas
the signal-to-noise ratio of $\V{z}$ in the likelihood model
$\propto e^{-\frac{1}{2\hat{q}}\|\bm{z}-\hat{q}\bm{w}\|^2}$ is
$\hat{q}=\beta^2\tilde{\chi}\to\infty$, implying that
the likelihood model is asymptotically noiseless. 
It can be understood as corresponding to the deterministic nature of
the minimum-loss estimator $\hat{\V{w}}=\argmin_{\V{w}:\|\V{w}\|^2=N}\mathcal{H}(\V{w}\mid D^M;b,s)$. One furthermore has 
\begin{align}
  &\hat{Q}+2\hat{q}+\hat{m}^2=(\hat{Q}+\hat{q})^2\to
  \tilde{\chi}+\tilde{m}^2=\tilde{Q}^2,
  \\
  &m=\frac{\hat{m}}{\hat{Q}+\hat{q}}\to
  m=\frac{\tilde{m}}{\tilde{Q}},
  \\
  &q=1-\frac{1}{\hat{Q}+\hat{q}}
  \to q=1-\frac{1}{\beta\tilde{Q}},
  \quad\chi=\frac{1}{\tilde{Q}}.
\end{align}
These reproduce \BReqss{Qt_sp}{chi_sp} of the EOS.

%%%%%%%%%%%%%%%%%%%%%%%%%%%%%%%%%%%%%%%%%%%%%%%%%%%%%
%%%%%%%%%%%%%%%%%%%%%%%%%%%%%%%%%%%%%%%%%%%%%%%%%%%%%
%%%%%%%%%%%%%%%%%%%%%%%%%%%%%%%%%%%%%%%%%%%%%%%%%%%%%
\section{Computational details for \Rsec{Analytical evidence}}
\Lsec{Computational}
Here we show the detailed algebras necessary for \Rsec{Analytical evidence}. For notational convenience, we write the derivative of a quantity $A$ w.r.t.\ $s_+$ as $\dot{A}:=\frac{dA}{ds_+}$, and the order parameter vectors as $\tilde{\V{\Omega}}=(\tilde{m},\tilde{\chi})^{\top}$ and $\V{\Omega}=(m,\chi)^{\top}$. 

Taking the derivatives w.r.t.\ $s_+$ of the EOS \NReq{EOS}, we can derive a set of linear equations of $\dot{\tilde{\V{\Omega}}}, \dot{\V{\Omega}}$. For example, the derivative of \Req{Qt_sp} yields
\be
\dot{\T{Q}}=\frac{\T{m}}{\T{Q}}\dot{\T{m}}+\frac{1}{2\T{Q}}\dot{\T{\chi}}.
\Leq{Qt_dot}
\ee
Similarly, taking the derivative of \BReqs{m_sp}{chi_sp} and rewriting $\dot{\T{Q}}$ by using \Req{Qt_dot}, we have a relation transforming $\dot{\T{\V{\Omega}}}$ to $\dot{\V{\Omega}}$ as
\subbe
\be
&&
\dot{\V{\Omega}}=\T{T}\dot{\T{\V{\Omega}}},
\Leq{Ot2Oapp}
\\ &&
\T{T}=
\frac{1}{\T{Q}^3}
\left(
\begin{array}{cc}
\T{\chi}  & -\frac{1}{2}\T{m}   \\
-\T{m}  & -\frac{1}{2}     
\end{array}
\right),
\Leq{Ttilde}
\ee
\subee
which corresponds to Eq.~\eqref{eq:Ot2O}. %####
In the same way, after lengthy algebras, the derivatives of \BReqs{mt_sp}{chit_sp} lead to another relation transforming $\dot{\V{\Omega}}$ to $\dot{\T{\V{\Omega}}}$:
\be
\dot{\T{\V{\Omega}}}=\V{c}+T\dot{\V{\Omega}},
\Leq{O2Otapp}
\ee
which corresponds to Eq.~\eqref{eq:O2Ot}, %####
where 
\subbe
\Leq{EOS_diff_s}
\be
&&
\dloss_{y}\coloneqq \dloss(h_{y}),
\\ &&
\dloss'_{y }\coloneqq   \frac{d\dloss(h)}{dh} \mid _{h=h_{y}},
\\ &&
D_{y}\coloneqq 1-\sigma \chi s_{y} \dloss'_{y},
\\ &&
\V{c}=
\alpha \left(
\begin{array}{c}
\sum_{y=\pm 1} yr_{y}\Dz{\frac{\dloss_y }{D_y}}  
    \\
\sum_{y=\pm 1} y r_y s_y\Dz{\frac{\dloss_y ^2}{D_y}}    
\end{array}
\right),
\\ &&
T=\left(\begin{array}{cc}
  T_{11}&T_{12}\\
  T_{21}&T_{22}
\end{array}\right),
\\ &&
T_{11}=\alpha  \lb \sum_{y=\pm 1} r_ys_y\Dz{\frac{\dloss_y '}{D_y}} \rb,
\\ &&
T_{12}=\sigma^2 \alpha \lb \sum_{y=\pm 1} r_y s_y^2 \Dz{\frac{\dloss_y \dloss_y '}{D_y}} \rb,
\\ &&
T_{21}=2\sigma^2 \alpha \lb \sum_{y=\pm 1} r_y s_y^2\Dz{\frac{\dloss_y \dloss_y '}{D_y}} \rb(=2T_{12}),
\\ &&
T_{22}=2\sigma^2 \alpha \lb  \sum_{y=\pm 1} r_y \sigma^2 s_y^3 \Dz{\frac{\dloss_y ^2\dloss_y '}{D_y}}\rb.
\ee
\subee
Note that the expressions in \Req{EOS_diff_s} assume that the loss $\lossS$ is twice differentiable on $\mR$. For losses having singularities such as the zero-one loss, those expressions should be interpreted as appropriate limits of those for their smoothed versions, as discussed in \Rsec{Derivation of}. The smoothness-controlling parameter ($\gamma$ in \Req{ell_01_smooth}) can be arbitrary because the following discussion holds irrespectively of its value. 

Using the identity 
$
\dloss'_{y}/D_{y}=\lb -1 + \frac{1}{D_{y}}\rb/(\sigma^2\chi s_{y})
$, we rewrite \Req{EOS_diff_s} as
\be
&&
T_{11}=\frac{\alpha }{\sigma^2\chi} \lb -1+\sum_{y=\pm 1} r_y\Dz{\frac{1}{D_y}} \rb,
\\ &&
T_{12}=\frac{ \alpha }{\chi} \lb -\sum_{y=\pm 1} r_y s_y \Dz{\dloss_y } + \sum_{y=\pm 1} r_y s_y\Dz{\frac{\dloss_y }{D_y}} \rb,
\\ &&
T_{21}=2T_{12},
\\ &&
T_{22}=\frac{2\sigma^2 \alpha }{\chi}
\lb
 - \sum_{y=\pm 1} r_y s_y^2\Dz{\dloss_y ^2} 
+ \sum_{y=\pm 1}r_y s_y^2 \Dz{ \frac{\dloss_y ^2}{D_y} } \rb.
\ee
Putting \BReqs{Ot2Oapp}{O2Otapp} together, we obtain the equation to $\dot{\V{\Omega}}$:
\be
\dot{\V{\Omega}}=\T{T}T\V{c}+\T{T}T\dot{\V{\Omega}} \eqqcolon \V{b}+A\dot{\V{\Omega}}.
\ee
If we assume the extremization condition $\dot{m}=\dot{\Omega}_1=0$, the following relation can be derived:
\be
-\frac{b_1}{A_{12}}=\frac{b_2}{1-A_{22}} (=\dot{\chi}).
\Leq{m_max_cond}
\ee
Our discussion completes by showing that this condition is satisfied if $s_+=s_-=1/2\eqqcolon s$. In the situation $s_+=s_-=s$ with $b=0$, the symmetry $\dloss_+ =\dloss_-\eqqcolon \dloss,~D_+ =D_-\eqqcolon D$ holds. This simplifies many terms:
\subbe
\be
&&
\T{m}\to \alpha s\Dz{\dloss},
\\ &&
\T{\chi} \to \sigma^2\alpha  s^2\Dz {\dloss^2},
\\ && 
c_{1}\to \alpha  (2r_{+}-1)\Dz {\frac{\dloss}{D}} ,
\\ && 
c_{2}\to 2\sigma^2 \alpha  s (2r_+-1)\Dz {\frac{\dloss^2}{D}} ,
\\ && 
T_{11}\to \frac{\alpha }{\sigma^2\chi} \lb -1+\Dz {\frac{1}{D}} \rb,
\\ && 
T_{12}\to \frac{\alpha   }{\chi} \lb -\Dz {\dloss'}+\Dz {\frac{\dloss'}{D}} \rb,
\\ && 
T_{21}=2T_{12},
\\ && 
T_{22}\to \frac{2\sigma^2\alpha }{\chi}\lb -\Dz {(\dloss')^2}+\Dz {\frac{(\dloss')^2}{D}}\rb.
\ee
\subee
Using these, we obtain
\be
&&
b_1=\T{T}_{11}c_1+\T{T}_{12}c_2
\to \frac{\sigma^2 \alpha ^2 s^2(2r_+-1)}{\T{Q}^3}
\lbb
\Dz {\dloss^2}\Dz {\frac{\dloss}{D}}
-\Dz {\dloss}\Dz {\frac{\dloss^2}{D}}
\rbb,
\\ &&
b_2
=\T{T}_{21}c_1+\T{T}_{22}c_2
\to 
-\frac{\alpha  s(2r_+-1)}{\T{Q}^3}
\lbb
\alpha  \Dz {\dloss}\Dz {\frac{\dloss}{D}}
+\sigma^2 \Dz {\frac{\dloss^2}{D}}
\rbb,
\\ &&
A_{12}
=\T{T}_{11}T_{12}+\T{T}_{12}T_{22}
\to 
\frac{\sigma^2 \alpha ^2 s^3}{\T{Q}^2}
\lbb
\Dz {\dloss^2}\Dz {\frac{\dloss}{D}}
-\Dz {\dloss}\Dz {\frac{\dloss^2}{D}}
\rbb,
\\ &&
A_{22}
=\T{T}_{21}T_{12}+\T{T}_{22}T_{22}
\to 
1
-
\frac{\alpha  s^2}{\T{Q}^2}
\lbb
\lb \alpha  \Dz {\dloss}\Dz {\frac{\dloss}{D}}+\sigma^2 \Dz {\frac{\dloss^2}{D}} \rb
\rbb.
\ee
In the transformations above, we have used the EOS relation $\chi=\T{Q}^{-1}$ and 
\be
\T{Q}^2=\T{m}^2+\T{\chi}\to 
\alpha ^2 s^2 \Dz {\dloss}^2+\sigma^2\alpha  s^2\Dz {\dloss^2}
\Rightarrow
\frac{\T{Q}^2}{\alpha  s^2} \to \alpha  \Dz {\dloss}^2+\sigma^2\Dz {\dloss^2}.
\ee
Then, we finally arrive at 
\be
&&
\frac{b_1}{A_{12}}=\frac{2r_+-1}{\T{Q}s}=-\frac{b_2}{1-A_{22}},
\ee
showing that \Req{m_max_cond} holds at $s_+=s_-=s$. This supports the empirical observation that the maximum of $m$ is achieved when no resampling/reweighting is applied. 

%%%%%%%%%%%%%%%%%%%%%%%%%%%%%%%%%%%%%%%%%%%%%%%%%%%%%
\subsubsection*{Acknowledgments}\Lsec{Acknowledgements}
This work was partially supported by JST, CREST Grant number JPMJCF1862, Japan (TO, TT), JSPS KAKENHI under Grant 22K12179 (TO), Grant-in-Aid for Transformative Research Areas (A), ``Foundation of machine learning physics'' (22H05117) (TO), and Grant-in-Aid for Transformative Research Areas (A), ``Shin-biology regulated by protein lifetime'' (24H01895) (TT).

%%%%%%%%%%%%%%%%%%%%%%%%%%%%%%%%%%%%%%%%%%%%%%%%%%%%%
%%%%%%%%%%%%%%%%%%%%%%%%%%%%%%%%%%%%%%%%%%%%%%%%%%%%%
%%%%%%%%%%%%%%%%%%%%%%%%%%%%%%%%%%%%%%%%%%%%%%%%%%%%%
%\bibliographystyle{plain}
\bibliography{obuchi}
\bibliographystyle{tmlr}

\end{document}